\documentclass{article}

\usepackage[affil-it]{authblk}

\usepackage{sectsty,relsize}
\sectionfont{\Large\selectfont}
\subsectionfont{\relsize{1.2}\normalsize\selectfont}
\subsubsectionfont{\normalsize\selectfont}

\usepackage{geometry}
\usepackage{url}
\usepackage{hyperref}
\usepackage{natbib}
\usepackage{graphicx}

\usepackage{subcaption}

\usepackage{amsmath,amssymb,mathtools,xfrac}

\usepackage{wrapfig}
\usepackage[shortlabels]{enumitem}

\usepackage{setspace}
\usepackage[linesnumbered,ruled]{algorithm2e}
\SetKwBlock{Cycle}{}{}
\SetAlCapHSkip{0em}

\usepackage{xifthen}
\usepackage{amsmath}
\usepackage{amsfonts}

\renewcommand{\vec}[1]{{\boldsymbol{#1}}}

\usepackage{color}

\newcommand{\textsub}[1]{\textsc{\tiny #1}} 

\usepackage{pifont}
\makeatletter
\newcommand{\colorlabel}[1]{%
	\global\tag@true%
	\nonumber%
	\refstepcounter{equation}%
	\gdef\df@tag{\maketag@@@{{\color{#1}(\theequation)}}\def\@currentlabel{\theequation}}}
\makeatother

\newcommand{\bigmid}{\;\ifnum\currentgrouptype=16 \middle\fi|\;} 
\newcommand{\mytilde}[0]{\mathds{\raise.17ex\hbox{$\scriptstyle\sim$}}} 

\DeclareMathOperator{\EV}{\mathbb{E}}
\newcommand{\EVV}[2]{\EV_{#1}\!\left[{#2}\right]}

\newcommand{\gaussian}{\mathcal N}
\newcommand{\realspace}{\mathbb R}

\newcommand{\bigO}{\mathcal{O}}

\newcommand{\action}{a}
\newcommand{\state}{s}
\newcommand{\params}{\theta}

\newcommand{\context}{c}

\newcommand{\reward}{R}

\newcommand{\statespace}{\mathcal S}
\newcommand{\actionspace}{\mathcal A}

\newcommand{\dstate}{d_{\textsub{\state}}}
\newcommand{\daction}{d_{\textsub{\action}}}

\providecommand{\rwd}{\mathcal{R}}
\providecommand{\prob}{\mathcal{P}}

\newcommand{\rmodel}[1][]{%
	\ifthenelse{\equal{#1}{}}{\rwd\left(\state,\action\right)}{\rwd\left(\state^{[#1]},\action^{[#1]}\right)}%
}
\newcommand{\rmodelt}[1][]{%
	\ifthenelse{\equal{#1}{}}{\rwd\left(\state_t,\action_t\right)}{\rwd\left(\state_t^{[#1]},\action_t^{[#1]}\right)}%
}

\newcommand{\pmodel}[1][]{%
	\ifthenelse{\equal{#1}{}}{\prob\left(\state'|\state,\action\right)}{\prob\left(\state^{[#1+1]}|\state^{[#1]},\action^{[#1]}\right)}%
}
\newcommand{\pmodelt}[1][]{%
	\ifthenelse{\equal{#1}{}}{\prob\left(\state_{t+1}|\state_t,\action_t\right)}{\prob\left(\state_{t+1}^{[#1]}|\state_t^{[#1]},\action_t^{[#1]}\right)}%
}

\newcommand{\rmodelctx}[1][]{%
	\ifthenelse{\equal{#1}{}}{\rwd\left(\context,\params\right)}{\rwd\left(\context^{[#1]},\params^{[#1]}\right)}%
}
\newcommand{\berror}[1][]{%
	\ifthenelse{\equal{#1}{}}{\delta\left(\state,\action\right)}{\delta_#1\left(\state,\action\right)}%
}

\newcommand{\vecrmodel}[1][]{%
	\ifthenelse{\equal{#1}{}}{\vec\rwd\left(\state,\action\right)}{\rwd\left(\state^{[#1]},\action^{[#1]}\right)}%
}

\usepackage{xspace}
\newcommand{\coeff}{\kappa}

\usepackage{booktabs,array}

\newcount\rowc

\makeatletter
\def\ttabular{%
	\hbox\bgroup
	\let\\\cr
	\def\rulea{\ifnum\rowc=\@ne \hrule height 1.3pt \fi}
	\def\ruleb{
		\ifnum\rowc=1\hrule height 1.3pt \else
		\ifnum\rowc=6\hrule height \heavyrulewidth 
		\else \hrule height \lightrulewidth\fi\fi}
	\valign\bgroup
	\global\rowc\@ne
	\rulea
	\hbox to 10em{\strut \hfill##\hfill}%
	\ruleb
	&&%
	\global\advance\rowc\@ne
	\hbox to 10em{\strut\hfill##\hfill}%
	\ruleb
	\cr}
\def\endttabular{%
	\crcr\egroup\egroup}

\begin{document}

\title{\textbf{Long-Term Visitation Value for Deep Exploration \\in Sparse-Reward Reinforcement Learning}}

\author[1]{S.~Parisi\thanks{Corresponding author: \texttt{sparisi@fb.com}}}
\author[2]{D.~Tateo}
\author[2]{M.~Hensel}
\author[2]{C.~D'Eramo}
\author[2]{J.~Peters}
\author[3]{J.~Pajarinen}
\date{\small Preprint (under review)}

\affil[1]{\small Meta AI Research, United States}
\affil[2]{\small Technische Universit{\"a}t Darmstadt, Germany}
\affil[3]{Aalto University, Finland}

\maketitle

\hrule height 1pt
\subsubsection*{Abstract}
Reinforcement learning with sparse rewards is still an open challenge. Classic methods rely on getting feedback via extrinsic rewards to train the agent, and in situations where this occurs very rarely the agent learns slowly or cannot learn at all. Similarly, if the agent receives also rewards that create suboptimal modes of the objective function, it will likely prematurely stop exploring. 
More recent methods add auxiliary intrinsic rewards to encourage exploration. 
However, auxiliary rewards lead to a non-stationary target for the Q-function. 
In this paper, we present a novel approach that (1) plans exploration actions far into the future by using a long-term visitation count, and (2) decouples exploration and exploitation by learning a separate function assessing the exploration value of the actions. 
Contrary to existing methods which use models of reward and dynamics, our approach is off-policy and model-free. 
We further propose new tabular environments for benchmarking exploration in reinforcement learning.
Empirical results on classic and novel benchmarks show that the proposed approach outperforms existing methods in environments with sparse rewards, especially in the presence of rewards that create suboptimal modes of the objective function. Results also suggest that our approach scales gracefully with the size of the environment. 
\\
The source code is available at \url{https://github.com/sparisi/visit-value-explore}

\bigskip
\noindent\emph{Keywords:} reinforcement learning, sparse reward, exploration, upper confidence bound, off-policy
\bigskip
\hrule height 1pt

\section{Introduction}
\label{sec:intro}
Reinforcement learning (RL) is a process where an agent learns how to behave in an environment by trial and error. The agent performs actions and, in turn, the environment may provide a \textit{reward}, i.e., a feedback assessing the quality of the action. The goal of RL is then to learn a \textit{policy}, i.e., a function producing a sequence of actions yielding the maximum cumulative reward.
Despite its simplicity, RL achieved impressive results, such as learning to play Atari videogames from pixels \citep{mnih2013playing,schulman2017proximal}, or beating world-class champions at Chess, Go and Shogi \citep{silver2017amastering}. However, a high-quality reward signal is typically necessary to learn the optimal policy, and without it RL algorithms may perform poorly even in small environments \citep{osband2019deep}.

\clearpage

\noindent\textbf{Characteristics of high-quality rewards.} 
One important factor that defines the quality of the reward signal is the \textit{frequency} at which rewards are emitted. Frequently emitted rewards are called ``dense'', in contrast to infrequent emissions which are called ``sparse''. Since improving the policy relies on getting feedback via rewards, the policy cannot be improved until a reward is obtained. In situations where this occurs very rarely, the agent learns slowly or cannot learn at all. 
Furthermore, reinforcement signals should encourage the agent to find the best actions to solve the given task. However, the environment may also provide \textit{distracting} rewards, i.e., rewards that create suboptimal modes of the objective function. In this case, the agent should be able to extensively explore the environment without prematurely converge to locally optimal solutions caused by distracting rewards.
\\[5pt]
\textbf{Reward engineering.}
In the presence of sparse distracting rewards, efficiently exploring the environment to learn the optimal policy is challenging.
Therefore, many RL algorithms rely on well-shaped reward functions such as quadratic costs.
For example, in a collect-and-deliver task, we could design an additional reward function that rewards proximity to the items to be collected.
These well-shaped functions help to guide the agent towards good solutions and avoid bad local optima. 
However, from the perspective of autonomous learning, this so-called reward engineering is unacceptable for three reasons.
First, the commonly utilized reward functions heavily restrict the solution space and may prevent the agent from learning optimal behavior (especially if no solution to the task is known). 
Second, it is easy to misspecify the reward function and cause unexpected behavior. For instance, by getting excessively high rewards for proximity to the items the agent may never learn to deliver them.
Third, reward engineering requires manual tuning of the reward function and the manual addition of constraints. This process requires expertise and the resulting reward function and constraints may not transfer to different tasks or environments.

\begin{figure}[t]
	\centering
	\includegraphics[width=\linewidth]{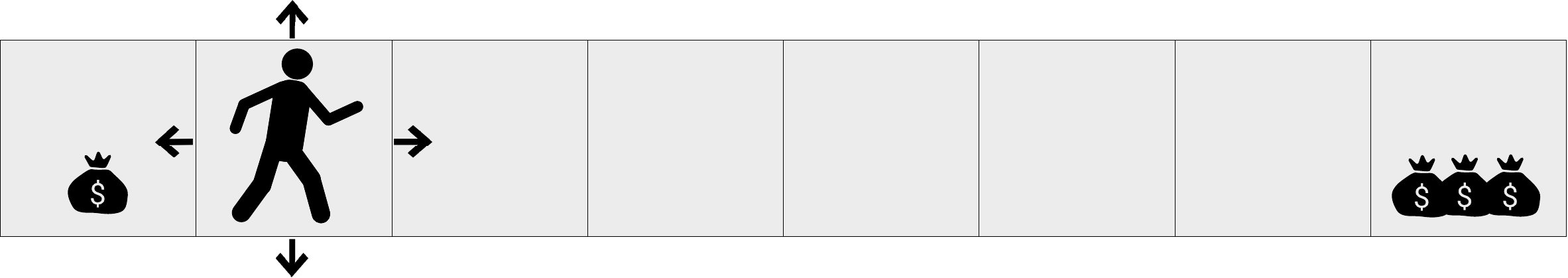}
	\caption{\label{fig:chain}A simple problem where naive exploration performs poorly. The agent starts in the second leftmost cell and is rewarded only for finding a treasure. It can move up / down / left / right, but if it moves up or down its position resets. 
	Acting randomly at each step takes on average $4^N$ attempts to find the rightmost and most valuable reward, where $N$ is the number of cells to the right of the agent.
	As no reward is given in any intermediate cell, $\epsilon$-greedy exploration 
	is likely to converge to the local optimum represented by the leftmost reward if $\epsilon$ decays too quickly.}
\end{figure}

\textit{In this paper, we address the problem of reinforcement learning with \textit{sparse} rewards without relying on reward engineering, or requiring prior knowledge about the environment or the task to solve. In particular, we highlight the problem of learning in the presence of rewards that create suboptimal modes of the objective function, which we call ``distractors''.}
\\[5pt]
\textbf{The exploration-exploitation dilemma.} Learning when rewards are sparse and possibly distracting challenges classic RL algorithms. Initially, the agent has no knowledge about the environment and needs to explore it in search for feedback. As it gathers rewards, the agent develops a better understanding of the environment and can estimate which actions are ``good'', i.e., yield positive rewards, and which are not.
Then, at any point, it needs to decide whether it should explore the environment in search for better rewards by executing new actions, or exploit its knowledge and execute known good actions.
This ``exploration-exploitation dilemma'' is frequently addressed naively by dithering \citep{mnih2013playing, lillicrap2015continuous}. In continuous action spaces Gaussian noise is added to the action, while in discrete spaces actions are chosen $\epsilon$-greedily, i.e., optimally with probability $1\! -\! \epsilon$ and randomly with probability $\epsilon$. Typically, the initial noise and $\epsilon$ are large and then decay over time. 
These approaches work in environments where random sequences of actions are likely to cause positive rewards. However, if rewards are sparse or are given only for specific sequences of actions, it is very unlikely that the agent will receive any feedback. In this case, the worst-case sample complexity for learning the optimal policy with dithering exploration is exponential in the number of states and actions \citep{kakade2003sample, szepesvari2010algorithms, osband2016generalization}. Furthermore, if distractor rewards are easily reachable, the agent may prematurely converge to local optima. An example of this problem is depicted in Figure \ref{fig:chain}.
\\[5pt]
\textbf{Deep exploration.}
Well-performing exploration strategies should solve long-horizon problems with sparse and possibly distracting rewards in large state-action spaces while remaining computationally tractable. This can be achieved if the agent takes several coherent actions to explore unknown states instead of just locally choosing the most promising actions. These behaviors are usually referred as ``deep exploration'' and ``myopic exploration'', respectively \citep{osband2016deep}.
RL literature has a long history of algorithms for deep exploration, even though the term ``deep'' became popular only in recent years.
The first algorithms to guarantee full exploration of tabular environments date back to \citet{kearns2002near} and \citet{brafman2002r}. These algorithms learn a model of the environment, i.e., the transition and reward functions, and plan actions accordingly to a state-action visitation count in order to favor less visited states.
Since then, other model-based \citep{jaksch2010near,hester2013texplore} as well as model-free algorithms \citep{strehl2006pac,bellemare2016unifying,jin2018iq,dong2020q} have been proposed. However, despite their strong guarantees, these algorithms either require to learn the complete model of the environment, rely on optimistic initialization, or have impractically high sample complexity.

In continuous environments, intrinsic motivation and bootstrapping are the most prominent approaches.
The former defines an additional \textit{intrinsic} reward added to the environment \textit{extrinsic} reward. If the extrinsic reward is sparse, the intrinsic reward can fill the gaps between the sparse signals, possibly giving the agent quality feedback at every timestep. This approach can be used in conjunction with any RL algorithm by just providing the modified reward signal. 
However, the quality of exploration strongly depends on the intrinsic reward, which may not scale gracefully with the extrinsic reward and therefore needs hand-tuning. Furthermore, combining exploration and exploitation by summing intrinsic and extrinsic rewards can result in undesired behavior due to the non-stationarity of the augmented reward function.
Moreover, convergence is harder to guarantee \citep{kolter2009near}, and in general, there is no agreement on the definition of the best intrinsic reward \citep{houthooft2016vime,pathak2017curiosity}.
Bootstrapping, instead, is used to estimate the actions value posterior distribution over the environment, which then drives exploration. Because actions are selected considering the level of uncertainty associated with their value estimates, bootstrapping incentivizes experimentation with actions of highly uncertain value and, thus, induces exploration \citep{osband2019deep}.
However, these methods rely on approximated posteriors and usually lack guarantees of convergence, unless either the environment model is learned or a time-dependent policy is used \citep{osband2019deep}.
\\[5pt]
\textbf{Deep exploration via long-term visitation value.}
\textit{In this paper, we present a novel approach that (1) plans exploration actions far into the future using a long-term visitation count, and (2) decouples exploration and exploitation by learning a separate function assessing the exploration value of the actions.}
Contrary to existing methods that use models of reward and dynamics, our approach is off-policy and model-free. 

\textit{We further comprehensively benchmark our approach against existing algorithms on several environments}, stressing the challenges of learning with sparse and distracting rewards. 
Empirical results show that the proposed approach outperforms existing algorithms, and suggest that it scales gracefully with the size of the environment.

\section{Preliminaries}
\label{sec:preliminaries}
We start with a description of the RL framework, providing the reader with the notation used in the remainder of the paper. Subsequently, we review the most prominent exploration strategies in literature, discuss their shortcomings, and identify open challenges.

\subsection{Reinforcement Learning and Markov Decision Processes}
\label{ssec:mdp}
We consider RL in an environment governed by a Markov Decision Process (MDP). An MDP is described by the tuple $\langle \statespace, \actionspace, \prob, \rwd, \mu_1 \rangle$, where $\statespace$ is the state space, $\actionspace$ is the action space, $\pmodel$ defines a Markovian transition probability density between the current $\state$ and the next state $\state'$ under action
$\action$, $\rmodel$ is the reward function, and $\mu_1$ is initial distribution for state $s_1$.
The state and action spaces could be finite, i.e., $|\statespace| = \dstate$ and $|\actionspace| = \daction$, or continuous, i.e., $\statespace \subseteq \realspace^{\dstate}$ and $\actionspace \subseteq \realspace^{\daction}$. The former case is often called \textit{tabular} MDP, as a table can be used to store and query all state-action pairs if the size of the state-action space is not too large. If the model of the environment is known, these MDPs can be solved exactly with dynamic programming.
The latter, instead, requires function approximation, and algorithms typically guarantee convergence to local optima.
In the remainder of the paper, we will consider only tabular MDPs and only model-free RL, i.e., the model of the environment is neither known nor learned.

Given an MDP, we want to learn to act. Formally, we want to find a \emph{policy} $\pi(a|s)$ to take an appropriate action $a$ when the agent is in state $s$. By following such a policy starting at initial state $s_1$, we obtain a sequence of states, actions and rewards $(\state_t, \action_t, r_t)_{t=1\ldots H}$, where $r_t = \rwd(s_t, a_t)$ is the reward at time $t$, and $H$ is the total timesteps also called \textit{horizon}, which can possibly be infinite. We refer to such sequences as \emph{trajectories} or \emph{episodes}.
Our goal is thus to find a policy that maximizes the expected return
\begin{align}
	Q^*(s,a) &:= \max_\pi \:\: \EVV{\mu_\pi(\state) \pi(\action|\state)}{Q^{\pi}(\state,\action)}, \label{eq:maxQ}
	\\
	\textrm{where} \:\:\:
	Q^{\pi}(\state_t,\action_t) &:= \EVV{\prod_{i=t}^H \pi(a_{i+1}|s_{i+1})\prob(s_{i+1}|s_{i},a_{i})}{\sum_{i=t}^H \gamma^{i-t} r_{i}},  \label{eq:Q}
\end{align}
where $\mu_\pi(\state)$ is the (discounted) state distribution under $\pi$, i.e., the probability of visiting state $\state$ under $\pi$, and $\gamma \in [0,1)$ is the discount factor which assigns weights to rewards at different timesteps. $Q^{\pi}(\state,\action)$ is the action-state value function (or Q-function) which is the expected return obtained by executing $a$ in state $s$ and then following $\pi$ afterwards. 
In tabular MDPs, the Q-function can be stored in a table, commonly referred as Q-table, and in stationary infinite-horizon MDPs the greedy policy $\pi^*(a|s) =\arg\max_a Q^*(s,a)$ is always optimal \citep{puterman1994markov}. 
However, the Q-function of the current policy is initially unknown and has to be learned. 

\paragraph{Q-learning.}
For tabular MDPs, Q-learning by \citet{watkins1992q} was one of the most  important breakthroughs in RL, and it is still at the core of recent successful algorithms. 
At each training step $i$, the agent chooses an action according to a \textit{behavior} policy $\beta(a|s)$, i.e., $a_t \sim \beta(\cdot|s_t)$, and then updates the Q-table according to
\begin{align}
	Q_{i+1}^\pi(s_t,a_t) &= Q_i^\pi(s_t,a_t) + \eta \delta(s_t,a_t,s_{t+1})
	\\
	\delta(s_t,a_t,s_{t+1}) &= 	\begin{cases}
	\textcolor{blue}{r_t + \gamma \max_a Q_i^\pi(s_{t+1},a)} - Q_i^\pi(s_t,a_t) & \text{if $s_t$ is non-terminal}
	\\
	\textcolor{blue}{r_t} - Q_i^\pi(s_t,a_t) & \text{otherwise}
	\end{cases}
\end{align}
where $\eta$ is the learning rate and $\delta(s,a,s')$ is the \textit{temporal difference (TD) error}. The blue term is often referred to as \textit{TD target}. This approach is called \textit{off-policy} because the policy $\beta(a|s)$ used to explore the environment is different from the target greedy policy $\pi(a|s)$.

\paragraph{The behavior policy and the ``exploration-exploitation dilemma''.}
As discussed in Section \ref{sec:intro}, an important challenge in designing the behavior policy is to account for the exploration-exploitation dilemma. 
\textit{Exploration} is a long-term endeavor where the agent tries to maximize the possibility of finding better rewards in states not yet visited. On the contrary, \textit{exploitation} maximizes the expected rewards according to the current knowledge of the environment.
Typically, in continuous action spaces Gaussian noise is added to the action, while in discrete spaces actions are chosen $\epsilon$-greedily, i.e., optimally with probability $1\! -\! \epsilon$ and randomly with probability $\epsilon$. In this case, the exploration-exploitation dilemma is simply addressed by having larger noise and $\epsilon$ at the beginning, and then decaying them over time. 
This naive dithering exploration is extremely inefficient in MDPs with sparse rewards, especially if some of them are ``distractors'', i.e., rewards that create suboptimal modes of the objective function. In the remainder of this section, we review more sophisticated approaches addressing exploration with sparse rewards. We first show theoretically grounded ideas for tabular or small MDP settings which are generally computationally intractable for large MDPs. We then give an overview of methods that try to apply similar principles to large domains by making approximations.

\subsection{Related Work}
Since finding high-value states is crucial for successful
RL, a large body of work has been devoted to
efficient exploration in the past years. We begin by discussing an optimal solution to exploration, then continue with confidence-bound-based approaches. Then, we discuss methods based on intrinsic motivation, and describe posterior optimality value-distribution-based methods. We conclude by
stating the main differences between our approach and these
methods.
\\[5pt]
\textbf{Optimal solution to exploration.} In this paper, we consider model-free RL, i.e., the agent does not know the MDP dynamics. 
It is well-known that in model-free RL the optimal solution to exploration can be found by a Bayesian approach. First,
assign an initial (possibly uninformative) prior distribution over
possible unknown MDPs. Then, at each time step the posterior belief
distribution over possible MDPs can be computed using the Bayes' theorem
based on the current prior distribution, executed action, and made
observation. The action that optimally explores is then found
by planning over possible future posterior distributions, e.g., using tree search, sufficiently far forward. Unfortunately,
optimal exploration is intractable even for very small
tasks~\citep{szepesvari2010algorithms,strens2000bayesian,poupart2006analytic}. The worst-case computational effort is exponential w.r.t. the planning horizon in tabular MDPs, and can be even more challenging with continuous state-action spaces.
\\[5pt]
\textbf{Optimism.}
In tabular MDPs, many of the provably efficient algorithms are
based on {optimism in the face of uncertainty} (OFU)
\citep{lai1985asymptotically}. In OFU, the agent acts greedily
w.r.t. an optimistic action value estimate composed of the value
estimate and a bonus term that is proportional to the uncertainty of
the value estimate. After executing an optimistic action, the agent
then either experiences a high reward learning that the value of the action was indeed high, or the agent experiences a low reward and learns that the action was not optimal. After visiting a state-action pair, the exploration bonus is reduced. This approach is superior to naive approaches in that it avoids actions where low value and low information gain are possible. Under the assumption that the agent can visit every
state-action pair infinitely many times, the overestimation will
decrease and almost optimal behavior can be obtained
\citep{kakade2003sample}. Most algorithms are optimal up to a
polynomial amount of states, actions, or horizon length. RL literature provides many variations of these algorithms which use
bounds with varying efficacy or different simplifying assumptions, such as \citep{kearns2002near,brafman2002r,kakade2003sample,auer2007logarithmic,jaksch2010near,dann2015sample}.
\\[5pt]
\textbf{Upper confidence bound.}
Perhaps the best well-known OFU algorithm is the upper confidence
bound (UCB) algorithm~\citep{auer2002finite}. UCB chooses actions based on a
bound computed from visitation counts: the lower the count, the higher the bonus. Recently, \citet{jin2018iq}
proved that episodic Q-learning with UCB exploration converges to a
regret proportional to the square root of states, actions, and timesteps, and to the square root of the cube of the episode
length. \citet{dong2020q} gave a similar proof for infinite horizon
MDPs with sample complexity proportional to the number of states and
actions. In our experiments, we compare our approach to Q-learning
with UCB exploration as defined by \citet{auer2002finite}.
\\[5pt]
\textbf{Intrinsic motivation.}
The previously discussed algorithms are often computationally intractable,
or their guarantees no longer apply in continuous state-action
spaces, or when the state-action spaces are too
large. Nevertheless, these provably efficient algorithms inspired more practical algorithms. Commonly, exploration is conducted by adding a bonus reward,
also called auxiliary reward, to interesting states. This is often
referred to as intrinsic motivation~\citep{ryan2000intrinsic}. For
example, the bonus can be similar to the
UCB~\citep{strehl2008analysis,bellemare2016unifying}, thus encouraging actions of high uncertainty or of low visitation count.
Recent methods, instead, compute the impact of actions based on state embeddings and reward the agent for performing actions changing the environment~\citep{raileanu2020ride}. 
\\
Other forms of bonus are based on the prediction error of some
quantity. For example, the agent may learn a model of the dynamics and try to predict the next state \citep{stadie2015incentivizing,houthooft2016vime,pathak2017curiosity,schmidhuber1991possibility,schmidhuber2006developmental}. By giving a bonus proportional to the prediction error, the agent is incentivized to explore unpredictable states. Unfortunately, in the presence of noise, unpredictable states are not necessarily interesting states.
Recent work addresses this issue by training an ensemble of models \citep{pathak2019self} or predicting the output of a random neural network \citep{burda2018exploration}.

As we will discuss in Section \ref{ssec:vv}, our approach has
similarities with the use of auxiliary reward. However, the
use of auxiliary rewards can be inefficient for long-horizon
exploration. Our approach addresses this issue by using a decoupled
long-horizon exploration policy and, as the experiments in
Section~\ref{sec:eval} show, it outperforms auxiliary-reward-based
approaches~\citep{strehl2008analysis}.
\\[5pt]
\textbf{Thompson sampling.}
Another principled well-known technique for exploration is
Thompson sampling~\citep{thompson1933likelihood}. Thompson sampling
samples actions from a posterior distribution which specifies the
probability for each action to be optimal. Similarly to UCB-based methods, Thompson sampling is guaranteed to converge to an optimal policy in multi-armed bandit problems~\citep{kaufmann2012thompson,agrawal2013further}, and has shown
strong empirical performance~\citep{scott2010modern,chapelle2011empirical}. For a discussion of known
shortcomings of Thompson sampling, we refer to~\citep{russo2014blearning,russo2017time,russo2018tutorial}.
\\
Inspired by these successes, recent methods have followed
the principle of Thompson sampling in
Q-learning~\citep{osband2013more,osband2019deep,deramo2019exploiting}.
These methods 
assume that an empirical distribution
over Q-functions --an ensemble of randomized Q-functions-- is similar to the
distribution over action optimalities used in Thompson sampling. Actions are thus sampled from such ensemble. Since the
Q-functions in the ensemble become similar when updated with new samples,
there is a conceptual similarity with the action optimality
distribution used in Thompson sampling for which variance also
decreases with new samples. While there are no general proofs for randomized Q-function approaches, \citet{osband2019deep} proved a bound on the Bayesian regret of an algorithm based on
randomized Q-functions in tabular time-inhomogeneous MDPs with a
transition kernel drawn from a Dirichlet prior, providing a starting
point for more general proofs.
\\
Other exploration methods approximating a distribution over action
optimalities include for example the work of \citet{fortunato2017noisy} and \citet{plappert2018parameter}, who apply noise to the parameters of the model. This may be interpreted as approximately inferring a posterior
distribution \citep{gal2015dropout}, although this is not without contention \citep{osband2016risk}. \citet{osband2016deep} more
directly approximates the posterior over Q-functions through
bootstrapping \citep{efron1982jackknife}. The lack of a proper prior
is an issue when rewards are sparse, as it causes the uncertainty of
the posterior to vanish quickly. \citet{osband2018randomized} and
\citet{osband2019deep} try to address this by enforcing a prior
via regularization or by adding randomly initialized but fixed neural
network on top of each ensemble member, respectively.
\\
Methods based on posterior distributions have yielded high performance in some tasks. However,
because of the strong approximations needed to model posterior optimality
distributions instead of maintaining visitation counts or explicit
uncertainty estimates, these approaches may often have problems in
assessing the state uncertainty correctly.
In our experiments, the proposed approach outperforms state-of-the-art methods based on approximate posterior distributions, namely the algorithms proposed by~\citet{osband2016deep,osband2019deep} and \citet{deramo2019exploiting}.

\medskip
Above, we discussed state-of-the-art exploration methods that (1) use
confidence bounds or distribution over action optimalities to take
the exploration into account in a principled way, (2) modify the reward by adding an intrinsic reward signal to encourage exploration, and (3) approximate a posterior distribution over
value functions. The main challenge that pertains to all these methods
is that they do not take long-term
exploration into account explicitly. 
\textit{Contrary to this,
similarly to how value iteration propagates state values, we propose an approach that uses dynamic programming to explicitly propagate visitation information backward in time.} 
This allows our approach to efficiently
find the most promising parts of the state space that have not been
visited before, and avoid getting stuck in suboptimal 
regions that at first appear promising.

\clearpage

\section{Long-Term Visitation Value for Deep Exploration}
\label{sec:vv}
In this section, we present our novel off-policy method for improving exploration with sparse and distracting rewards.
We start by motivating our approach with a toy example, showing the limitation of current count-based algorithms and the need for ``long-term visitation counts''. Subsequently, we formally define the proposed method, and show how it solves the toy example.

\subsection{Example of the Limitations of Existing Immediate Count Methods}
\label{ssec:toy_example}
\begin{wrapfigure}{l}{0.34\textwidth}
	\begin{center}
		\vspace*{-0.7cm}
		\includegraphics[width=0.9\linewidth]{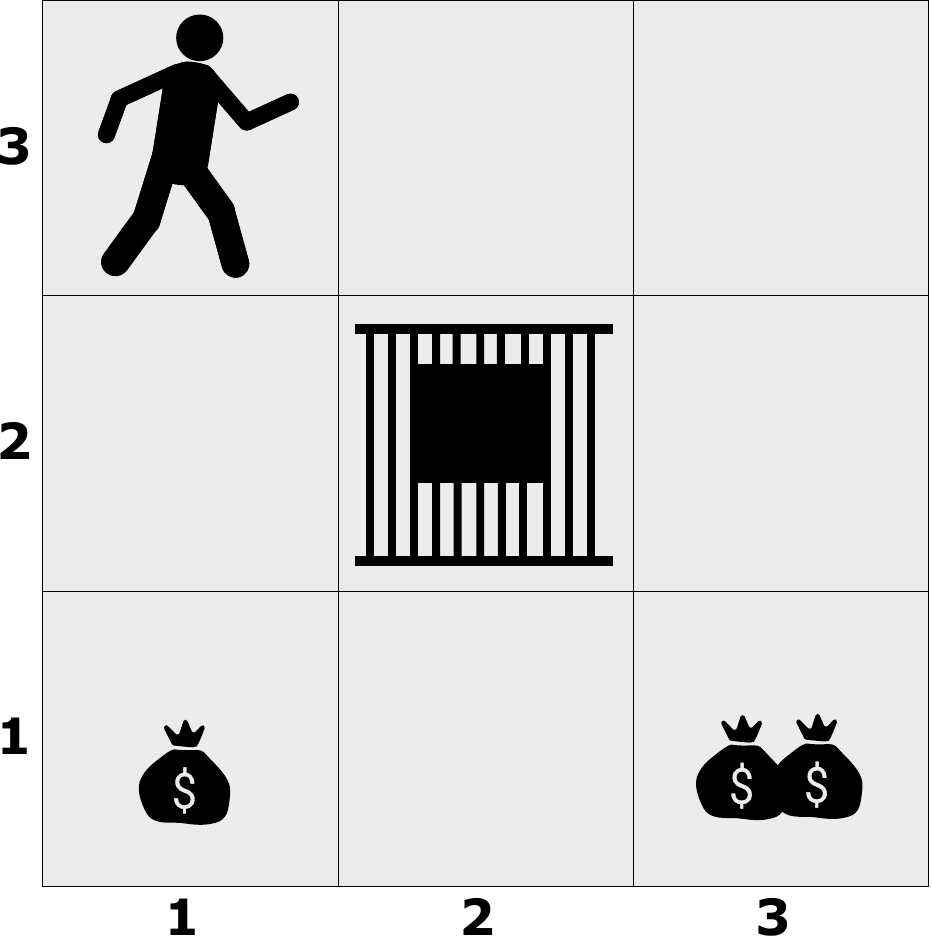}
	\end{center}
	\vspace*{-0.5cm}
	\caption{Toy problem domain.}
	\label{fig:toy_env}
	\vspace*{-0.5cm}
\end{wrapfigure}
Consider the toy problem of an agent moving in a grid, shown in Figure \ref{fig:toy_env}. Reward is 0 everywhere except in $(1,1)$ and $(3,1)$, where it is 1 and 2, respectively. The reward is given for executing an action in the state, i.e., on transitions from reward states.
The agent initial position is fixed in $(1,3)$ and the episode ends when a reward is collected or after five steps.
Any action in the prison cells $(2,2)$ has only an arbitrarily small probability of success. If it fails, the agent stays in $(2,2)$. In practice, the prison cell almost completely prevents any further exploration. Despite its simplicity, this domain highlights two important challenges for exploration in RL with sparse rewards. First, there is a locally optimal policy collecting the lesser reward, which acts as distractor. Second, the prison state is particularly dangerous as it severely hinders exploration. 

\medskip

\begin{figure}[t]
	\centering 
	\begin{subfigure}[t]{.24\linewidth} 
		\centering 
		\includegraphics[width=\linewidth]{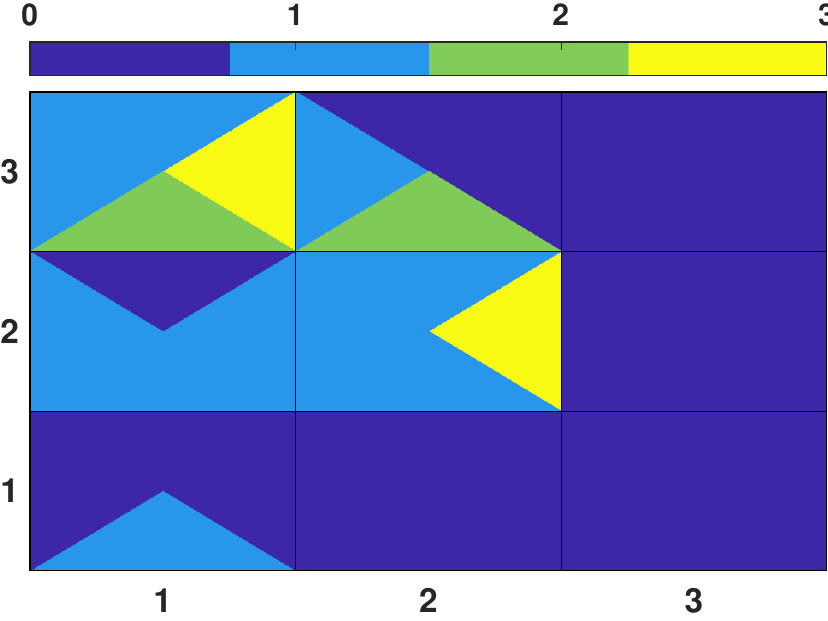}
		\caption{\label{fig:toy_vca}State-action count}
	\end{subfigure}
	\hfill
	\begin{subfigure}[t]{.24\linewidth} 
		\centering 
		\includegraphics[width=\linewidth]{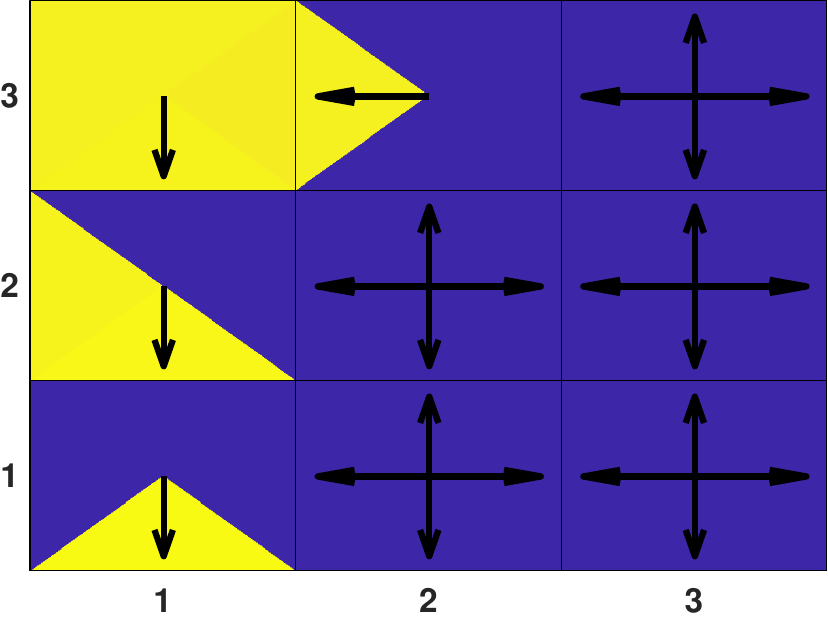}
		\caption{\label{fig:toy_greedy}Greedy} 
	\end{subfigure} 
	\hfill
	\begin{subfigure}[t]{.24\linewidth} 
		\centering 
		\includegraphics[width=\linewidth]{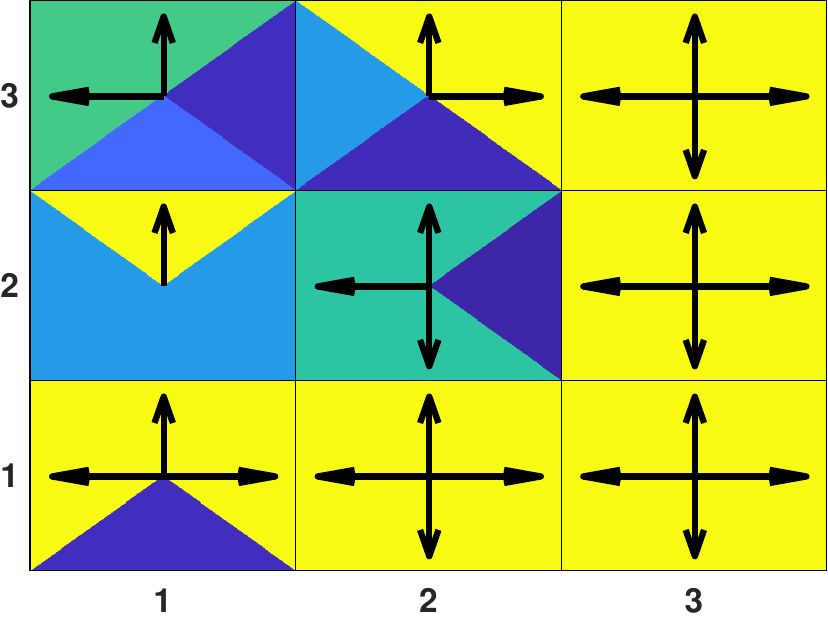}
		\caption{\label{fig:toy_ucb}UCB1}
	\end{subfigure}
	\hfill
	\begin{subfigure}[t]{.24\linewidth} 
		\centering 
		\includegraphics[width=\linewidth]{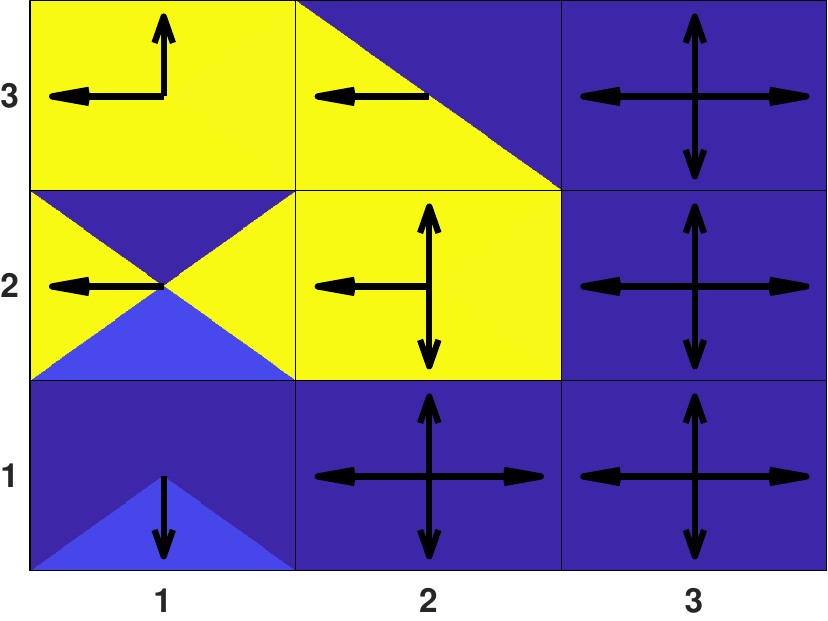}
		\caption{\label{fig:toy_mbie}Greedy (with bonus)} 
	\end{subfigure} 
	\caption{\label{fig:toy_example}\textbf{Toy example (part 1).} Visitation count (\subref{fig:toy_vca}) and behavior policies (\subref{fig:toy_greedy}, \subref{fig:toy_ucb}, \subref{fig:toy_mbie}) for the toy problem below. Cells are divided into four triangles, each one representing the count (\subref{fig:toy_vca}) or the behavior Q-value (\subref{fig:toy_greedy}, \subref{fig:toy_ucb}, \subref{fig:toy_mbie}) for the actions ``up / down / left / right'' in that cell. Arrows denote the action with the highest Q-value. None of the behavior policies exhibits long-term deep exploration. In (1,3) both UCB1 (\subref{fig:toy_ucb}) and the augmented Q-table (\subref{fig:toy_mbie}) myopically select the action with the smallest immediate count. The latter also acts badly in (1,2) and (2,3). There, it selects ``left'' which has count one, instead of ``up'' which has count zero.}
\end{figure}

Assume that the agent has already explored the environment for four episodes performing the following actions: (1) left, down, left, down, down (reward 1 collected); (2) up, down, right, right (fail), right (fail); (3) right, left, right, down, down (fail); (4) right, down, left (fail), up (fail), right (fail). The resulting state-action visitation count is shown in Figure \ref{fig:toy_vca}. Given this data, we initialize $Q^\pi(s,a)$ to zero, train it with Q-learning with $\gamma = 0.99$ until convergence, and then derive three greedy policies. The first simply maximizes $Q^\pi(s,a)$. The second maximizes the behavior $Q^\beta(s,a)$ based on UCB1 \citep{auer2002finite}, defined as
\begin{align}
Q^\beta(s,a) &= Q^\pi(s,a) + \coeff \sqrt{\frac{2\log \sum_{a_j} n(s_t,a_j)}{n(s_t,a)}},\label{eq:ucb1}
\end{align}
where $n(s,a)$ is the state-action count, and $\coeff$ is a scaling coefficient\footnote{In classic UCB, the reward is usually bounded in $[0,1]$ and no coefficient is needed. This is not usually the case for MDPs, where Q-values can be larger/smaller than 1/0. We thus need to scale the square root with $\coeff = (r_{\max} - r_{\min}) / (1 - \gamma)$, which upper-bounds the difference between the largest and smallest possible Q-value.}.
The third maximizes an augmented Q-function $Q^+(s,a)$ trained by adding the intrinsic reward based on the immediate count \citep{strehl2008analysis}, i.e.,
\begin{equation}
r^+_t = r_t + {\alpha}\:{{n(s_t,a_t)}^{-1/2}},\label{eq:exp_bonus}
\end{equation} 
where $\alpha$ is a scaling coefficient set to $\alpha = 0.1$ as in \citep{strehl2008analysis,bellemare2016unifying}.
Figure \ref{fig:toy_vca} shows the state-action count after the above trajectories, and Figures \ref{fig:toy_greedy}, \ref{fig:toy_ucb}, and \ref{fig:toy_mbie} depict the respective Q-functions (colormap) and policies (arrows). Given the state-action count, consider what the agent has done and knows so far.
\begin{enumerate}[a)]
\setlength{\itemsep}{0pt}
\setstretch{0.9}
\item It has executed all actions at least once in (1,3) and (2,2) (the prison cell),
\item It still did not execute ``up'' in (1,2),
\item It visited (1,3) once and experienced the reward of 1, and
\item It still did not execute ``up'' and ``right'' in (2,1).
\end{enumerate}
In terms of exploration, the best decision would be to select actions not executed yet, such as ``up'' in (1,2). \textit{But what should the agent do in states where it has already executed all actions at least once? Ideally, it should select actions driving it to unvisited states, or to states where it has not executed all actions yet.}
However, none of the policies above exhibits such behavior. 

\begin{enumerate}[a)]
\setlength{\itemsep}{0pt}
\item The greedy policy (Figure \ref{fig:toy_greedy}), as expected, points to the only reward collected so far.
\item The UCB1 policy (Figure \ref{fig:toy_ucb}) yields Q-values of much larger magnitude. Most of the actions have been executed few times, thus their UCB is larger than their Q-value. The policy correctly selects unexecuted actions in (1,2) and (2,3), but fails (1,3). There it also selects the least executed actions (`up'' and ``left'') which, however, have already been executed once. The agent should know that these actions do not bring it anywhere (the agent hits the environment boundaries and does not move) and should avoid them. However, it selects them because the policy $Q^\beta(s,a)$ is based on the \textit{immediate} visitation count.
Only when the count will be equal for all actions, the policy will select a random action. This behavior is clearly myopic and extremely inefficient.
\item The policy learned using the auxiliary bonus (Figure \ref{fig:toy_mbie}) is even worse, as it acts badly not only in (1,3) but also in (1,2) and (2,3). There, it chooses ``left'', which was already selected once, instead of ``up'' (or ``right in (2,3)), which instead has not been selected yet.
The reason for such behavior is that this auxiliary reward requires optimistic initialization \citep{strehl2008analysis}. The agent, in fact, is positively rewarded for simply executing any action, with value proportional to the visitation count. However, if the Q-function is initialized to zero, the positive feedback will make the agent believe that the action is good. This mechanism encourages the agent to execute the same action again and hinders exploration.
This problem is typical of all methods using an auxiliary reward based on visitation count, unless optimistic initialization of the Q-table is used \citep{strehl2008analysis}. More recent methods use more sophisticated rewards \citep{jin2018iq,dong2020q}.
However, despite their strong theoretical convergence guarantees, for large-size problems these methods may require an impractical number of samples and are often outperformed by standard algorithms. 
\end{enumerate}

\subsection{The Long-Term Visitation Value Approach}
\label{ssec:vv}
Despite its simplicity, the above toy problem highlights the limitations of current exploration strategies based on the immediate count. More specifically, (1) considering the immediate count may result in shallow exploration, and (2) directly augmenting the reward to train the Q-function, i.e., coupling exploration and exploitation, can have undesired effects due to the non-stationarity of the augmented reward function.
To address these limitations, we propose an approach that (1) plans exploration actions far into the future by using a \textit{long-term visitation count}, and (2) decouples exploration and exploitation by learning a separate function assessing the \textit{exploration value} of the actions. 
Contrary to existing methods that use models of reward and dynamics~\citep{hester2013texplore}, which can be hard to come by, our approach is off-policy and model-free.

\paragraph{Visitation value.}
The key idea of our approach is to train a visitation value function using the temporal difference principle. The function, called W-function and denoted by $W^\beta(s,a)$, approximates the cumulative sum of an intrinsic reward $r^W$ based on the visitation count, i.e.,
\begin{align}
W^\beta(s_t, a_t) &:= \EVV{\prod_{i=t} \beta(a_{t+1}|s_{t+1}) \prob(s_{i+1}|s_i,a_i)}{ \sum_{i=t}^H \gamma_w^{i-t} r^W_{i}},
\end{align}
where $\gamma_w$ is the visitation count discount factor. The higher the discount, the more in the future the visitation count will look. 
In practice, we estimate $W^\beta(s,a)$ using TD learning on the samples
produced using the behavior policy $\beta$. 
Therefore, we use the following update 
\begin{align}
W_{i+1}^\beta(s_t,a_t) &= W_i^\beta(s_t,a_t) + \eta \delta^W(s_t,a_t,s_{t+1}),
\end{align}
where $\eta$ is the learning rate, and $\delta^W(s_t,a_t,s_{t+1})$ is the W-function TD error, which depends on the reward $r^W$ (can be either maximized or minimized).
Subsequently, following the well-known upper confidence bound (UCB) algorithm \citep{lai1985asymptotically}, the behavior policy is greedy w.r.t. the sum of the Q-function and an upper confidence bound, i.e.,
\begin{align}
	\beta(a|s) &= \arg\max_{a}\left\lbrace Q^\pi(s,a) + \coeff U_W(s,a)\right\rbrace, \label{eq:vv_beta_generic}
\end{align}
where $U_W(s,a)$ is an upper confidence bound based on $W^\beta(s,a)$, and $\coeff$ is the same exploration constant used by UCB1 in Eq. \eqref{eq:ucb1}.
This proposed approach (1) considers long-term visitation count by employing the W-function, which is trained to maximize the cumulative --not the immediate-- count. Intuitively, the W-function encourages the agent to explore state-actions that have not been visited before. Given the current state $s$, in fact, $W^\beta(s,a)$ specifies how much exploration we can expect on (discount-weighted) average in future states if we choose action $a$. 
Furthermore, since the Q-function is still trained greedily w.r.t. the extrinsic reward as in Q-learning, while the W-function is trained greedily on the intrinsic reward and favors less-visited states, our approach (2) effectively decouples exploration and exploitation. This allows us to more efficiently prevent underestimation of the visitation count exploration term.
Below, we propose two versions of this approach based on two different rewards $r^W$ and upper bounds $U_W(s,a)$.

\paragraph{W-function as long-term UCB.} The visitation reward is the UCB of the state-action count at the current time, i.e., 
\begin{align}
r_t^W &= \begin{cases}
\sqrt{\frac{2\log \sum_{a_j} n(s_t,a_j)}{ n(s_t,a_t)}} & \text{if $s_t$ is non-terminal}
\\
\frac{1}{1 - \gamma_w}\sqrt{\frac{2\log \sum_{a_j} n(s_t,a_j)}{ n(s_t,a_t)}} & \text{otherwise.}
\end{cases}
\label{eq:vv3_rwd}
\intertext{This W-function represents the discount-weighted average UCB along the trajectory, i.e.,
}
W^\beta_{\textsub{ucb}}(s_{t},a_{t}) &= \sum_{i=t}^H \gamma_w^{i-t} \sqrt{\frac{2\log \sum_{a_j} n(s_{i},a_j)}{ n(s_{i},a_{i})}},
\intertext{and its TD error is}
\delta^W(s_t,a_t,s_{t+1}) &= 	\begin{cases}
r^W_t + \gamma_w \max_a W_{\textsub{ucb},i}^\beta(s_{t+1},a) - W_{\textsub{ucb},i}^\beta(s_t,a_t) & \text{if $s_t$ is non-terminal}
\\
r^W_t - W_{\textsub{ucb},i}^\beta(s_t,a_t) & \text{otherwise}. \label{eq:vv3_td}
\end{cases}
\end{align}
Notice how we distinguish between terminal and non-terminal states for the reward in Eq.~\eqref{eq:vv3_rwd}. The visitation value of terminal states, in fact, is equal to the visitation reward alone, as denoted by the TD target in Eq.~\eqref{eq:vv3_rwd}.
Consequently, the visitation value of terminal states could be significantly smaller than the one of other states, and the agent would not visit them again after the first time. This, however, may be detrimental for learning the Q-function, especially if terminal states yield high rewards. For this reason, we assume that the agent stays in terminal states forever, getting the same visitation reward at each time step. The limit of the sum of the discounted reward is the reward in Eq.~\eqref{eq:vv3_rwd}.
Furthermore, given proper initialization of $W^\beta_{\textsub{ucb}}$ and under some assumptions, the target of non-terminal states will still be higher than the one of terminal states despite the different visitation reward, if the count of the former is smaller.
We will discuss this in more detail later in this section.
\\
Another key aspect of this W-function regards the well-know overestimation bias of TD learning.
Since we can only update the W-function approximately based on limited samples, it is beneficial to overestimate its target with the $\max$ operator. While it is known that in highly stochastic environments the overestimation can degrade the performance of value-based algorithm~\citep{hasselt2010double, deramo2017estimating}, it has been shown that underestimation does not perform well when the exploration is challenging~\citep{tateo2017exploiting}, and for a wide class of environments, overestimation allows finding the optimal solution due to self-correction~\citep{sutton2018reinforcement}. 
\\
Assuming identical visitation over state-action pairs, the W-function becomes the discounted immediate UCB, i.e.,
\begin{align}
W^\beta_{\textsub{ucb}}(s_{t},a_{t}) &= \frac{1}{1 - \gamma_w}\sqrt{\frac{2\log \sum_{a_j} n(s_{t},a_j)}{ n(s_{t},a_{t})}}, \label{eq:vv3_limit}
\intertext{thus, we can use $W^\beta_{\textsub{ucb}}(s,a)$ directly in the behavior policy}
\beta(a_t|s_{t}) &= \arg\max_a \Big\lbrace Q^\pi(s_{t},a) + \coeff\underbrace{ (1 - \gamma_w) W^\beta_{\textsub{ucb}}(s_{t},a)}_{U_{W}^{\textsub{ucb}}(s,a)} \Big\rbrace,
\label{eq:vv3_beta}
\end{align}
In Section \ref{ssec:vv_init}, we discuss how to initialize $W^\beta_{\textsub{ucb}}(s,a)$. Finally, notice that when $\gamma_w = 0$, Eq.~\eqref{eq:vv3_limit} is the classic UCB, and Eq.~\eqref{eq:vv3_beta} is equivalent to the UCB1 policy in Eq.~\eqref{eq:ucb1}. 

\paragraph{W-function as long-term count.} The visitation reward is the state-action count at the current time, i.e.,
\begin{align}
r_t^W &= \begin{cases}
{n(s_{t},a_t)} & \text{if $s_t$ is non-terminal}
\\
\frac{1}{1 - \gamma_w} {n(s_{t},a_t)} & \text{otherwise}.
\end{cases}\label{eq:vv2_rwd}
\intertext{This W-function represents the discount-weighted average count along the trajectory, i.e.,
}
W^\beta_{\textsub{n}}(s_{t},a_{t}) &= \sum_{i=t}^H \gamma_w^{i-t} n(s_i, a_i),
\intertext{and its TD error is}
\delta^W(s_t,a_t,s_{t+1}) &= 	\begin{cases}
r^W_t + \gamma_w \min_a W_{\textsub{n},i}^\beta(s_{t+1},a) - W_{\textsub{n},i}^\beta(s_t,a_t) & \text{if $s_t$ is non-terminal}
\\
r^W_t - W_{\textsub{n},i}^\beta(s_t,a_t) & \text{otherwise}.
\end{cases}\label{eq:vv2_td}
\end{align}
Once again we distinguish between terminal and non-terminal states, but in the TD error the $\max$ operator has been replaced by the $\min$ operator. 
Contrary to the previous case, in fact, the lower the W-function the better, as we want to incentivize the visit of lower-count states.
If we would use only the immediate count as reward for terminal states, the agent would be constantly driven there. Therefore, similarly to the UCB-based reward, we assume that the agent stays in terminal states forever getting the same reward.
\\
In the TD target, since we want to prioritize low-count state-action pairs, the $\min$ operator yields an optimistic estimate of the W-function, because the lower the W-function the more optimistic the exploration is. 
Assuming identical visitation over state-action pairs, the W-function becomes the discounted cumulative count, i.e.,
\begin{align}
W^\beta_{\textsub{n}}(s_{t},a_{t}) &= \frac{1}{1 - \gamma_w} n(s_{t},a_t).\label{eq:vv2_limit}
\intertext{We then compute the pseudocount $\hat n(s_t,a_t) = {(1 - \gamma_w) W^\beta_{\textsub{n}}(s_t,a_t)}$ and use it in the UCB policy}
\beta(a_t|s_t) &= \arg\max_a \Big\lbrace Q^\pi(s_t,a) + \coeff\underbrace{ \sqrt{\frac{2\log \sum_{a_j} \hat n(s_t,a_j)}{\hat n(s_t,a)}}}_{U_{W}^{\textsub{n}}(s,a)} \Big\rbrace. \label{eq:vv2_beta}
\end{align}
When the counts are zero, $W^\beta_{\textsub{n}}(s,a) = 0$, thus we need to initialize it to zero. We discuss how to avoid numerical problem in the square root in Section \ref{ssec:vv_init}. Finally notice that when $\gamma_w = 0$, Eq.~\eqref{eq:vv2_beta} is equivalent to the UCB1 policy in Eq.~\eqref{eq:ucb1}.


\begin{figure}[t]
	\centering 
	\begin{subfigure}[t]{.325\linewidth} 
		\centering 
		\includegraphics[width=\linewidth]{img/grid_3x3_vca-eps-converted-to.pdf}
		\caption{\label{fig:toy_vca2}State-action count}
	\end{subfigure}
	\hfill
	\begin{subfigure}[t]{.325\linewidth} 
		\centering 
		\includegraphics[width=\linewidth]{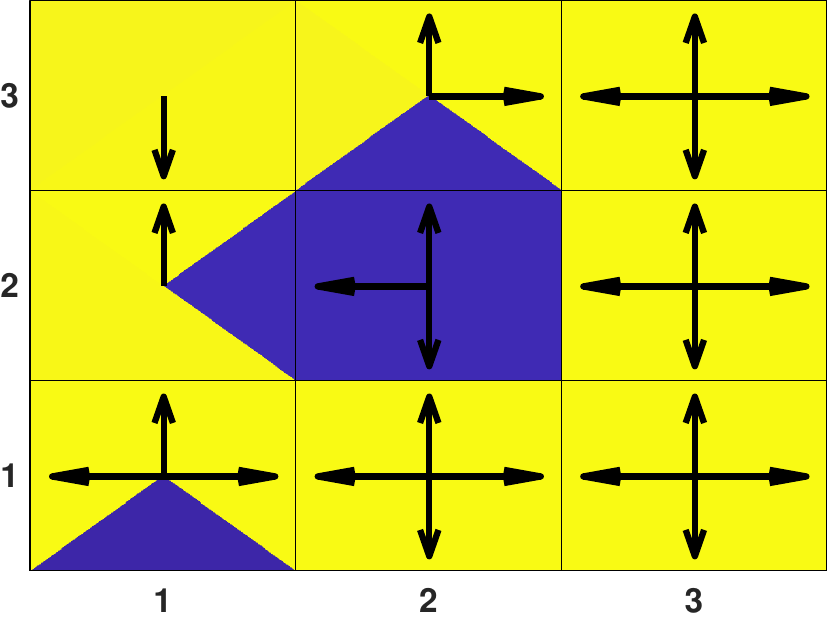}
		\caption{\label{fig:toy_vv3}UCB W-function policy} 
	\end{subfigure} 
	\hfill
	\begin{subfigure}[t]{.325\linewidth} 
		\centering 
		\includegraphics[width=\linewidth]{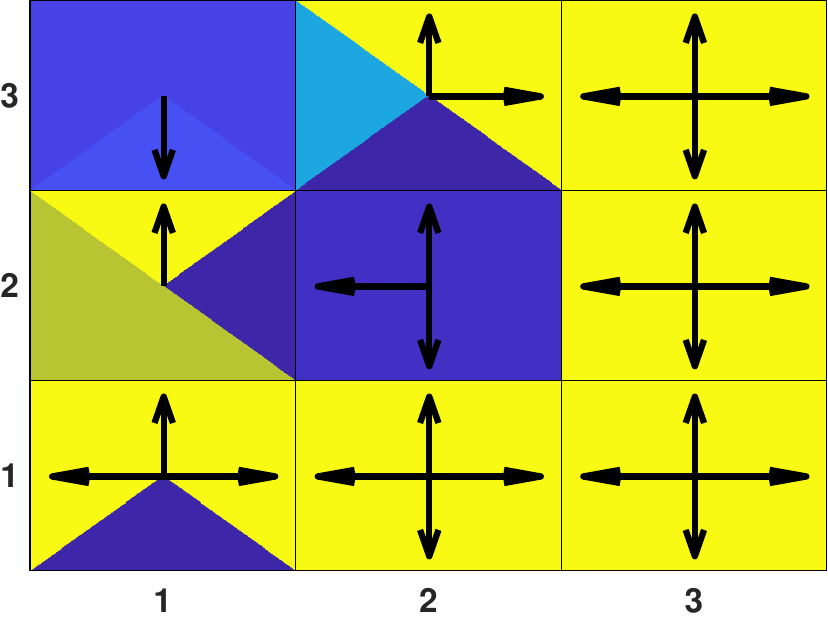}
		\caption{\label{fig:toy_vv2}Count W-function policy}
	\end{subfigure}
	\centering 
	\caption{\label{fig:toy_example_contd}\textbf{Toy example (part 2).} Visitation count (\subref{fig:toy_vca2}) and behavior policies derived from the proposed W-functions (\subref{fig:toy_vv3}, \subref{fig:toy_vv2}). Due to the low count, the exploration upper bound dominates the greedy Q-value, and the proposed behavior policies successfully achieve long-term exploration. Both policies prioritize unexecuted actions, and avoid terminal and prison states.}
\end{figure}

\paragraph{Example of Long-Term Exploration with the W-function.}
Consider again the toy problem presented in Section \ref{ssec:toy_example}. Figure \ref{fig:toy_example_contd} shows the behavior policies of Eq.~\eqref{eq:vv3_beta} and Eq.~\eqref{eq:vv2_beta} with $\gamma_w = 0.99$. The policies are identical and both drive the agent to unvisited states, or to states where it has not executed all actions yet, successfully exhibiting long-term exploration. The only difference is the magnitude of the state-action pair values (the colormap, especially in unvisited states) due to their different initialization.
\begin{enumerate}[a)]
\setlength{\itemsep}{0pt}
\setstretch{0.9}
\item In (1,3), they both select ``down'' despite ``up'' and ``left'' having lower count. The agent, in fact, knows that from (1,2) it can go ``down'' to (1,1) where it executed only one action. Thus, it knows that there it can try new actions. On the other hand, the only thing it knows about (2,3) is that it leads to the prison state, where the agent already executed all actions. Thus, (1,2) has higher long-term exploration value than (2,3).
\item In (1,2) both select ``up'', overturning the greedy action ``down'' (Figure \ref{fig:toy_greedy}). This is important, because to fully explore the environment we need to select unexecuted actions.
\item The same happens in (2,3), where actions with count zero are selected, rather than the greedy ones.
\item None of the policies lead the agent to the prison state, because it has already been visited and all actions have been executed in there.
\item They both do not select ``right'' in (2,2) because it has been executed one more time than other actions. Nonetheless, the difference in the action value is minimal, as shown by the action colormap (almost uniform).
\end{enumerate}

\subsection{W-function Initialization and Bounds}
\label{ssec:vv_init}
In UCB bandits, it is assumed that each arm/action will be executed once before actions are executed twice. In order to enforce this, we need to make sure that the upper confidence bound $U_W(s,a)$ is high enough so that an action cannot be executed twice before all other actions are executed once.
Below, we discuss how to initialize the W-functions and set bounds in order to achieve the same behavior.
For the sake of simplicity, below we drop subscripts and superscripts, i.e., we write $Q(s,a)$ in place of $Q^\pi(s,a)$, $W(s,a)$ in place of $W^\beta_{\textsub{ucb}}(s,a)$ in the first part, and $W(s,a)$ in place of $W^\beta_{\textsub{n}}(s,a)$ in the second part.

\paragraph{$W^\beta_{\textsub{ucb}}$ upper bound.}
To compute the initialization value we consider the worst-case scenario, that is, in state $s$ all actions $a$ but $\bar{a}$ have been executed. In this scenario, we want that $W(s,\bar{a})$ is an upper bound of $W(s,a)$. Following Eq. \eqref{eq:vv3_beta} and \eqref{eq:vv3_td}, we have
\begin{align}
Q_{\max} + \coeff (1 - \gamma_w) W(s,\bar{a}) &= 
Q_{\max} +
\coeff (1 - \gamma_w) \left(\textcolor{blue}{\sqrt{2 \log(|\actionspace| - 1)} + \gamma_w \max\nolimits_a W(s',a)}\right), \label{eq:vv3_init0}
\intertext{where the blue term is the TD target for non-terminal states, and $\sqrt{2 \log(|\actionspace| - 1)}$ is the immediate visitation reward for executing $\bar{a}$ in the worst-case scenario (all other $|\actionspace| - 1$ actions have been executed, and $\hat{a}$ has not been executed yet). In this scenario, $W(s,\bar{a})$ is an upper bound of $W(s,a)$ under the assumption of uniform count, i.e.,}
\forall (s,a) \neq (s,\bar{a}), & \quad n(s,a) = \bar{n}(s) \Rightarrow  W(s,\bar{a}) \geq W(s,a).
\intertext{We can then write Eq. \eqref{eq:vv3_init0} as}
Q_{\max} + \coeff (1 - \gamma_w) W(s,\bar{a}) &= 
Q_{\max} + \coeff (1 - \gamma_w) \left(\textcolor{blue}{\sqrt{2 \log(|\actionspace| - 1)} + \gamma_w W(s,\bar{a})}\right).
\end{align}
In order to guarantee to select $\bar{a}$, which has not been select yet, we need to initialize $W(s,\bar{a})$ such that the following is satisfied
\begin{align}
Q_{\min} + \coeff (1 - \gamma_w) W(s,\bar{a}) &> Q_{\max} + \coeff (1 - \gamma_w) \left(\sqrt{2 \log(|\actionspace| - 1)} + \gamma_w W(s,\bar{a})\right),
\intertext{where $Q_{\min}$ denotes the smallest possible Q-value. We get}
W(s,\bar{a}) &> \frac{Q_{\max} - Q_{\min}}{\coeff (1 - \gamma_w)} + \sqrt{2 \log(|\actionspace| - 1)} + \gamma_w W(s,\bar{a}) \nonumber
\\
W(s,\bar{a}) &> \frac{Q_{\max} - Q_{\min}}{\coeff (1 - \gamma_w)^2} + \frac{\sqrt{2 \log(|\actionspace| - 1)}}{(1 - \gamma_w)}. \label{eq:vv3_init}
\end{align}
Initializing the W-function according to Eq. \eqref{eq:vv3_init} guarantees to select $\bar{a}$.

\paragraph{$W^\beta_{\textsub{n}}$ upper bound.}
Since this W-function represents the discounted cumulative count, it is initialized to zero. 
However, numerical problems may occur in the square root of Eq. \eqref{eq:vv2_beta} because the pseudocount can be smaller than one or even zero. In these cases, the square root of negative numbers and the division by zero are not defined.
To address these issues, we add +1 inside the logarithm and replace the square root with an upper bound when $\hat{n}(s,a) = 0$. Similarly to the previous section, the bound needs to be high enough so that an action cannot be executed twice before all other actions are executed once. Following Eq. \eqref{eq:vv2_beta}, we need to ensure that 
\begin{align}
Q_{\min} + \coeff U_W(s,\bar{a}) &> Q_{\max} + \coeff U_W(s,a), \nonumber
\\
U_W(s,\bar{a}) &> \frac{Q_{\max} - Q_{\min}}{\coeff} + U_W(s,a). \label{eq:vv2_condition}
\end{align}
However, this time assuming uniform count does not guarantee a uniform \textit{pseudocount}, i.e., 
\begin{equation}
\forall (s,a) \neq (s,\bar{a}), \quad n(s,a) = \bar{n}(s) \nRightarrow W(s,\bar{a}) \geq W(s,a).
\end{equation}

\begin{wrapfigure}{l}{0.45\textwidth}
	\begin{center}
		\vspace*{-0.7cm}
		\includegraphics[width=0.99\linewidth]{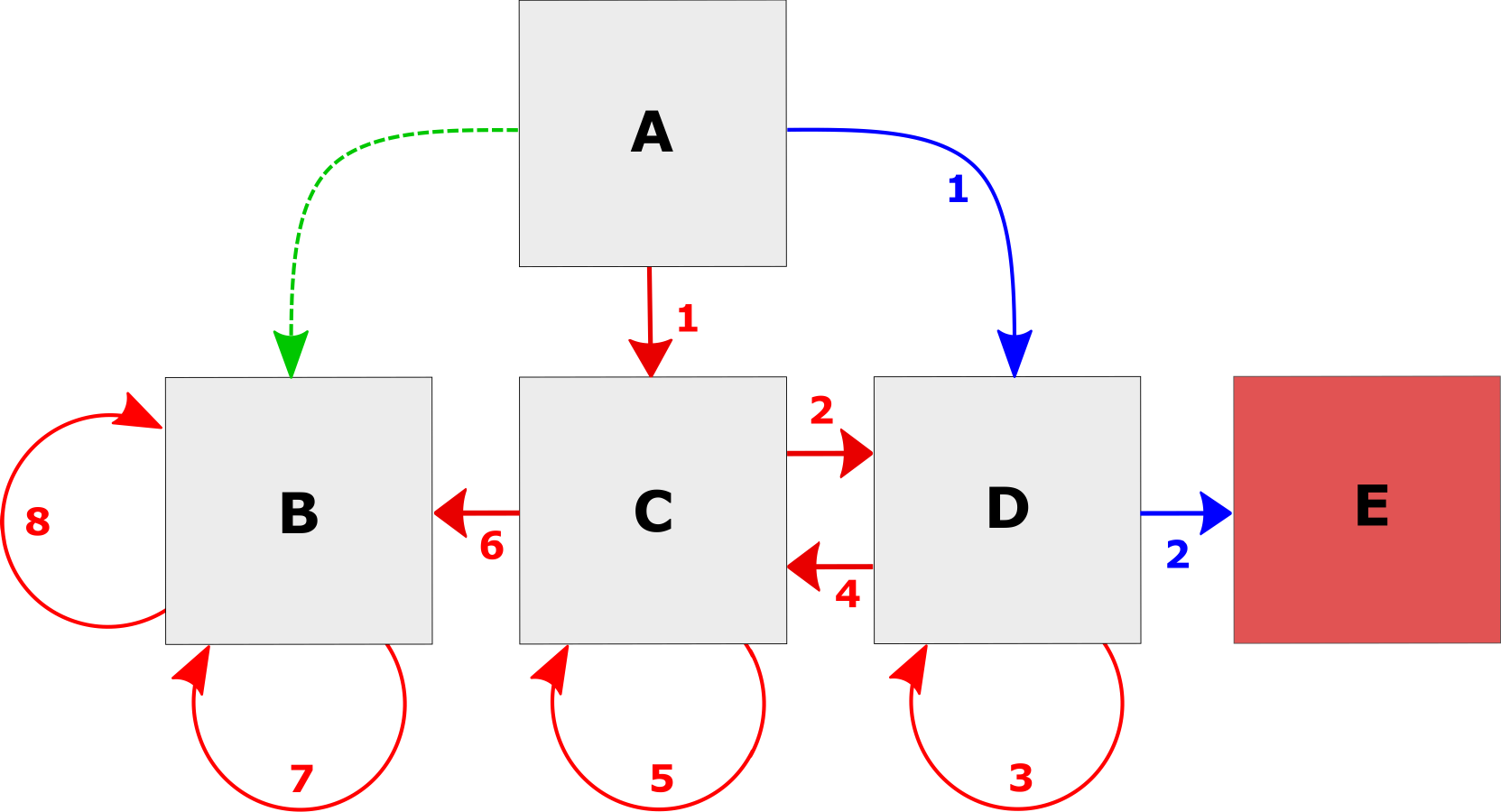}
	\end{center}
	\vspace*{-0.3cm}
	\caption{\label{fig:chain_vv2}Chainworld with uniform count.}
	\vspace*{-0.3cm}
\end{wrapfigure}
To illustrate this issue, consider the chain MDP in Figure \ref{fig:chain_vv2}. The agent always starts in $A$ and can move left, right, and down' The episode ends after eight steps, or when terminal state $E$ is reached. The goal of this example is to explore all states uniformly.
In the first episode, the agent follows the blue trajectory ($A, D, E$). In the second episode, it follows the red trajectory ($A, C, D, D, C, B, B, B$). In both trajectories, an action with count zero was always selected first. At the beginning of the third episode, all state-action pairs have been executed once, except for ``left'' in $A$, i.e., $n(s,a)=1 \:\: \forall (s,a) \neq (A,left), \:\: n(A,left)=0$. Thus, we need to enforce the agent to select it. 
However, the discounted \textit{cumulative} count $\hat n(s,a) = (1 - \gamma_w) W(s,a)$ is \textit{not} uniform. For example with $\gamma_w = 0.9$, $\hat n(A,right) = 3.8$ and $\hat n(A,down) = 4.52$.

\clearpage

More in general, the worst-case scenario we need to consider to satisfy Eq. \eqref{eq:vv2_condition} is the one where (1) action $\bar{a}$ has not been executed yet, (2) an action $\dot{a}$ has been executed once and has the smallest pseudocount, and (3) all other actions have been executed once and have the highest pseudocount. In this scenario, to select $\bar{a}$ we need to guarantee $U_W(s,\bar{a}) > U_W(s,\dot{a})$. Since we assume that counts are uniform and no action has been executed twice, the smallest pseudocount is $\hat n(s,\dot{a}) = 1 - \gamma_w$, while the highest is 1. Then, Eq. \eqref{eq:vv2_condition} becomes
\begin{align}
U_W(s,\bar{a}) &> \frac{Q_{\max} - Q_{\min}}{\coeff} + U_W(s,\dot{a}) \nonumber
\\
U_W(s,\bar{a}) &> \frac{Q_{\max} - Q_{\min}}{\coeff} + \sqrt{\frac{2\log\left((1 - \gamma_w) + |\actionspace| - 2\right)}{1 - \gamma_w}}. \label{eq:vv2_condition2}
\end{align}
Replacing $U_W(s,a)$ in Eq. \eqref{eq:vv2_beta} with the above bound when $n(s,\bar{a}) = 0$ guarantees to select $\bar{a}$.

\begin{figure}[t]
	\centering 
	\raisebox{22pt}{\rotatebox[origin=t]{90}{\textbf{Terminal}}}	\begin{subfigure}[t]{.18\linewidth} 
	\centering 
	\includegraphics[width=\linewidth]{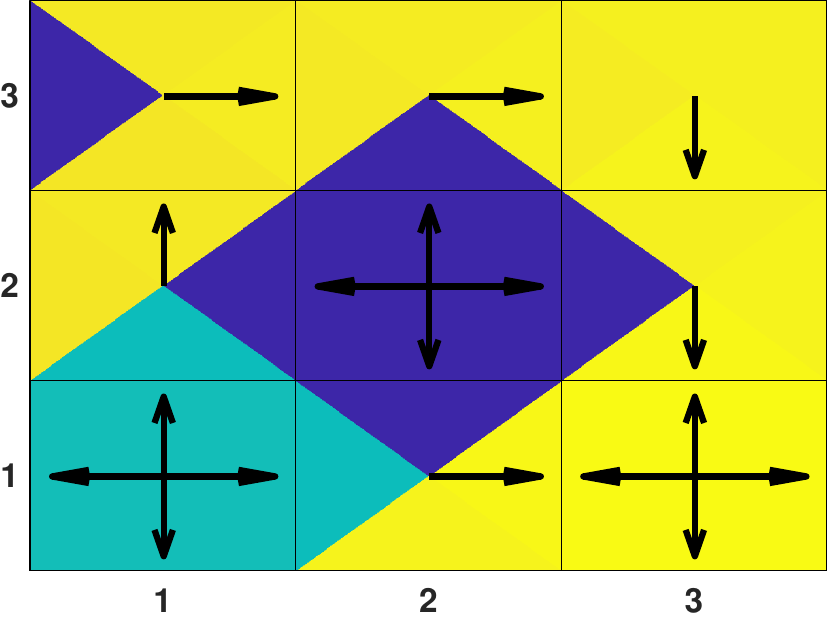}
\end{subfigure}
\hfill
\begin{subfigure}[t]{.18\linewidth} 
	\centering 
	\includegraphics[width=\linewidth]{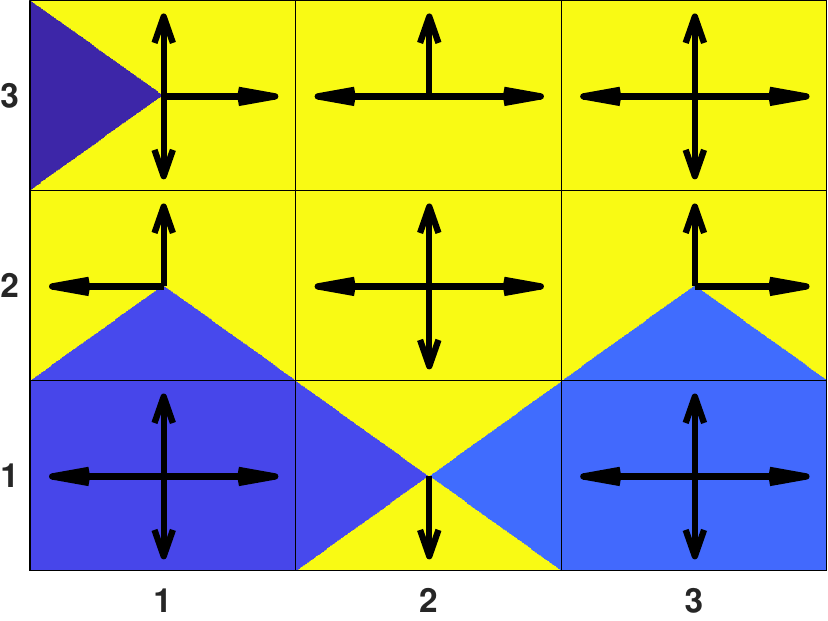}
\end{subfigure}
\hfill
\begin{subfigure}[t]{.18\linewidth} 
	\centering 
	\includegraphics[width=\linewidth]{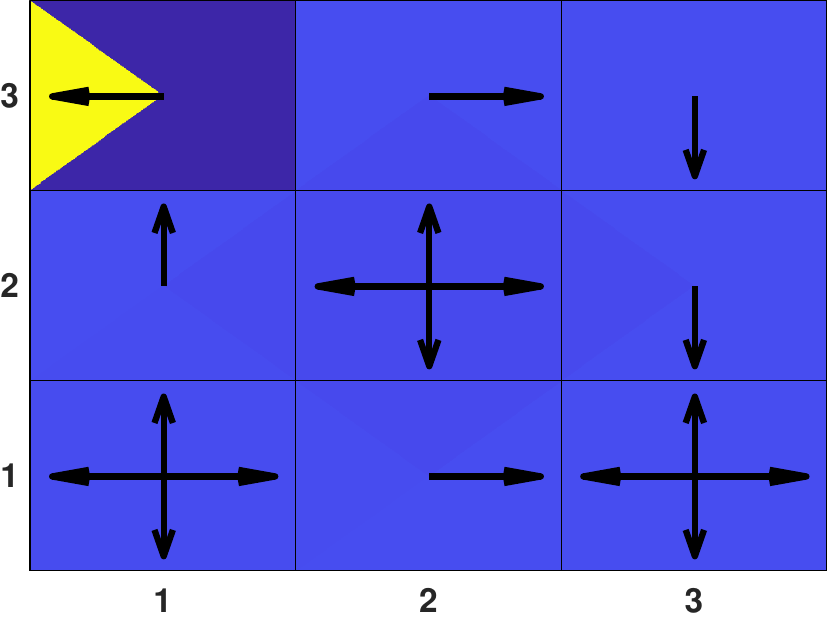}
\end{subfigure} 
\hfill
\begin{subfigure}[t]{.18\linewidth} 
	\centering 
	\includegraphics[width=\linewidth]{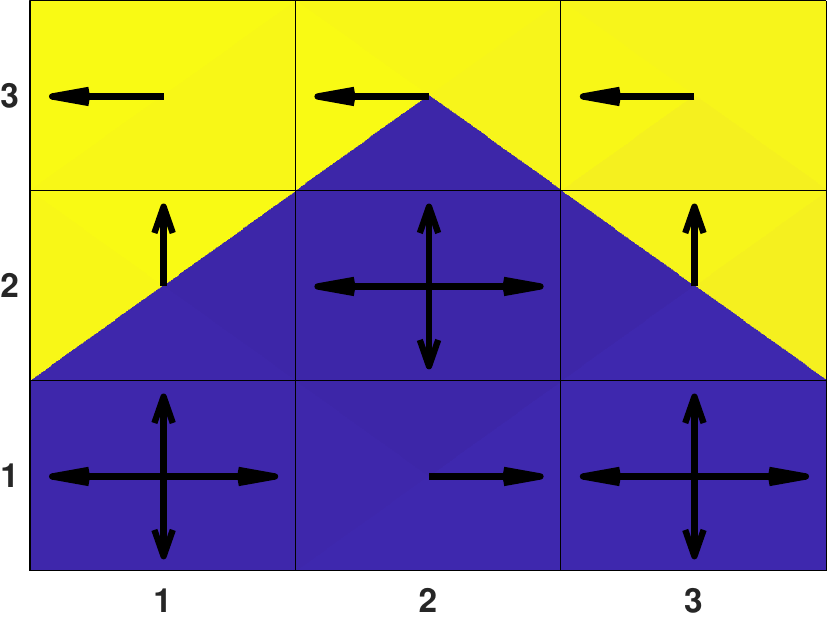}
\end{subfigure}
\hfill
\begin{subfigure}[t]{.18\linewidth} 
	\centering 
	\includegraphics[width=\linewidth]{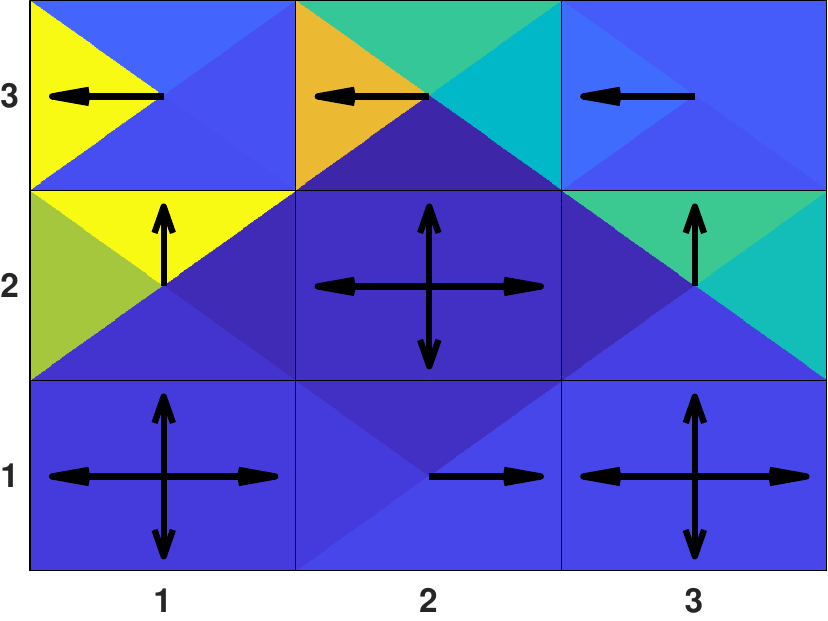}
\end{subfigure}
\\[0.5em]
	\raisebox{22pt}{\rotatebox[origin=t]{90}{\textbf{Non-Term.}}}
	\begin{subfigure}[t]{.18\linewidth} 
		\centering 
		\includegraphics[width=\linewidth]{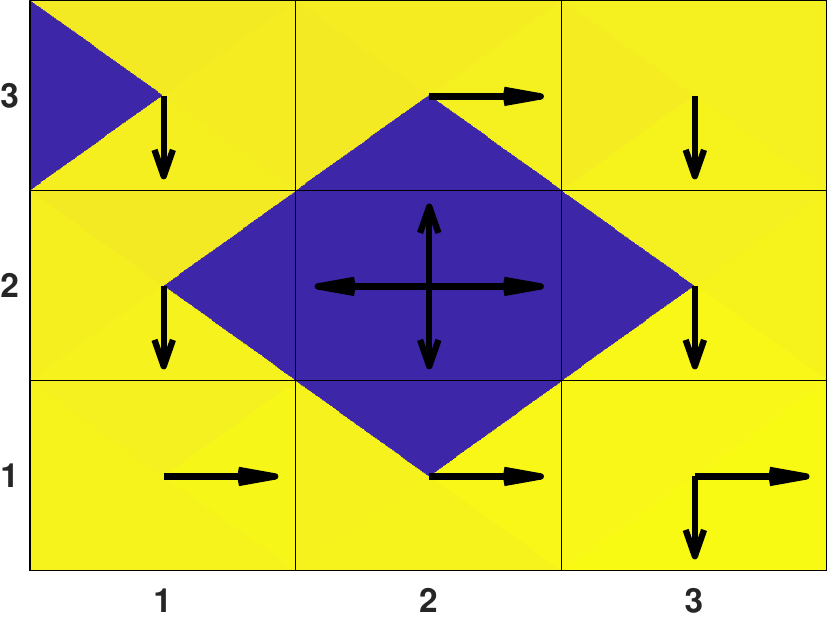}
		\caption{\label{fig:toy_greedy_0}Greedy}
	\end{subfigure}
	\hfill
	\begin{subfigure}[t]{.18\linewidth} 
		\centering 
		\includegraphics[width=\linewidth]{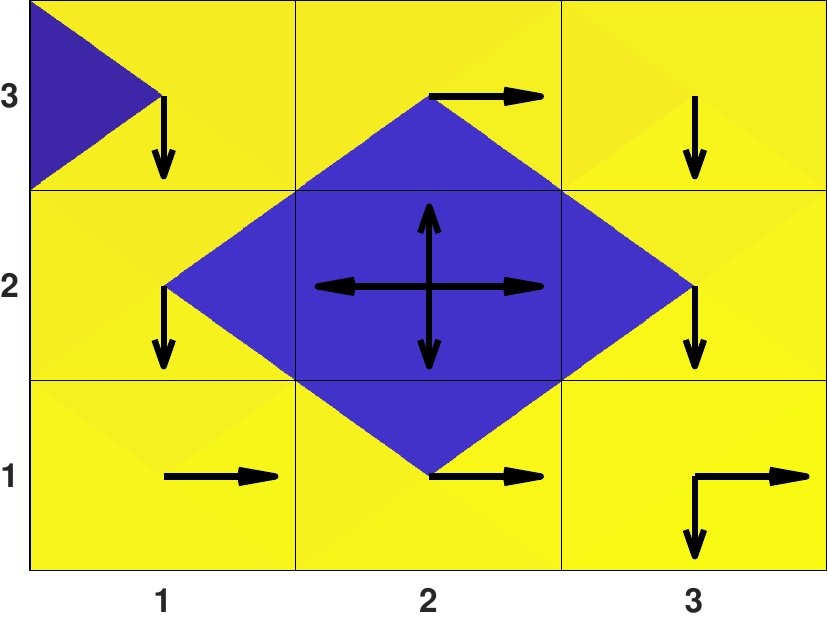}
		\caption{\label{fig:toy_mbie_0}Augmented}
	\end{subfigure}
	\hfill
	\begin{subfigure}[t]{.18\linewidth} 
		\centering 
		\includegraphics[width=\linewidth]{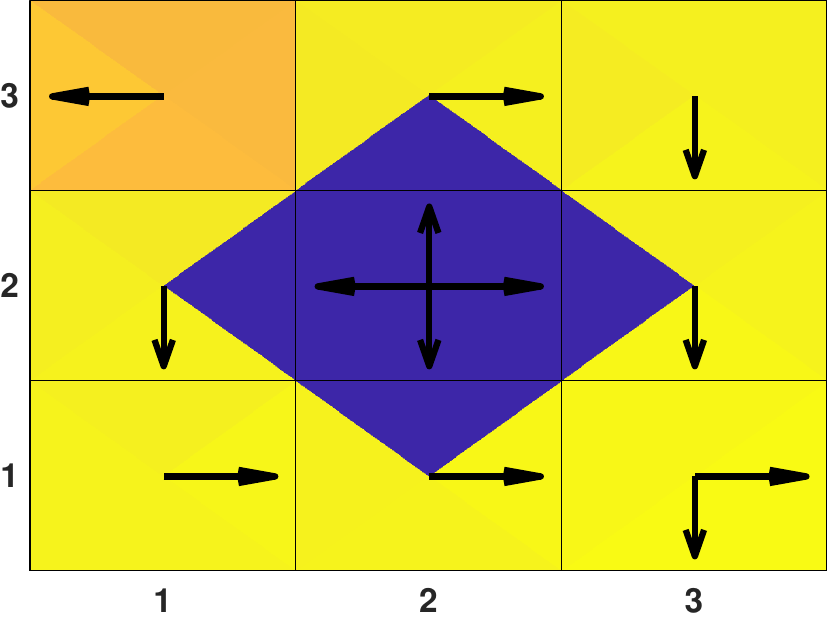}
		\caption{\label{fig:toy_ucb_0}UCB1}
	\end{subfigure} 
	\hfill
	\begin{subfigure}[t]{.18\linewidth} 
		\centering 
		\includegraphics[width=\linewidth]{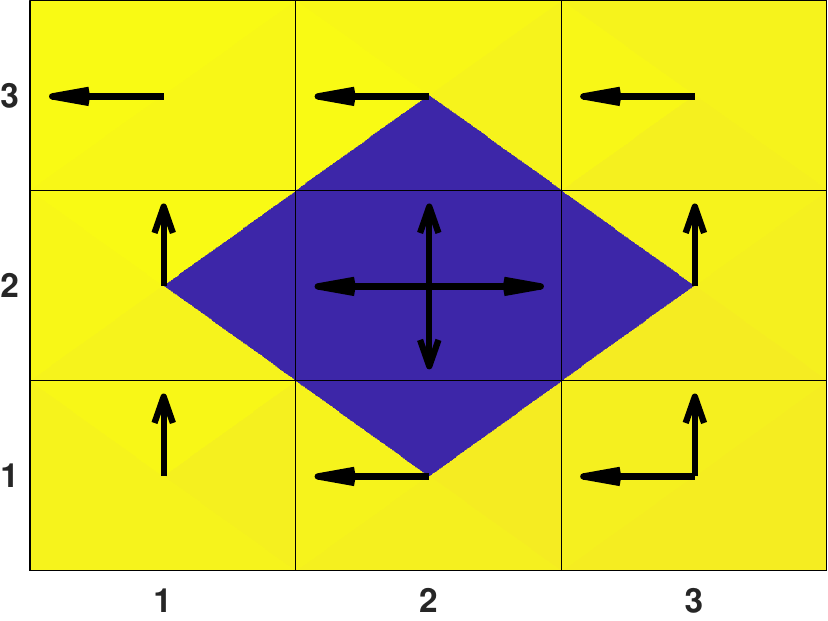}
		\caption{\label{fig:toy_vv3_0}UCB W-func.}
	\end{subfigure}
	\hfill
	\begin{subfigure}[t]{.18\linewidth} 
		\centering 
		\includegraphics[width=\linewidth]{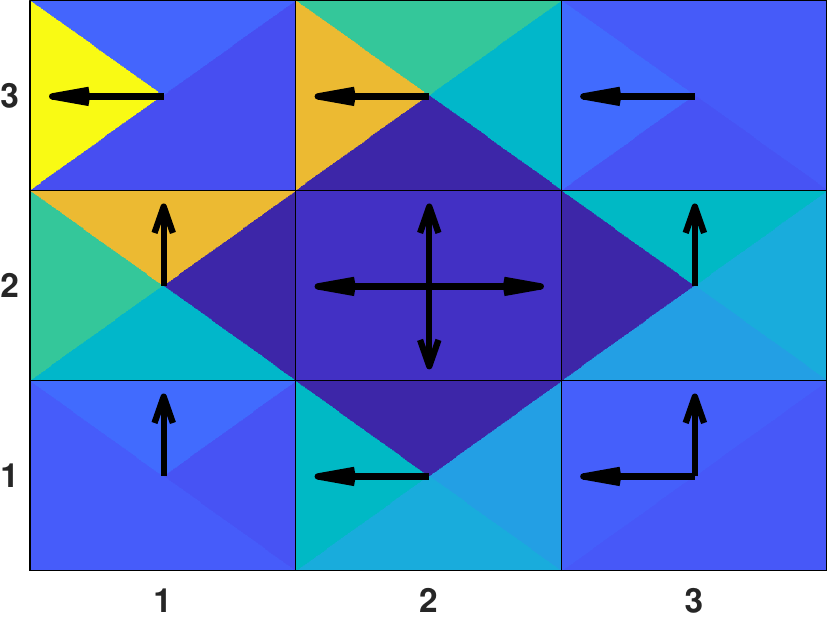}
		\caption{\label{fig:toy_vv2_0}Count W-func.}
	\end{subfigure}
	\caption{\label{fig:toy_uniform}\textbf{Toy example (part 3).} Behavior policies under uniform visitation count. In the top row, reward states (1,1) and (3,1) are terminal. In the bottom row, they are not and the MDP has infinite horizon. All state-action pairs been executed once, except for ``left'' in (1,3). Only the proposed methods (\subref{fig:toy_vv3_0}, \subref{fig:toy_vv2_0}) correctly identify such state-action pair as the most interesting for exploration, and propagate this information to other states thanks to TD learning. When the action is executed, all policies converge to the greedy one (\subref{fig:toy_greedy_0}).}
\end{figure}

\paragraph{Example of exploration behavior under uniform count.}
Consider our toy problem again. This time, all state-action pairs have been executed once except for ``left'' in (1,3). The Q-table is optimal for visited state-action pairs, thus the greedy policy simply brings the agent to (3,1) where the highest reward lies. 
Figure \ref{fig:toy_uniform} shows baseline and proposed behavior policies in two scenarios, depending on whether reward states are terminal (upper row) or not (bottom row). 
\begin{enumerate}[a)]
\setlength{\itemsep}{0pt}
\item The augmented reward policy (Figure \ref{fig:toy_mbie_0}) has no interest in trying ``left'' in (1,3), and its value is the lowest. The agent, in fact, cannot get the visitation bonus without first executing the action. This shows once more that this passive way of exploration is highly inefficient.
\item UCB1 policy (Figure \ref{fig:toy_ucb_0}) acts greedily everywhere except in (1,3), where it correctly selects the unexecuted action. However, since UCB1 considers only the \textit{immediate} visitation, it has no way of propagating this information to other state-action pairs.
\item By contrast, for the proposed W-function policies (Figure \ref{fig:toy_vv3_0} and \ref{fig:toy_vv2_0}) the unexecuted state-action pair is the most interesting, and the policy brings the agent there from any state. Thanks to the separate W-function, we can propagate the value of unexecuted actions to other states, always prioritizing the exploration of new state-action pairs. Under uniform count $n(s,a) = 1 \:\: \forall (s,a)$, then the agent acts greedily w.r.t. the Q-function.
\end{enumerate}

Finally, if the count is uniform for all state-action pairs, including for ``left'' in (1,3), then all policies look like the greedy one (Figure \ref{fig:toy_greedy_0}).

\clearpage

\subsection{Final Remarks}
\label{ssec:notes}
In this section, we discuss the uniform count assumption used to derive the bounds, the differences between the proposed W-functions and intrinsic motivation, the benefit of using the count-based one, and potential issues in stochastic environments and continuous MDPs. 

\paragraph{Uniform count assumption.}
In Section \ref{ssec:vv_init} we derived an initialization for $W^\beta_{\textsub{ucb}}$ and a bound for $W^\beta_{\textsub{n}}$ to guarantee that an action is not executed twice before all other actions are executed once. For that, we assumed the visitation count to be uniform for all state-action pairs. 
We acknowledge that this assumption is not true in practice. First, it may be necessary to revisit old states in order to visit new ones. Second, when an episode is over the state is reset and a new episode starts. 
This is common practice also for infinite horizon MDPs, when usually the environment is reset after some steps.
Thus, the agent's initial state will naturally have a higher visitation count. Nonetheless, in Section \ref{sssec:visit_count_map}, we will empirically show that our approach allows the agent to explore the environment as uniformly as possible.

\paragraph{W-function vs auxiliary rewards.}
Using an auxiliary reward such as an exploration bonus or an intrinsic
motivation signal has similarities with the proposed approach, especially with $W^\beta_{\textsub{ucb}}$, but the two methods are not equivalent.
Consider augmenting the reward with the same visitation reward used to train the W-function, i.e., $r^+_t = r_t + {\alpha}\:r_t^W$. Using TD learning, the augmented Q-function used for exploration can be rewritten as\footnote{For the sake of simplicity, we consider only non-terminal states.}
\begin{align}
Q^+(s_t,a_t) &= \textcolor{red}{r_t + \alpha r_t^W} + \textcolor{blue}{\gamma \max_a Q^+(s_{t+1},a)}.
\intertext{Then consider the behavior policy derived from $W^\beta_{\textsub{ucb}}$ in Eq. \eqref{eq:vv3_beta}, and decompose it as well}
Q^\pi(s_{t},a_t) + \alpha W^\beta_{\textsub{ucb}}(s_{t},a_t) 
&= \textcolor{red}{r_t} + \textcolor{blue}{\gamma \max_a Q^\pi(s_{t+1},a)} + \textcolor{red}{\alpha r_t^W} + \textcolor{blue}{\alpha \gamma_w \max_{a} W^\beta_{\textsub{ucb}}(s_{t+1},a)},
\end{align}
where we considered $\alpha = \coeff(1 - \gamma_w)$. The red terms are equivalent, but the blue terms are not because of the $\max$ operator, since $\max_x\{f(x) + g(x)\} \neq \max_x\{f(x)\} + \max_x\{g(x)\}$.

This explicit separation between exploitation ($Q$) and exploration ($W$) has two important consequences. First, it is easy to implement the proposed behavior policy for any off-policy algorithm, as the exploration term is separate and independent from the Q-function estimates. This implies that we can propagate the exploration value independently from the
action-value function estimate. With this choice we can select a
discount factor $\gamma_w$ which differs from the MDP discount $\gamma$. By choosing a high exploration
discount factor $\gamma_w$, we do long-term exploration allowing us to find the
optimal policy when exploration is crucial. By choosing a low exploration
discount factor $\gamma_w$, instead, we perform short-term exploration which may converge faster to a greedy policy. In the evaluation in Section \ref{ssec:eval_part2}, we show experiments for which the discount factors are the same, as well as experiments where they differ. Furthermore, we investigate the effect of changing the W-function discount while keeping the Q-function discount fixed.
\\
Second, auxiliary rewards are non-stationary, as the visitation count changes at every step. This leads to a non-stationary target for the Q-function. With our approach, by decoupling exploration and exploitation, we have stationary target for the Q-function while moving all the non-stationarity to the W-function.

\paragraph{Terminal states and stochasticity.}
The visitation rewards for training the W-functions penalize terminal state-action pairs (Eq. \eqref{eq:vv3_rwd} and \eqref{eq:vv2_rwd}). Consequently, once visited, the agent will avoid visiting them again. One may think that this would lead to poor exploration in stochastic environments, where the same state-action pair can lead to both terminal and non-terminal states. 
In this scenario, trying the same action again in the future may yield better exploration depending on the stochastic transition function. However, as we empirically show in Section \ref{sssec:stochastic_grid}, stochasticity in the transition function does not harm our approach.

\paragraph{Pseudo-counts and continuous MDPs.}
In the formulation of our method --and later in the experiments section-- we have considered only tabular MDPs, i.e., with discrete states and actions. This makes it easy to exactly count state-action pairs and accurately compute visitation rewards and value functions.
We acknowledge that the use of counts may be limited to tabular MDPs, as real-world problems are often characterized by continuous states and actions. Nonetheless, pseudo-counts~\citep{tang2017exploration} have shown competitive performance in continuous MDPs and can be used in place of exact counts without any change to our method. Alternatively, it is possible to replace counts with density models~\citep{ostrovski2017count} with little modifications to our method. Finally, if neural networks would be used to learn the visitation-value function $W(s,a,s')$, it may be necessary to introduce target networks similarly to deep Q-networks~\citep{mnih2013playing}.

\section{Evaluation}
\label{sec:eval}
In this section, we evaluate the proposed method against state-of-the-art methods for exploration in model-free RL on several domains. 
The experiments are designed to highlight the challenges of learning with sparse rewards, and are split into two parts. In the first, we present the environments and we address (1) learning when only few states give a reward, (2) learning when a sequence of correct actions is needed to receive a reward, (3) exploring efficiently when few steps are allowed per training episode, (4) avoiding local optima and distracting rewards, and (5) avoiding dangerous states that stop the agent preventing exploration. 
In the second, we further investigate (6) the visitation count at convergence, (7) the empirical sample complexity, (8) the impact of the visitation value hyperparameter $\gamma_w$, (9) the performance in the infinite horizon scenario, and (10) with stochastic transition function.
\\
To ensure reproducibility, the experiments are performed over fixed random seeds. For deterministic MDPs, the seeds are $1, 2, \ldots, 20$. For stochastic MDPs, the seeds are $1, 2, \ldots, 50$.
Pseudocode and further details of the hyperparameters are given in Appendix \ref{app:tab_details}.
The source code is available at \url{https://github.com/sparisi/visit-value-explore}

\paragraph{Baseline Algorithms.} The evaluated algorithms all use Q-learning~\citep{watkins1992q} and share the same hyperparameters, e.g., learning rate and discount factor. In most of the experiments, we use infinite replay memory as presented by \citet{osband2019deep}.
In Appendix \ref{app:dyna_vs_q} we also present an evaluation without replay memory on some domains.
The algorithms all learn two separate Q-tables, a behavior one for exploration, and a target one for testing the quality of the greedy policy. 
The difference among the algorithms is how the behavior policy performs exploration. In this regard, we compare against the following exploration strategies: random, $\epsilon$-greedy, bootstrapped, count-based.
\\
For bootstrapping, we evaluate the original version proposed by~\citet{osband2016deep}, as well as the more recent versions of~\citet{deramo2019exploiting} and \citet{osband2019deep}.
These methods all share the same core idea, i.e., to use an ensemble of Q-tables, each initialized differently and trained on different samples, to approximate a distribution over Q-values via bootstrapping. 
The differences are the following. \citet{osband2016deep} select a random Q-table at the beginning of each episodes, and use it until the episode is over. \citet{osband2019deep} further regularize the Q-tables using the squared $\ell_2$-norm to ``prior tables'' representing a prior over the Q-function.
\citet{deramo2019exploiting} instead select the Q-table randomly within the ensemble at each step to approximate Thompson sampling, but do not use any prior.
\\
For count-based algorithms, we compare against UCB1 exploration~\citep{auer2002finite} on the immediate state-action count as in Eq.~\eqref{eq:ucb1}, and against augmenting the reward with the exploration bonus proposed by~\citet{strehl2008analysis} as in Eq.~\eqref{eq:exp_bonus}. Notice that~\citet{strehl2008analysis} apply the exploration bonus to a model-based algorithm, but the same bonus has since then been used by many model-free algorithms~\citep{bellemare2016unifying,dong2020q}.
More recent methods use more sophisticated bonus to derive strong convergence guarantees \citep{jin2018iq,dong2020q}.
However, for large-size problems these methods may require an impractical number of samples and are often outperformed by standard algorithms.

\clearpage

\subsection{Part 1: Discounted Return and States Discovered}
\label{ssec:eval_part1}
We use the following environments: deep sea~\citep{osband2018randomized,osband2019deep}, taxi driver~\citep{asadi2017alternative}, and four novel benchmark gridworlds.
All the environments have sparse discounted rewards, and each has peculiar characteristics making it challenging to solve.

\paragraph{Evaluation Criteria.} For each environment, we evaluate the algorithms on varying two different conditions: optimistic vs zero initialization of the behavior Q-tables, and short vs long horizon, for a total of four scenarios. 
For each of them, we report the expected discounted return of $\pi(a|s)$, which is greedy over $Q^\pi(s,a)$, and the number of visited states over training samples. In all environments, the initial state is fixed.

Optimistic initialization \citep{lai1985asymptotically} consists of initializing the Q-tables to a large value. After visiting a state-action pair, either the agent experiences a high reward confirming his belief about its quality, or the agent experiences a low reward and learns that the action was not optimal and will not execute it again in that state. By initializing the Q-tables to the upper bound $r_{\max} / (1 - \gamma)$ we are guaranteed to explore all the state-action pairs even with just an $\epsilon$-greedy policy. However, this requires prior knowledge of the reward bounds, and in the case of continuous states and function approximation the notion of a universally ``optimistic prior'' does not carry over from the tabular setting \citep{osband2019deep}. Thus, it is important that an algorithm is able to explore and learn even with a simple zero or random initialization.

The length of the horizon is also an important factor for exploration. If the agent can perform a limited number of actions during an episode, it has to carefully choose them in order to explore as efficiently as possible. Prominent examples in this regard are real-world tasks such as videogames, where the player has limited time to fulfill the objective before the game is over, or robotics, where the autonomy of the robot is limited by the hardware. Thus, it is important that an algorithm is able to explore and learn even when the training episode horizon is short. In the ``short horizon'' scenario, we allow the agent to explore for a number of steps slightly higher than the ones needed to find the highest reward. For example, if the highest reward is eight steps away from the initial position, the ``short horizon'' is ten steps. The ``long horizon'' scenario is instead always twice as long.

Results are first presented in plots showing the number of states discovered and discounted return against training steps averaged over 20 seeds. 
Due to the large number of evaluated algorithms, confidence intervals are not reported in plots. At the end of the section, a recap table summarizes the results and reports the success of an algorithm, i.e., the number of seeds the algorithm learned the optimal policy, and the percentage of states visited at the end of the learning, both with 95\% confidence interval.

\subsubsection{Deep Sea}
\label{sssec:deep_sea}
\begin{wrapfigure}{l}{0.42\textwidth}
	\begin{center}
		\vspace*{-0.7cm}
		\includegraphics[width=0.94\linewidth]{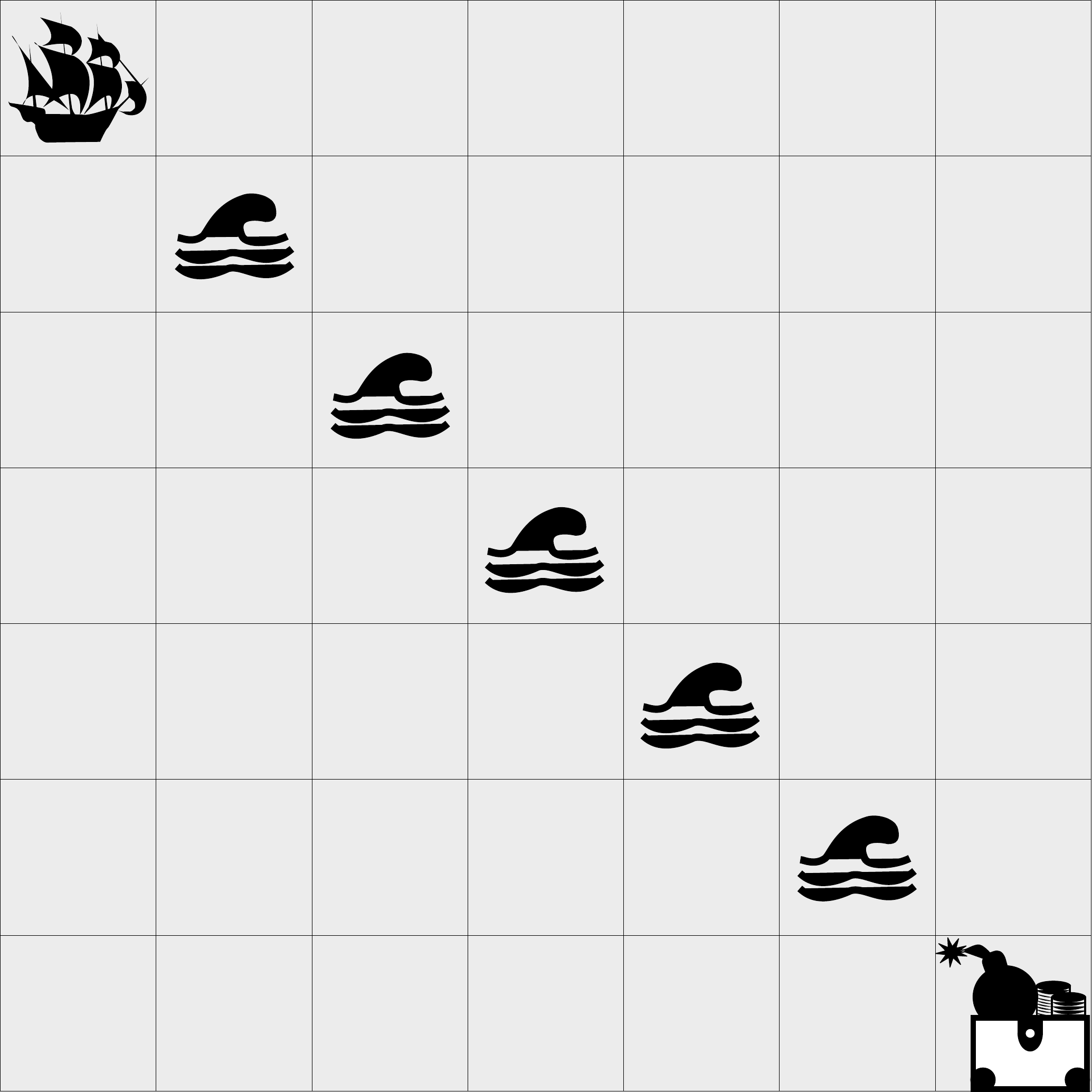}
	\end{center}
	\vspace*{-0.3cm}
	\caption{\label{fig:deep_sea_env}The deep sea.}
	\vspace*{-0.3cm}
\end{wrapfigure}
In this environment (Figure \ref{fig:deep_sea_env}), the agent (the ship) always starts at the top-left corner of a gridworld of size $N\!\times\!N$, and has to descend through the grid to a chest in the bottom-right corner. At the beginning of the episode, either a treasure or a bomb appear inside the chest, with a 50-50. The outcome is known to the agent, as the state consists of the ship position and a flag denoting if the bomb or the treasure has appeared. At every step, the ship descends by one step and must choose to turn left or right. If it hits the left boundary, it will descend without turning. For example, from the starting position $(1,1)$, doing ``left'' will move the ship to $(2,1)$, while doing ``right'' will move it to $(2,2)$. The episode ends when the ship reaches the bottom of the grid, and thus the horizon is fixed to $N$.
The reward is 1 if the ship finds the treasure, -1 if it finds the bomb, and $-0.01/N$ if the ship goes ``right'' in any cell along the diagonal (denoted by sea waves). 
The optimal policy selects always ``right'' if there is the treasure, otherwise ``left'' at the initial state and then can act randomly.

The challenge of this environment lies in having to select the same action (``right'') $N$ times in a row in order to receive either 1 or -1. Doing ``left'' even only once prevents the ship from reaching the chest. At the same time, the agent is discouraged from doing ``right'', since doing so along the diagonal yields an immediate penalty. 
Consequently, an $\epsilon$-greedy policy without optimistic initialization would perform poorly, as the only short-term reward it receives would discourage it from doing ``right''. A random policy is also highly unlikely to reach the chest. In particular, the probability such a policy reaches it in any given episode is $(1/2)^N$. Thus, the expected number of episodes before observing either the bomb or the treasure is $2^N$. Even for a moderate value of $N = 50$, this is an impractical number of episodes \citep{osband2019deep}.

\begin{figure}[t]
	\centering
	\includegraphics[width=0.95\linewidth,trim=0 1.1em 16em 0, clip]{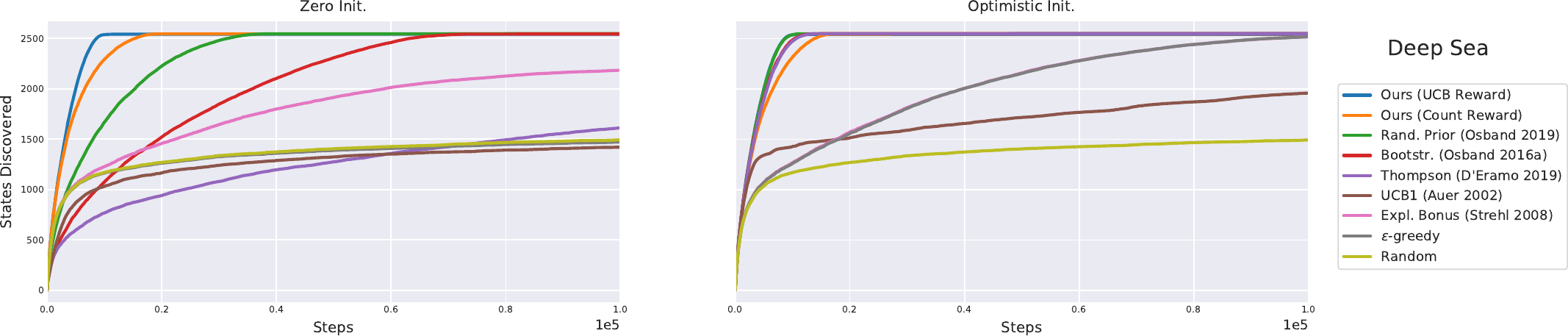}
	\\[0.5em]
	\includegraphics[width=0.95\linewidth,trim=0 0 16em 1em, clip]{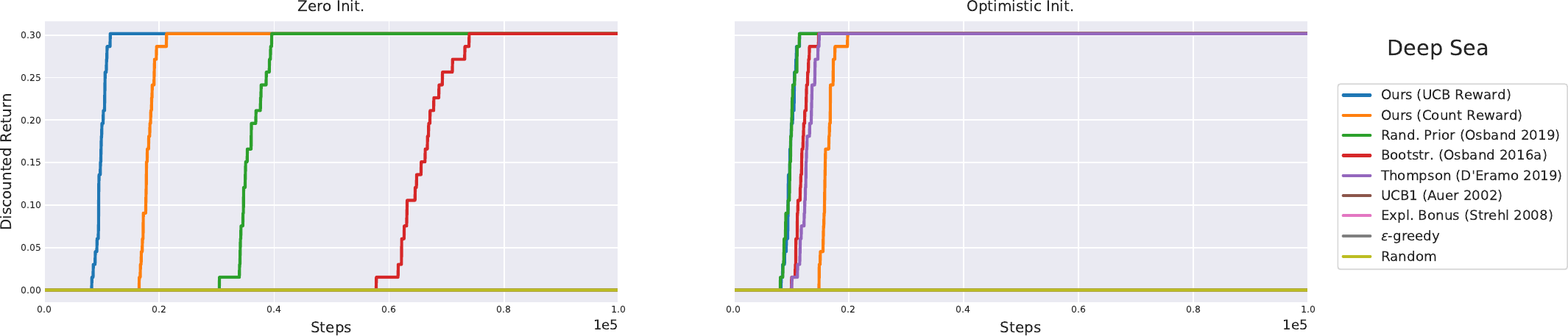}
	\\[0.5em]
	\includegraphics[width=\linewidth]{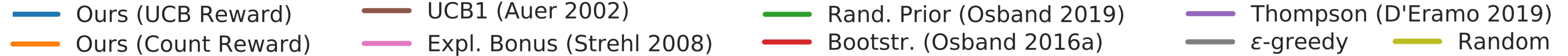}
	\caption{\label{fig:deepsea_res}\textbf{Results on the deep sea domain} averaged over 20 seeds.
	The proposed exploration achieves the best results, followed by bootstrapped exploration.}
\end{figure}

\paragraph{Results.}
Figure \ref{fig:deepsea_res} shows the results on a grid of depth $N = 50$. Since this domain has a fixed horizon, the evaluation does not include the ``short vs long horizon'' comparison.
With zero initialization, only our algorithms and bootstrapping by \citet{osband2016deep} and \citet{osband2019deep} discover all states and converge to the optimal policy within steps limit. Both proposed visitation-count-based (blue and orange) outperform the other two, converging twice or more as faster. Bootstrapped exploration with random prior (green) comes second, reproducing the results reported by \citet{osband2018randomized} and \citet{osband2019deep}.
Results with optimistic initialization are similar, with the main difference being that also approximate Thompson sampling by \citet{deramo2019exploiting} (purple) converges. Bootstrapping by \citet{osband2018randomized} (green) matches the performance of visitation-count-based with UCB (blue), and the two lines almost overlap.
For more details about the percentage of successful trials and the percentage of states discovered by each algorithm with confidence intervals, we refer to Table \ref{tab:recap}.

This evaluation shows that the proposed exploration outperforms state-of-the-art bootstrapping on a domain with a fixed horizon and no local optima. In the next sections, we investigate domains with absorbing states, i.e., that can prematurely end the episode, and distractor reward.
In Section \ref{sssec:empirical_sample_deep}, we also present an empirical study on sample complexity on varying the deep sea depth $N$. Furthermore, in Section \ref{sssec:visit_count_map}, we show the state visitation count at the end of the learning for the deep sea and other domains, discussing why the proposed visitation-value-based exploration achieves new state-of-the-art results.

\clearpage

\subsubsection{Multi-Passenger Taxi Driver}
\label{sssec:taxi}
\begin{wrapfigure}{l}{0.28\textwidth}
	\begin{center}
		\vspace*{-0.7cm}
		\includegraphics[width=0.93\linewidth]{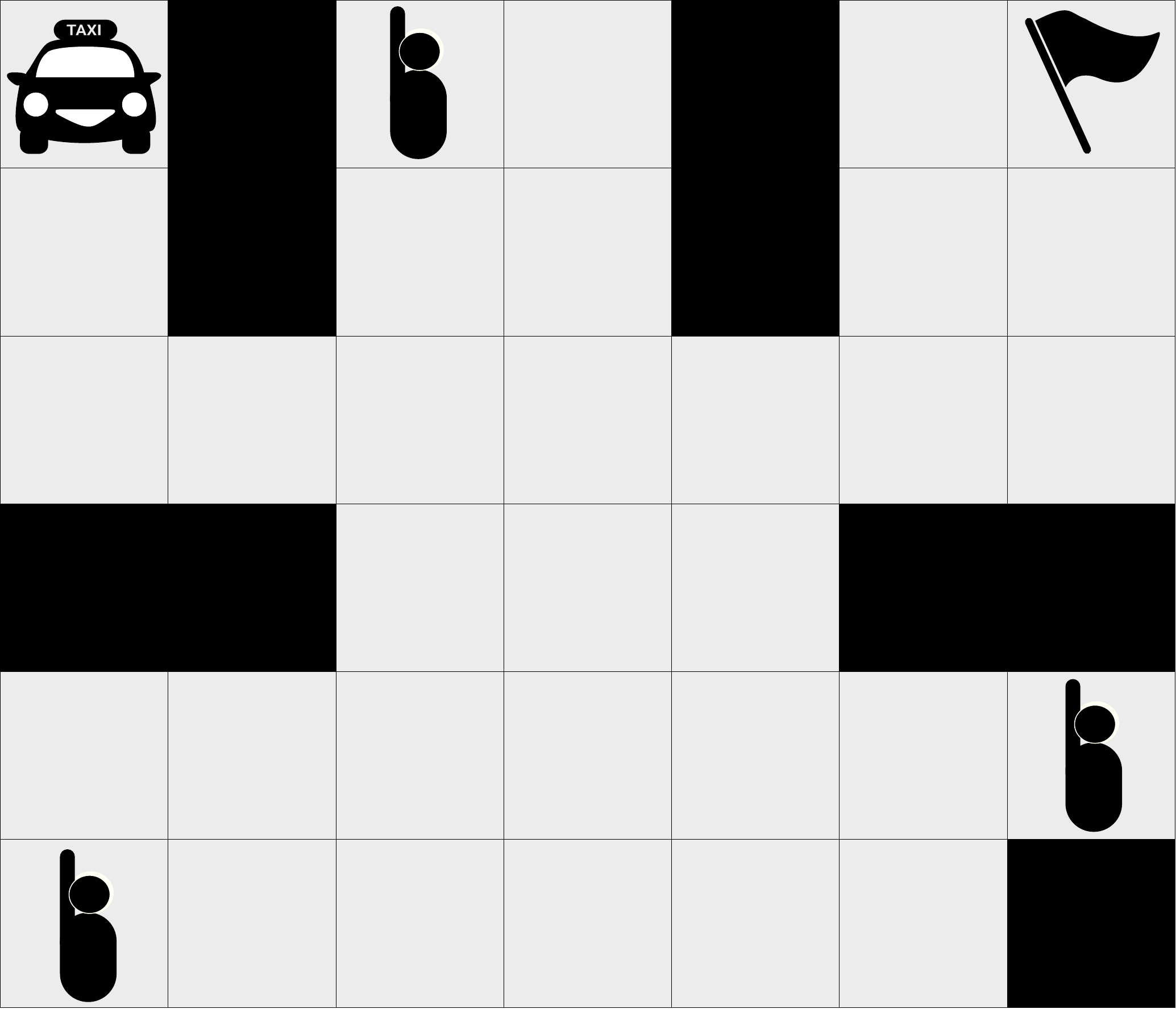}
	\end{center}
	\vspace*{-0.5cm}
	\caption{The taxi driver.}
	\label{fig:taxi_env}
	\vspace*{-0.4cm}
\end{wrapfigure}In this environment (Figure \ref{fig:taxi_env}), the agent (the taxi) starts at the top-left corner of a grid and has to pick up three passengers and drop them off at a goal (the flag). It can carry multiple passengers at the same time. If one, two, or all three passengers reach the goal, the agent is rewarded with 1, 3, or 15, respectively. Otherwise the reward is 0. 
The state consists of the taxi position and three booleans denoting if a passenger is in the taxi. The agent can move left, right, up, or down. Black cells cannot be visited. 
An episode ends if the taxi goes to the flag, even without passengers.
The optimal policy picks up all passengers and drops them off in 29 steps\footnote{28 steps to pick all passengers and reach the goal, plus an additional action to get the reward.}.

This domain is challenging for two reasons. First, it admits many locally optimal policies, depending on how many passengers reach the goal. Second, with a short horizon learning to pick up and drop off all passengers is difficult and requires effective exploration.

\begin{figure}[t]
	\centering
	\includegraphics[width=0.95\linewidth,trim=0 1.1em 15.5em 0em, clip]{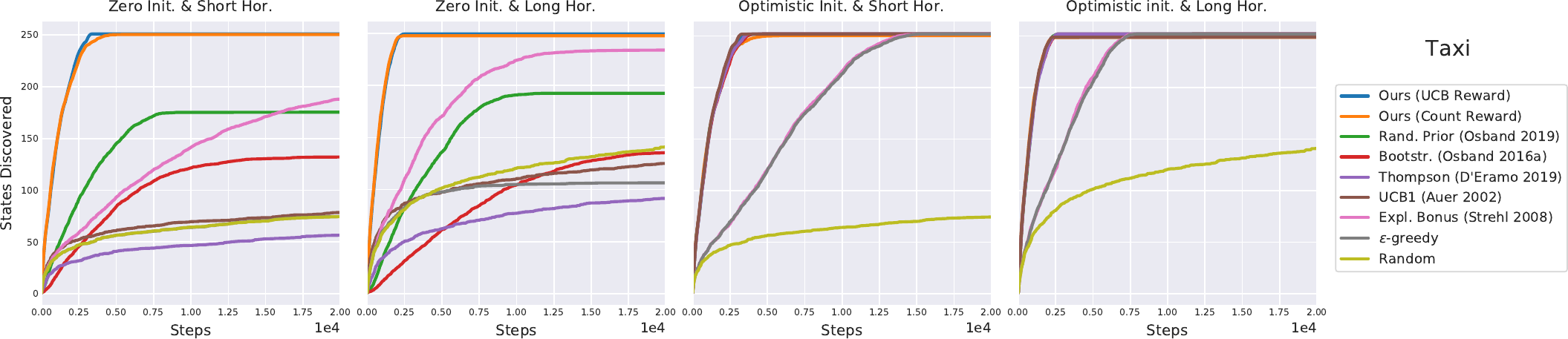}
	\\[0.5em]
	\includegraphics[width=0.95\linewidth,trim=0 0 15.5em 1em, clip]{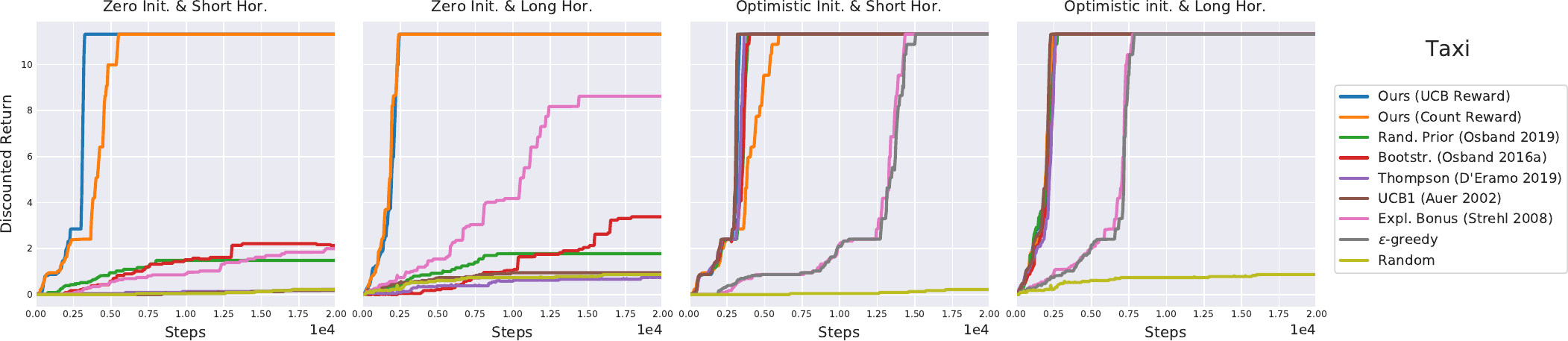}
	\\[0.5em]
	\includegraphics[width=\linewidth]{plot/legend_hor}
	\caption{\label{fig:taxi_res}\textbf{Results on the taxi} domain averaged over 20 seeds. The proposed algorithms outperform all others, being the only solving the task without optimistic initialization.}
\end{figure}

\paragraph{Results.}
In this domain, the ``short horizon'' consists of 33 steps\footnote{Our algorithms could learn also with a horizon of only 29 steps. However, with zero initialization other algorithms performed extremely poorly. We thus decided to give the agent 33 steps.}, while the ``long'' of 66.
As shown in Figure \ref{fig:taxi_res}, bootstrapped algorithms perform significantly worse than before. None of them, in fact, learned to pick all passengers if Q-tables are initialized to zero, and the ``long horizon'' does not help. In particular, random prior bootstrapping (green) converged prematurely due to the presence of local optima (the algorithm does not discover new states after 750-1,000 steps).
The proposed algorithms (blue and orange), instead, perform extremely well, quickly discovering all states and then learning the optimal policy. Other algorithms learned to pick one or two passengers, and only the auxiliary visitation bonus (pink) learned to pick all passengers in some seeds but only with a long horizon. 
With optimistic initialization, most of the algorithms match the performance of our proposed ones. Unsurprisingly, $\epsilon$-greedy (gray) learns slowly.
Bonus-based exploration (pink) is also slow, either because the small bonus coefficient (see Eq.~\eqref{eq:exp_bonus}), or due to the issues discussed in Section \ref{ssec:toy_example}.
Only random exploration (light green) still cannot pick all passengers. For the percentage of successful trials and the percentage of states discovered by each algorithm with confidence intervals, we refer to Table \ref{tab:recap}.

This evaluation shows that the proposed exploration outperforms existing algorithms in the presence of local optima and that its performance is not affected by the length of the horizon. Next, we investigate what happens when we combine the challenges of the deep sea and the taxi.

\subsubsection{Deep Gridworld}
\label{sssec:deep_grid}
\begin{wrapfigure}{l}{0.34\textwidth}
	\begin{center}
		\vspace*{-0.7cm}
		\includegraphics[width=.97\linewidth]{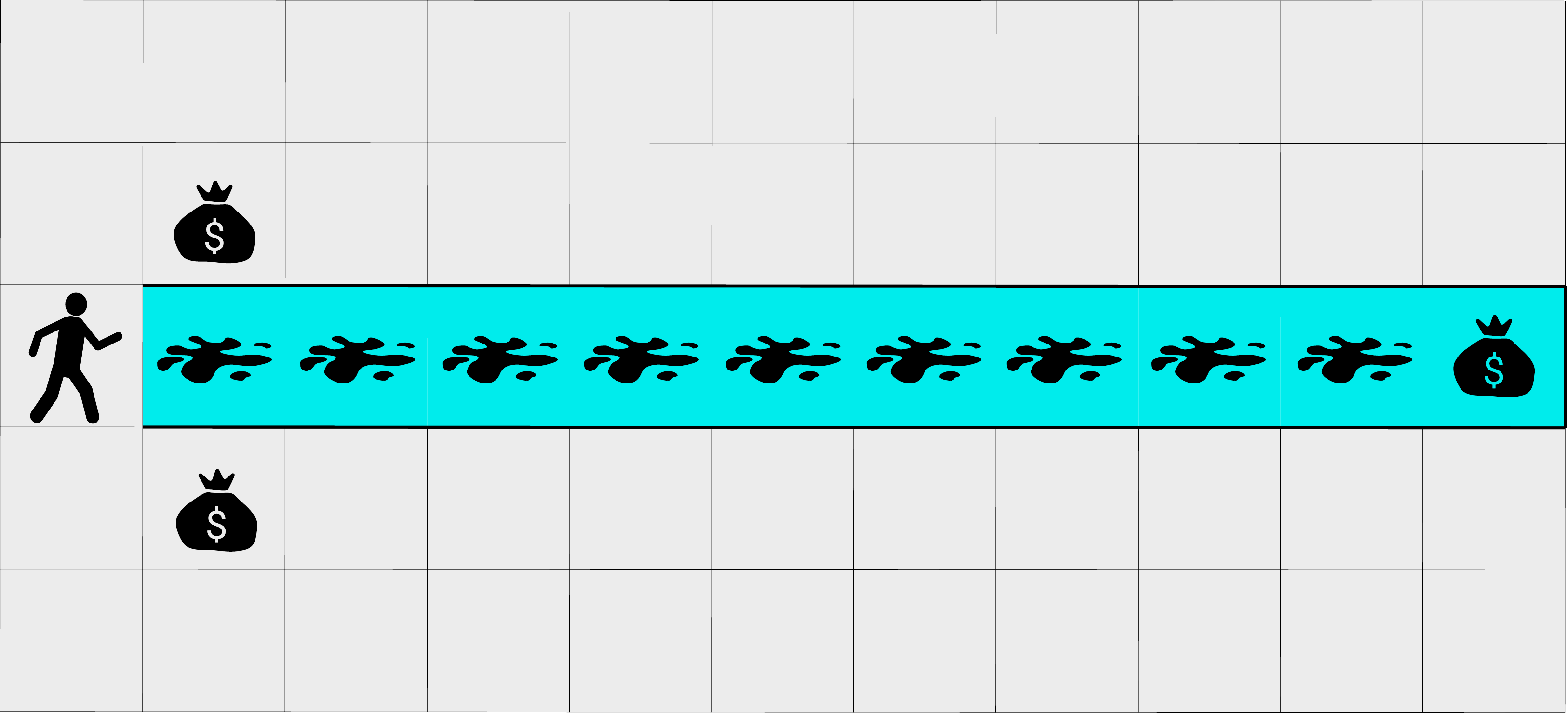}
	\end{center}
	\vspace*{-0.5cm}
	\caption{The deep gridworld.}
	\label{fig:deep_grid_env}
	\vspace*{-0.4cm}
\end{wrapfigure}
The first gridworld we propose is inspired by the deep sea domain. 
In this environment (Figure \ref{fig:deep_grid_env}), the agent navigates in a $5\times11$ grid by moving left, right, up, or down. All states can be visited, but the blue corridor can be entered only from its left side. The agent can exit the corridor anytime by moving either ``up'' or ``down''. A treasure of value 2 lies at the end of the corridor, while two treasures of value 1 serve as distractors next to the corridor entrance. The corridor is filled with puddles giving a penalty of -0.01. The agent always starts next to the entrance and the episode ends when the agent collects any of the treasures. The optimal policy consists of going always ``right''.

The deep gridworld combines the challenges of both the deep sea and the taxi domains. Like the former, to discover the highest reward the agent needs to execute the action ``right'' multiple times in a row, receiving negative immediate rewards due to the puddles. However, doing a different action does not prevent reaching the end of the corridor, because the agent can still go back to the entrance and try again within the same episode. At the same time, the presence of the two distractor treasures results in local optima, as in the taxi driver domain.

\begin{figure}[t]
	\centering
	\includegraphics[width=0.95\linewidth,trim=0 1.1em 15.5em 0em, clip]{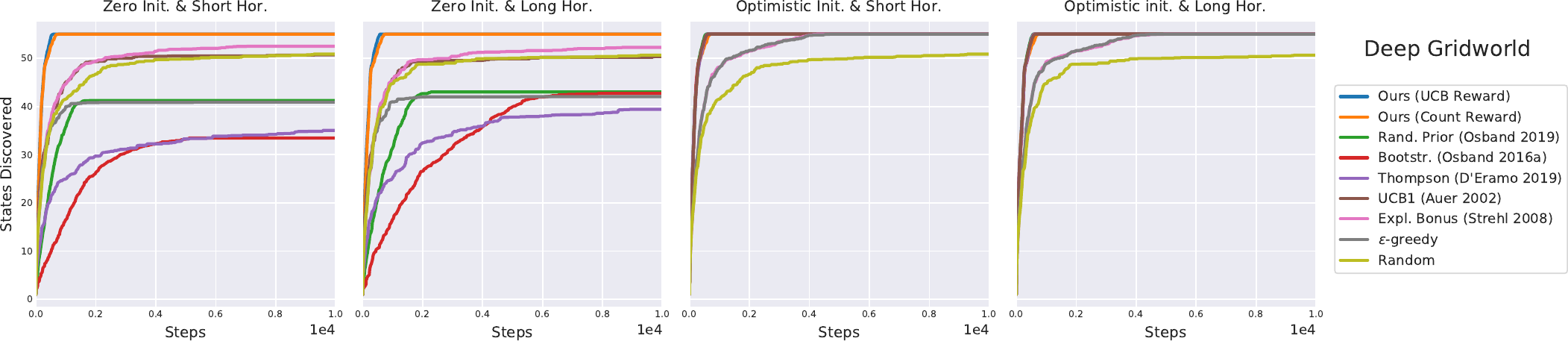}
	\\[0.5em]
	\includegraphics[width=0.95\linewidth,trim=0 0 15.5em 1em, clip]{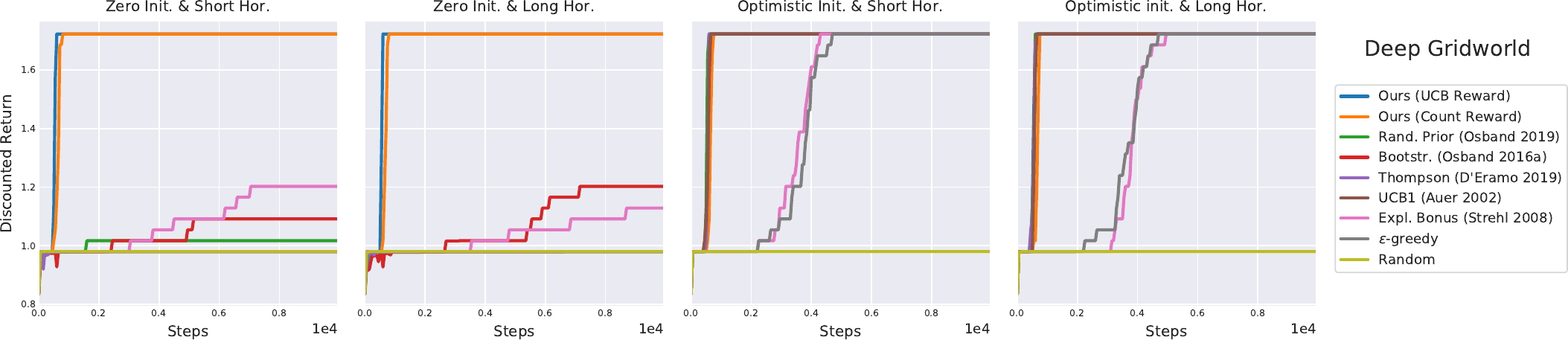}
	\\[0.5em]
	\includegraphics[width=\linewidth]{plot/legend_hor}
	\caption{\label{fig:deep_grid_res}\textbf{Results on the deep gridworld} averaged over 20 seeds. Once again, the proposed algorithms are the only solving the task without optimistic initialization.}
\end{figure}

\paragraph{Results.}
Because it is possible to exit the corridor without ending the episode, we set a longer horizon compared to the previous domains, i.e., 55 steps for the ``short'' scenario and 110 for the ``long'' one. Results shown in Figure \ref{fig:deep_grid_res} confirm previous results. The proposed exploration quickly discovers all states and learns to navigate through the corridor. By contrast, other algorithms, including bootstrapping, get stuck in local optima and learn to collect one of the two lesser treasures. Similarly to the taxi domain, with optimistic initialization all algorithms --but random exploration (light green)-- learn to navigate through the corridor. 

This evaluation shows that distractor rewards are challenging for existing algorithms but not for ours. Next, we increase the difficulty by adding more distractors and new special states.

\clearpage

\begin{figure}[h]
	\centering
	\textbf{The Gridworlds Environments}
	\\[1em]
	\begin{minipage}[t]{.45\textwidth}
		\centering
		\includegraphics[width=\linewidth]{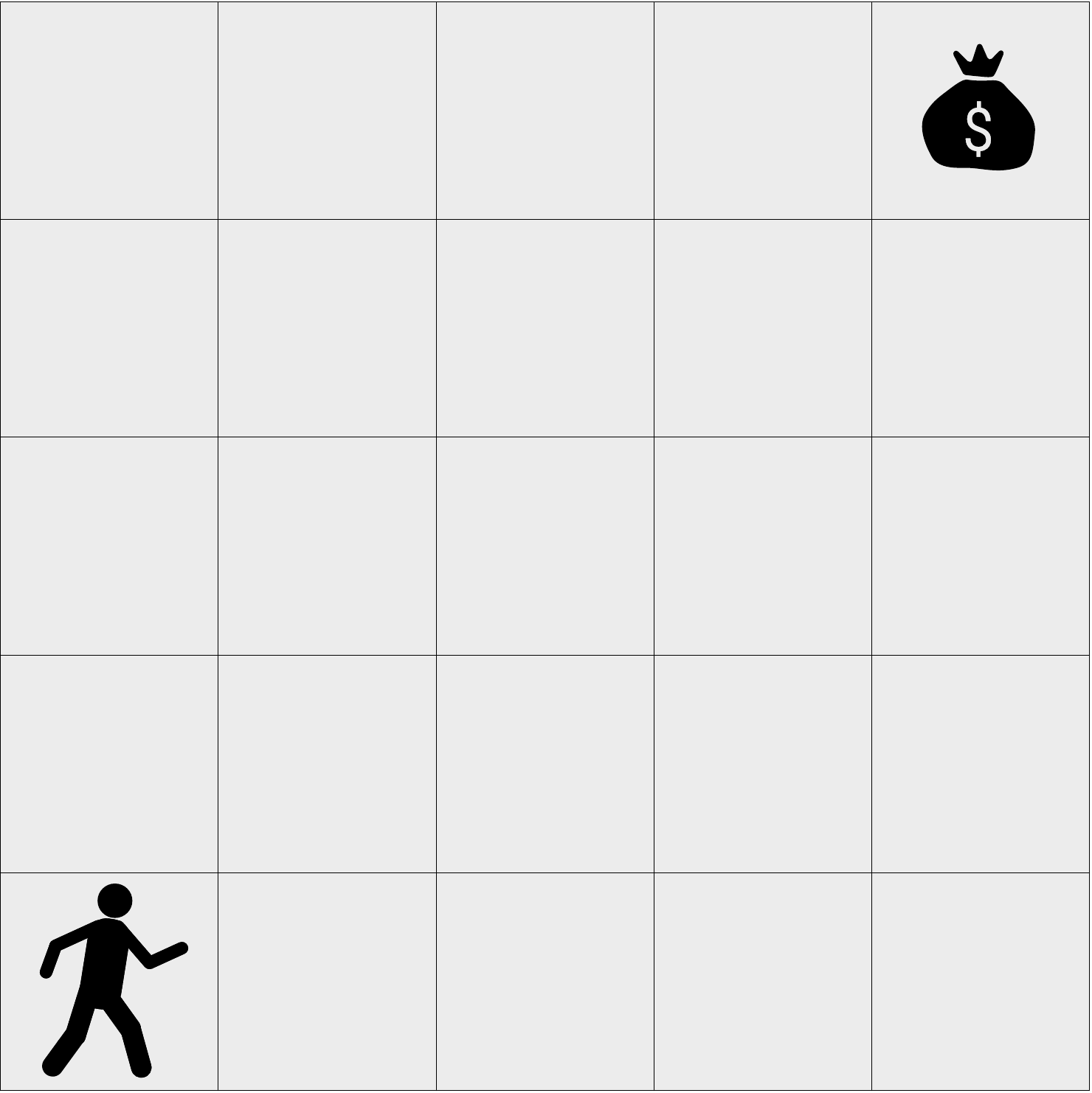}
		\caption{\label{fig:grid_simple_env}The ``toy'' gridworld.}
	\end{minipage}\hfill
	\begin{minipage}[t]{.45\textwidth}
		\centering
		\includegraphics[width=\linewidth]{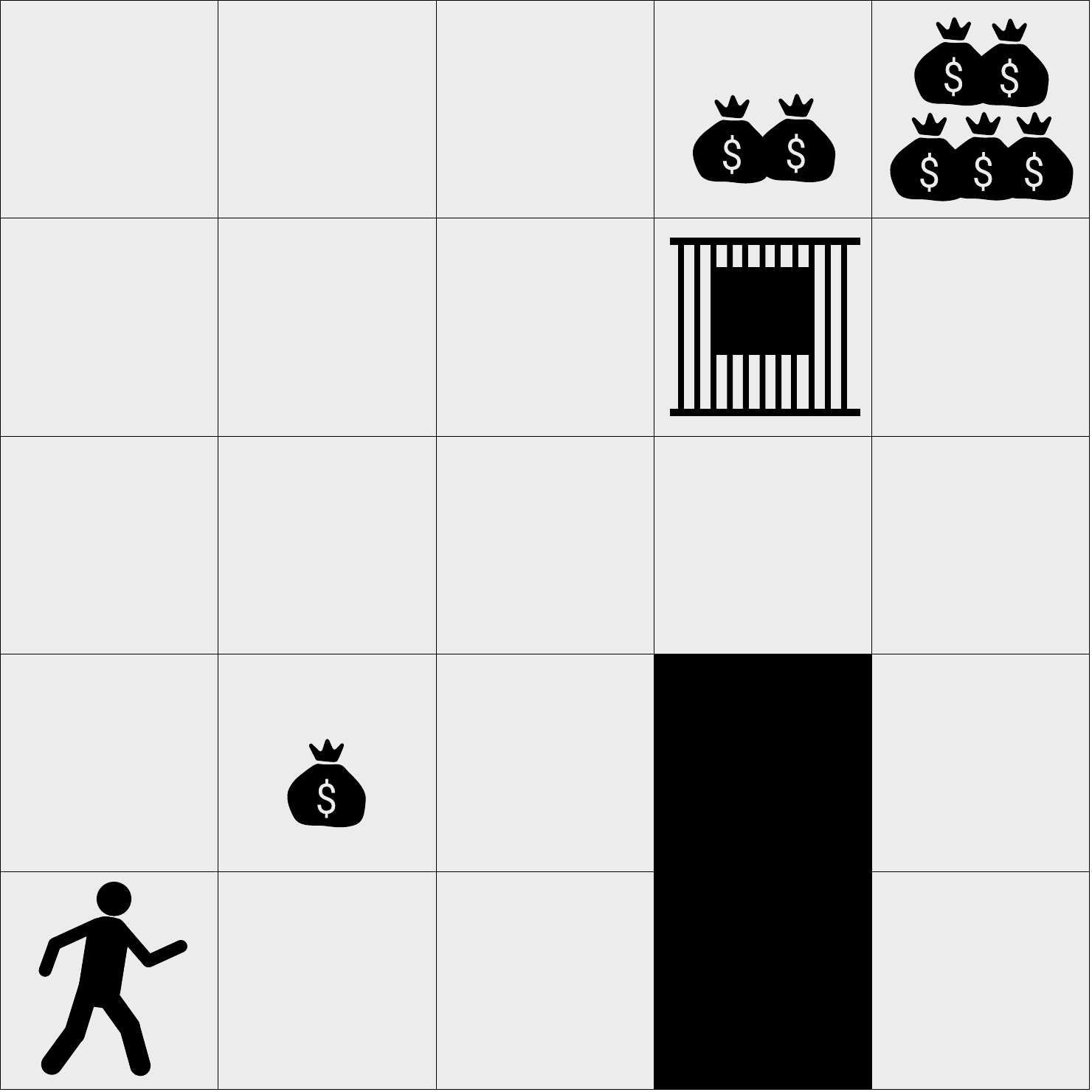}
		\caption{\label{fig:grid_small_env}The ``prison'' gridworld.}
	\end{minipage}\\[1em]
	\begin{minipage}[t]{.99\textwidth}
		\centering
		\includegraphics[width=0.95\linewidth]{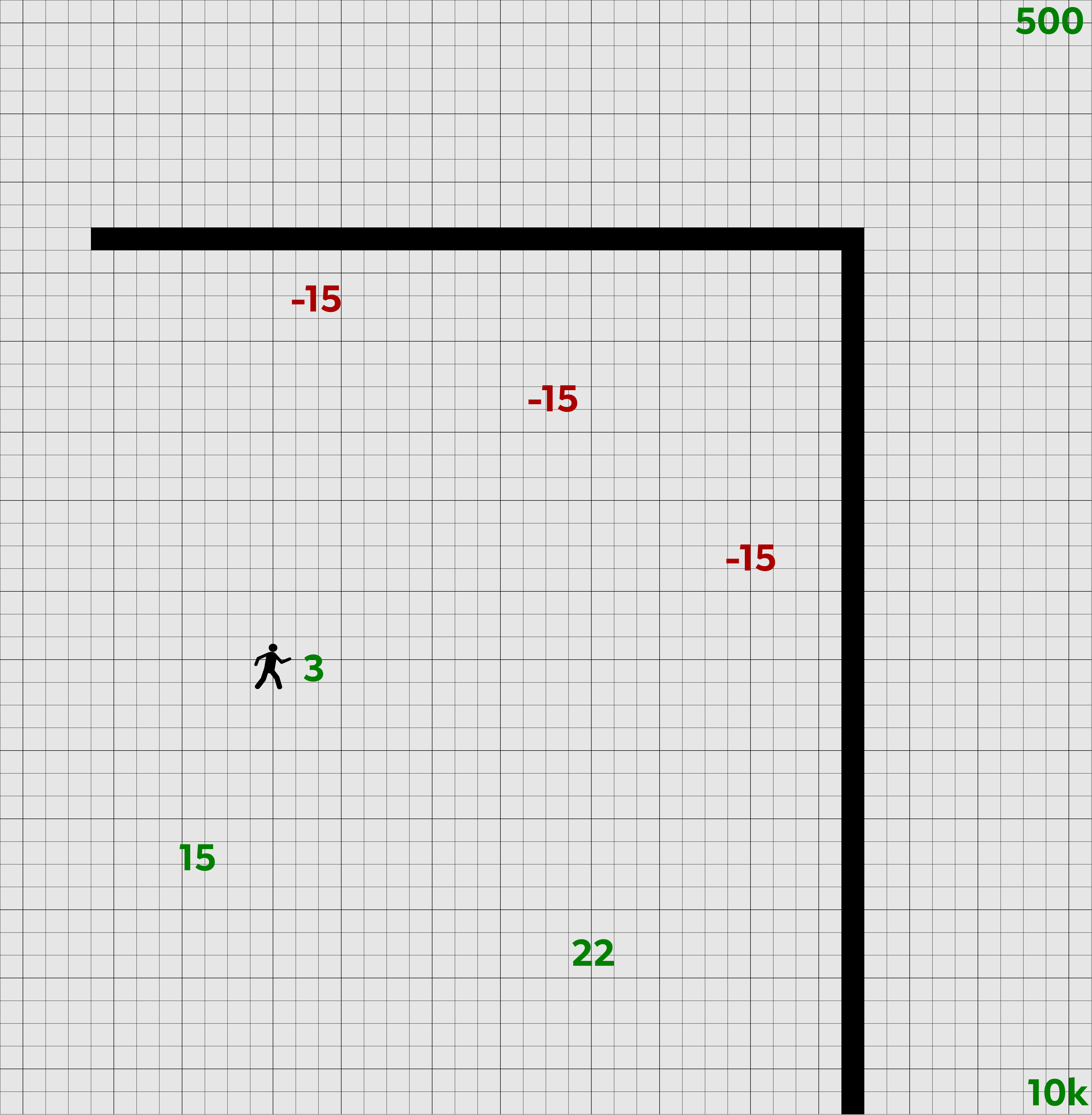}
		\caption{\label{fig:grid_wall_env}The ``wall'' gridworld.}
	\end{minipage}
\end{figure}

\clearpage

\subsubsection{Gridworlds}
\label{sssec:grids}
These environments (Figure \ref{fig:grid_simple_env}, \ref{fig:grid_small_env}, and \ref{fig:grid_wall_env}) share the same structure. The agent navigates in a grid by moving left, right, up, or down. 
Black cells cannot be visited, while any action in the prison has only an arbitrarily small probability of success. If it fails, the agent does not move. In practice, the prison almost completely prevents any further exploration. 
The reward is 0 everywhere except in treasure or penalty cells and is given for executing an action in the state, i.e., on transitions.
Treasure cells (denoted by a money bag or a green number) give a bonus reward of different magnitude and end the episode. Penalty cells (denoted by a red number) give a penalty reward and end the episode. The agent also gets a small penalty of -0.01 at each step. The goal, thus, is to find the biggest treasure using as few steps as possible. 
The initial position is fixed such that it is far from the biggest treasure and close to the smallest one.

\begin{figure}
	\centering
	\includegraphics[width=0.9\linewidth,trim=0 1.1em 15.5em 0em, clip]{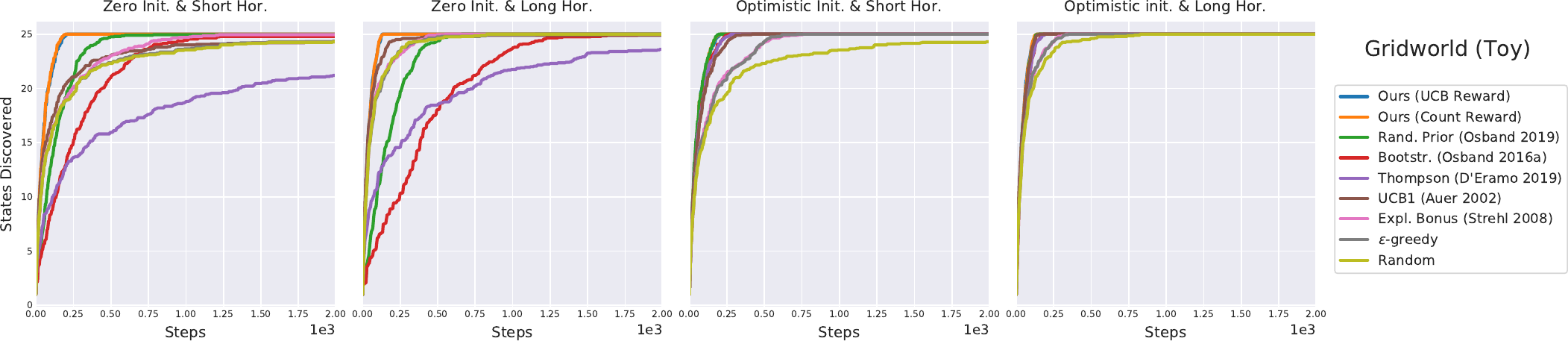}\\
	\includegraphics[width=0.9\linewidth,trim=0 0 15.5em 1em, clip]{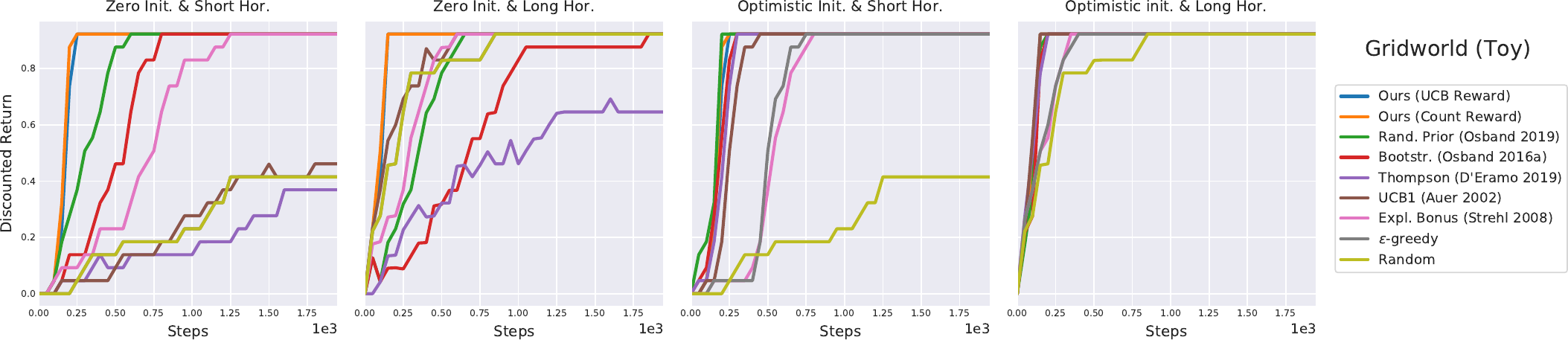}
	\caption{\label{fig:grid_simple_res}\textbf{Results on the toy gridworld} averaged over 20 seeds. Bootstrapped and bonus-based exploration perform well, but cannot match the proposed one (blue and orange line overlap).}
\end{figure}

\begin{figure}
	\centering
	\includegraphics[width=0.9\linewidth,trim=0 1.1em 15.5em 0em, clip]{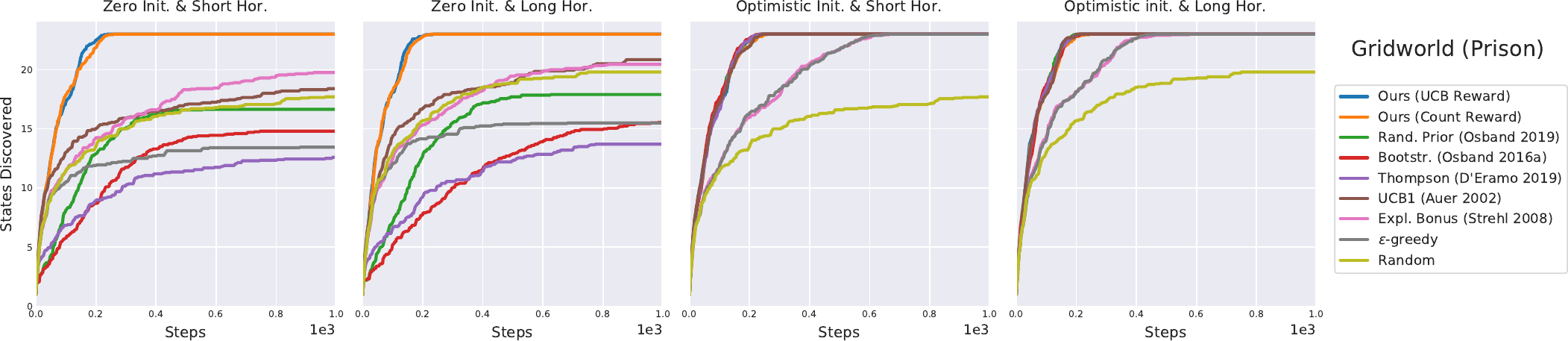}\\
	\includegraphics[width=0.9\linewidth,trim=0 0 15.5em 1em, clip]{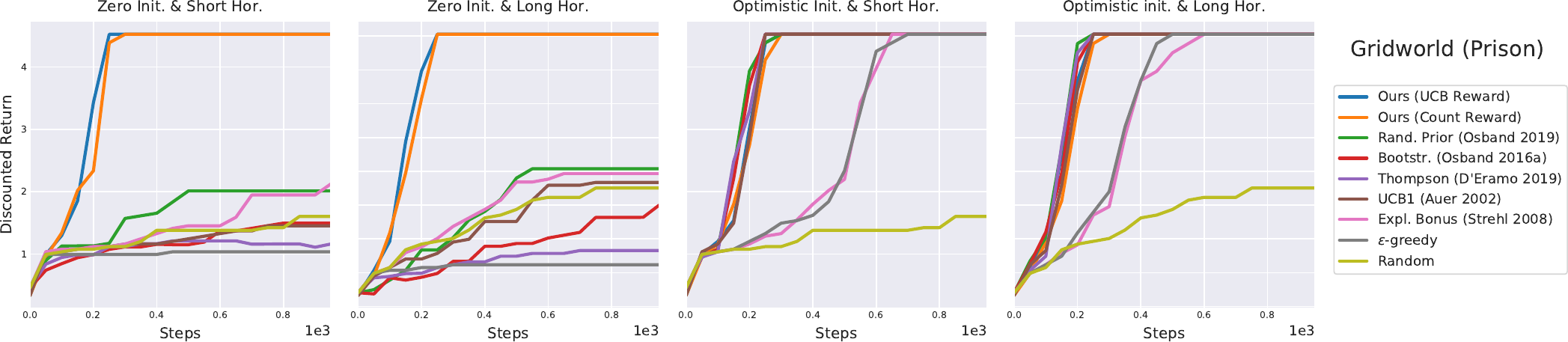}
	\caption{\label{fig:grid_small_res}\textbf{The ``prison'' and distractors} affect the performance of all algorithms but ours.}
\end{figure}

\begin{figure}
	\centering
	\includegraphics[width=0.9\linewidth,trim=0 1.1em 15.5em 0em, clip]{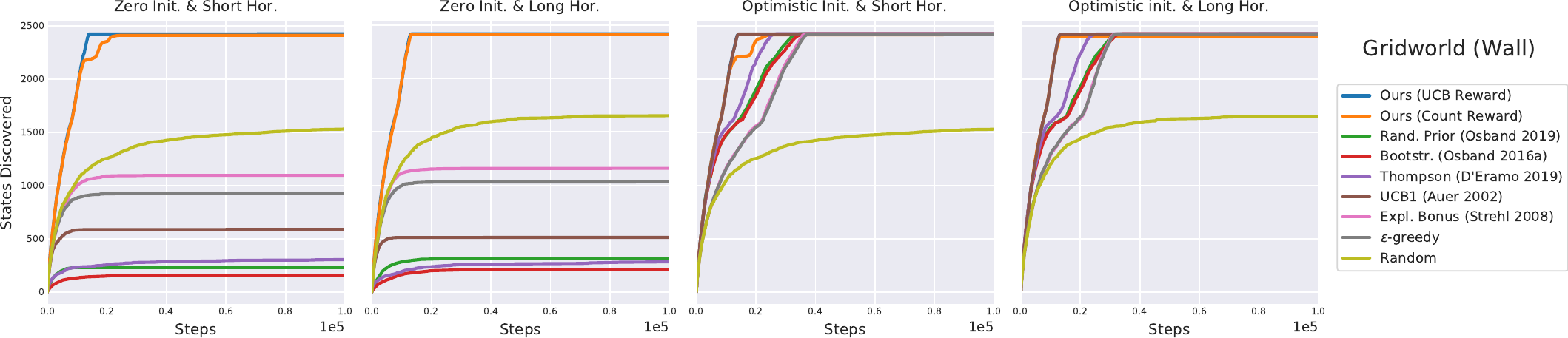}\\
	\includegraphics[width=0.9\linewidth,trim=0 0 15.5em 1em, clip]{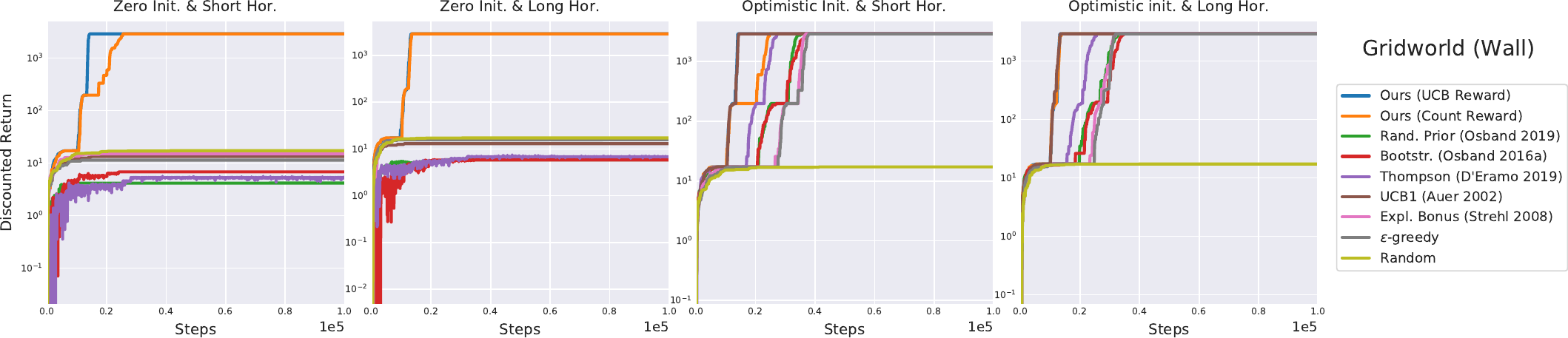}
	\caption{\label{fig:grid_wall_res}\textbf{The final gridworld} emphasizes the better performance of the proposed algorithms, especially the one using UCB reward. Return plot is in log scale.}
\end{figure}

Similarly to the deep gridworld, these domains are designed to highlight the difficulty of learning with many distractors (the smaller treasures) and, thus, of many local optima. The addition of the constant penalty at each step further discourages exploration and makes the distractors more appealing, since ending the episode will stop receiving penalties. Each domain has increasing difficulty and introduces additional challenges, as we explain below. 
\\[5pt]
\textbf{The ``toy'' gridworld.} We start with a simple $5\times5$ gridworld with one single reward of value 1, as shown in Figure \ref{fig:grid_simple_env}. The focus of this domain is learning with sparse reward, without distractors or additional difficulties. The optimal policy takes only nine steps to reach the treasure, and the ``short horizon'' scenario ends after eleven steps.
\\[5pt]
\textbf{The ``prison'' gridworld.} Shown in Figure \ref{fig:grid_small_env}, this domain increases the value of the furthest reward to 5, adds two distractors of value 1 and 2, and adds a prison cell. The first distractor is close to the initial position, while the second is close to the goal reward. The prison cell is also close to the goal. The optimal policy navigates around the first distractor, then right below the prison, and finally up to the goal. The prison 
highlights the importance of seeking states with low long-term visitation count, rather than low immediate count. As discussed in Section \ref{ssec:toy_example}, in fact, current count-based algorithms cannot deal with these kinds of states efficiently.
\\[5pt]
\textbf{The ``wall'' gridworld.} Shown in Figure \ref{fig:grid_wall_env}, the last gridworld is characterized by increased grid size ($50\times50$) and reward magnitude, and by the wall separating the grid into two areas. 
The first area, where the agent starts, has six small rewards (both treasures and penalties). The second area has two bigger treasures in the upper-right and bottom-right corners, of value 500 and 10,000, respectively. The optimal policy brings the agent beyond the wall and then to the bottom right corner, where the highest reward lies. The wall significantly increases the difficulty, due to the narrow passage that the agent needs to find in order to visit the second area of the grid. Learning is even harder when the horizon is short, as the agent cannot afford to lose time randomly looking for new states. To increase the chance of success of all algorithms, we set the ``short horizon'' to 330 steps, and 135 are needed to reach the reward of 10,000. 
\\[5pt]
\noindent
\textbf{Results.}
The gridworld environments confirm previous results. Without distractors (toy gridworld, Figure \ref{fig:grid_simple_res}), bootstrapped algorithms (green and red) perform well, and so does using the auxiliary visitation bonus (pink). Neither, however, match ours (blue and orange, almost overlapping). Increasing the horizon helps the remainder algorithms, including random exploration (light green), except for approximate Thompson sampling (purple).
In the next gridworld, however, the distractors and the prison cell substantially harm all algorithms except ours (Figure \ref{fig:grid_small_res}). Without optimistic initialization, in fact, existing algorithms cannot find the highest reward even with a long horizon, and all converge to local optima. This behavior was expected, given the study of Section \ref{ssec:toy_example}. Finally, the wall gridworld results emphasize even more the superiority of our algorithms (Figure \ref{fig:grid_wall_res}). With zero initialization, in fact, every other algorithm cannot go beyond the wall and find even the reward of 500. The results also stress that using the UCB reward visitation value (blue) over the count reward (orange) performs slightly best.

This evaluation strengthens the findings of previous experiments. First, it stresses how difficult it is for existing algorithms to learn with distracting rewards and a short horizon. Second, it shows that our proposed approach overcomes these challenges. Next, we show how efficiently the algorithms explore in terms of final visitation count and sample complexity.

\clearpage

\begin{figure}
\vspace*{-3em}
\includegraphics[width=\linewidth]{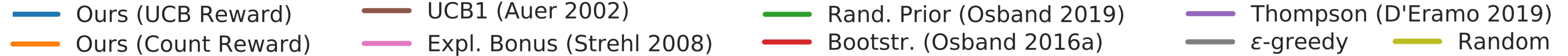}
\end{figure}

\clearpage

\begin{table}
	\centering
	\caption{\label{tab:recap}\textbf{Results recap} for the ``zero initialization short horizon'' scenarios. Only the proposed exploration strategy always discovers all states and solves the tasks in all 20 seeds.}
	\setstretch{0.95}
	\begin{tabular}{ c | c | c c }
	& Algorithm & Discovery (\%) & Success (\%) \\
	\hline
	& \textbf{Ours (UCB Reward)} & $\mathbf{100 \pm 0}$ & $\mathbf{100 \pm 0}$ \\
	& \textbf{Ours (Count Reward)} & $\mathbf{100 \pm 0}$ & $\mathbf{100 \pm 0}$ \\
	& Rand. Prior (Osband 2019) & $99.90 \pm 0.01$ & $100 \pm 0$ \\
	& Bootstr. (Osband 2016a) & $99.77 \pm 0.05$ & $100 \pm 0$ \\
	Deep Sea & Bootstr. (D'Eramo 2019) & $63.25 \pm 3.31$ & $0 \pm 0$ \\
	& UCB1 (Auer 2002) & $55.72 \pm 0.34$ & $0 \pm 0$ \\
	& Expl. Bonus (Strehl 2008) & $85.65 \pm 1.0$ & $0 \pm 0$ \\
	& $\epsilon$-greedy & $57.74 \pm 1.11$ & $0 \pm 0$ \\
	& Random & $58.59 \pm 1.35$ & $0 \pm 0$ \\
	\hline
	& \textbf{Ours (UCB Reward)} & $\mathbf{100 \pm 0}$ & $\mathbf{100 \pm 0}$ \\
	& \textbf{Ours (Count Reward)} & $\mathbf{100 \pm 0}$ & $\mathbf{100 \pm 0}$ \\
	& Rand. Prior (Osband 2019) & $69.60 \pm 2.96$ & $13.07 \pm 2.96$ \\
	& Bootstr. (Osband 2016a) & $52.44 \pm 7.55$ & $18.77 \pm 9.24$ \\
	Taxi & Bootstr. (D'Eramo 2019) & $22.44 \pm 1.81$ & $1.9 \pm 1.47$ \\
	& UCB1 (Auer 2002) & $31.17 \pm 0.70$ & $1.53 \pm 1.37$ \\
	& Expl. Bonus (Strehl 2008) & $74.62 \pm 2.24$ & $17.6 \pm 2.56$ \\
	& $\epsilon$-greedy & $29.64 \pm 0.98$ & $1.92 \pm 1.49$ \\
	& Random & $29.56 \pm 0.98$ & $1.92 \pm 1.49$ \\
	 \hline
	 & \textbf{Ours (UCB Reward)} & $\mathbf{100 \pm 0}$ & $\mathbf{100 \pm 0}$ \\
	 & \textbf{Ours (Count Reward)} & $\mathbf{100 \pm 0}$ & $\mathbf{100 \pm 0}$ \\
	 & Rand. Prior (Osband 2019) & $75 \pm 9.39$ & $59.03 \pm 4.2$ \\
	 & Bootstr. (Osband 2016a) & $60.82 \pm 11.78$ & $63.35 \pm 6.89$ \\
	 Deep Gridworld & Bootstr. (D'Eramo 2019) & $63.73 \pm 10.35$ & $56.85 \pm 0.06$ \\
 & UCB1 (Auer 2002) & $92.18 \pm 0.36$ & $56.88 \pm 0$ \\
	 & Expl. Bonus (Strehl 2008) & $95.45 \pm 1.57$ & $69.81 \pm 8.84$ \\
	 & $\epsilon$-greedy & $74.36 \pm 4.42$ & $56.88 \pm 0$ \\
	 & Random & $92.45 \pm 0.64$ & $56.88 \pm 0$ \\
	 \hline
	 & \textbf{Ours (UCB Reward)} & $\mathbf{100 \pm 0}$ & $\mathbf{100 \pm 0}$ \\
	 & \textbf{Ours (Count Reward)} & $\mathbf{100 \pm 0}$ & $\mathbf{100 \pm 0}$ \\
	 & Rand. Prior (Osband 2019) & $99.8 \pm 0.39$ & $100 \pm 0$ \\
	 & Bootstr. (Osband 2016a) & $99.2 \pm 0.72$ & $100 \pm 0$ \\
	 Gridworld (Toy) & Bootstr. (D'Eramo 2019) & $84.8 \pm 4.33$ & $40 \pm 21.91$ \\
	 & UCB1 (Auer 2002) & $97.4 \pm 0.99$ & $50 \pm 22.49$ \\
	 & Expl. Bonus (Strehl 2008) & $99.8 \pm 0.39$ & $100 \pm 0$ \\
	 & $\epsilon$-greedy & $97.4 \pm 1.17$ & $45 \pm 22.25$ \\
	 & Random & $97.2 \pm 1.15$ & $45 \pm 22.25$ \\
	 \hline
	 & \textbf{Ours (UCB Reward)} & $\mathbf{100 \pm 0}$ & $\mathbf{100 \pm 0}$ \\
	 & \textbf{Ours (Count Reward)} & $\mathbf{100 \pm 0}$ & $\mathbf{100 \pm 0}$ \\
	 & Rand. Prior (Osband 2019) & $72.39 \pm 6.9$ & $44.44 \pm 14.69$ \\
	 & Bootstr. (Osband 2016a) & $64.35 \pm 11.1$ & $33.03 \pm 8.54$ \\
	 Gridworld (Prison) & Bootstr. (D'Eramo 2019) & $54.78 \pm 6.69$ & $25.61 \pm 8.39$ \\
	 & UCB1 (Auer 2002) & $80.78 \pm 2.56$ & $32.31 \pm 4.11$ \\
	 & Expl. Bonus (Strehl 2008) & $85.87 \pm 3$ & $46.96 \pm 12.35$ \\
	 & $\epsilon$-greedy & $58.38 \pm 5.27$ & $22.84 \pm 2.48$ \\
	 & Random & $76.96 \pm 3.08$ & $35.36 \pm 10.37$ \\
	 \hline
	 & \textbf{Ours (UCB Reward)} & $\mathbf{100 \pm 0}$ & $\mathbf{100 \pm 0}$ \\
	 & \textbf{Ours (Count Reward)} & $\mathbf{100 \pm 0}$ & $\mathbf{100 \pm 0}$ \\
	 & Rand. Prior (Osband 2019) & $9.44 \pm 3.14$ & $0.14 \pm 0.06$ \\
	 & Bootstr. (Osband 2016a) & $6.35 \pm 2.74$ & $0.23 \pm 0.08$ \\
	 Gridworld (Wall) & Bootstr. (D'Eramo 2019) & $12.55 \pm 4.45$ & $0.18 \pm 0.08$ \\
	 & UCB1 (Auer 2002) & $24.7 \pm 4.45$ & $0.46 \pm 0.03$ \\
	 & Expl. Bonus (Strehl 2008) & $45.16 \pm 2.29$ & $0.52 \pm 0.04$ \\
	 & $\epsilon$-greedy & $38.15 \pm 1.95$ & $0.39 \pm 0.08$ \\
	 & Random & $65.28 \pm 0.45$ & $0.59 \pm 0$ \\
 \end{tabular}
\end{table}

\clearpage

\subsection{Part 2: In-Depth Investigation}
\label{ssec:eval_part2}
In this section, we present an in-depth analysis of issues associated with learning with sparse rewards, explaining in more detail why our approach outperformed existing algorithms in the previous evaluation.
We start by reporting the visitation count at the end of the learning, showing that our algorithms explore the environment uniformly. We continue with a study on the impact of the visitation discount $\gamma_w$, showing how it affects the performance of the W-function. Next, we present the empirical sample complexity analysis of all algorithms on the deep sea domain, showing that our approach scales gracefully with the number of states. Finally, we evaluate the algorithms in the infinite horizon setting and in stochastic MDPs, evaluating the accuracy of the Q-function and the empirical sample complexity. Even in these scenarios, our approach outperforms existing algorithms.

\begin{figure}[t]
	\centering 
	\raisebox{22pt}{\rotatebox[origin=t]{90}{\textbf{Deep Sea}}}
	\begin{subfigure}[b]{.18\linewidth} 
		\centering 
		\raisebox{3pt}{\textbf{Ours}}
		\includegraphics[width=\linewidth]{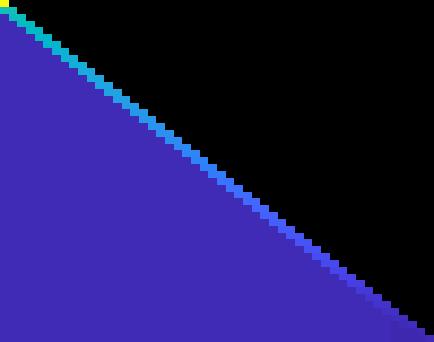}
	\end{subfigure}
	\hfill
	\begin{subfigure}[b]{.18\linewidth} 
		\centering 
		\raisebox{3pt}{\textbf{Bootstr.}}
		\includegraphics[width=\linewidth]{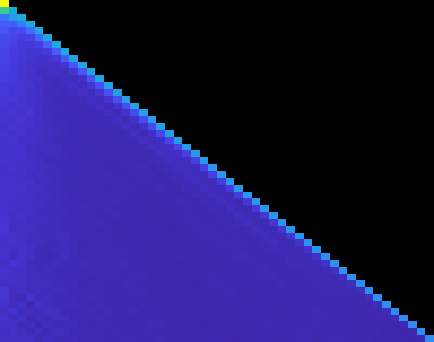}
	\end{subfigure} 
	\hfill
	\begin{subfigure}[b]{.18\linewidth} 
		\centering 
		\raisebox{3pt}{\textbf{UCB1}}
		\includegraphics[width=\linewidth]{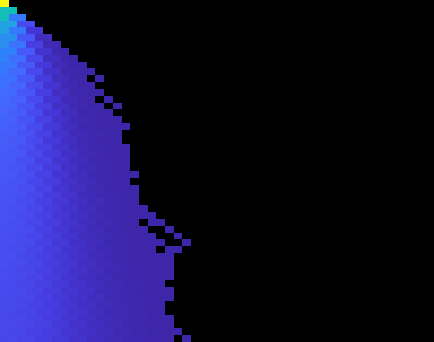}
	\end{subfigure}
	\hfill
	\begin{subfigure}[b]{.18\linewidth} 
		\centering 
		\raisebox{3pt}{\textbf{Expl. Bonus}}
		\includegraphics[width=\linewidth]{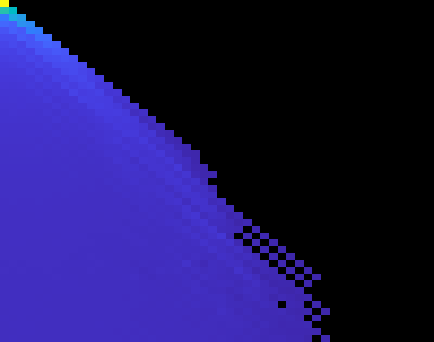}
	\end{subfigure} 
	\hfill
	\begin{subfigure}[b]{.18\linewidth} 
		\centering 
		\raisebox{3pt}{\textbf{$\epsilon$-greedy}}
		\includegraphics[width=\linewidth]{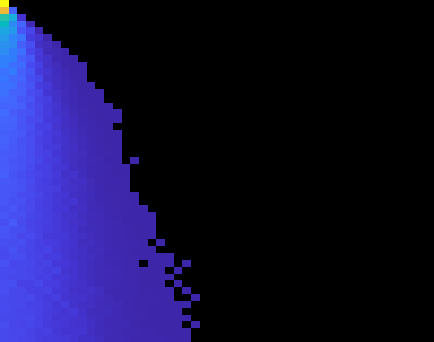}
	\end{subfigure} 
	\\[0.5em]
	\raisebox{22pt}{\rotatebox[origin=t]{90}{\textbf{Deep Grid}}}
	\begin{subfigure}[b]{.18\linewidth} 
		\centering 
		\includegraphics[width=\linewidth]{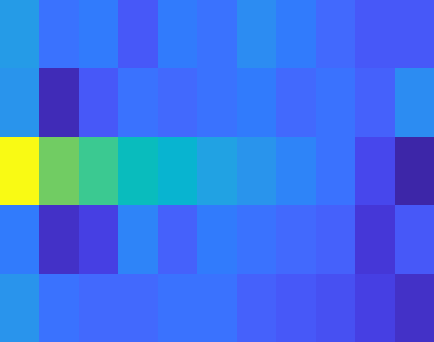}
	\end{subfigure}
	\hfill
	\begin{subfigure}[b]{.18\linewidth} 
		\centering 
		\includegraphics[width=\linewidth]{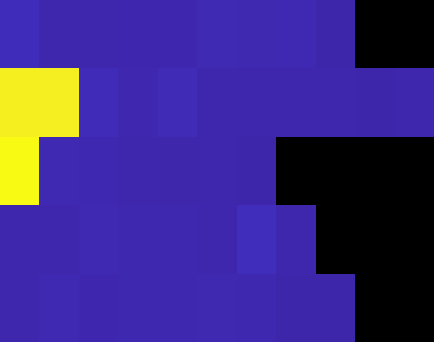}
	\end{subfigure} 
	\hfill
	\begin{subfigure}[b]{.18\linewidth} 
		\centering 
		\includegraphics[width=\linewidth]{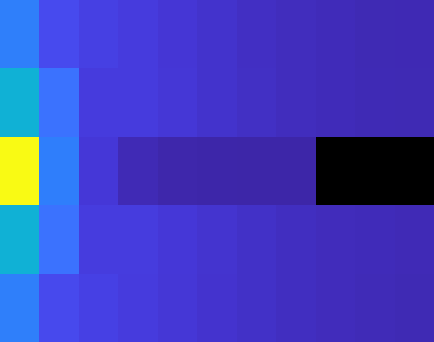}
	\end{subfigure}
	\hfill
	\begin{subfigure}[b]{.18\linewidth} 
		\centering 
		\includegraphics[width=\linewidth]{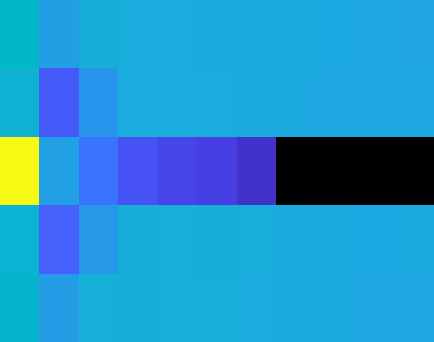}
	\end{subfigure} 
	\hfill
	\begin{subfigure}[b]{.18\linewidth} 
		\centering 
		\includegraphics[width=\linewidth]{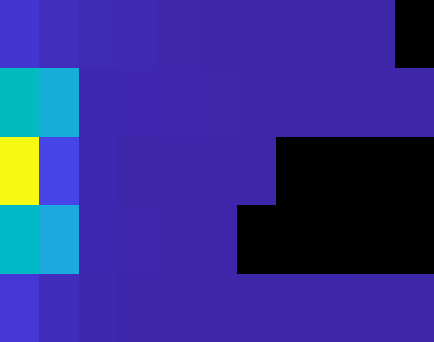}
	\end{subfigure} 
	\\[0.5em]
	\raisebox{22pt}{\rotatebox[origin=t]{90}{\textbf{Grid (Prison)}}}	\begin{subfigure}[b]{.18\linewidth} 
		\centering 
		\includegraphics[width=\linewidth]{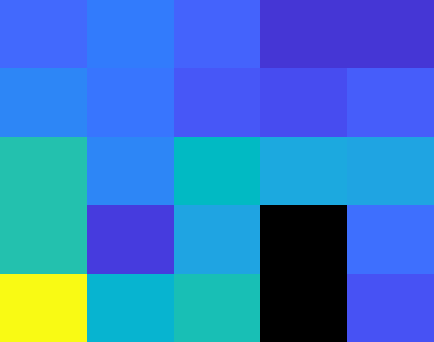}
	\end{subfigure}
	\hfill
	\begin{subfigure}[b]{.18\linewidth} 
		\centering 
		\includegraphics[width=\linewidth]{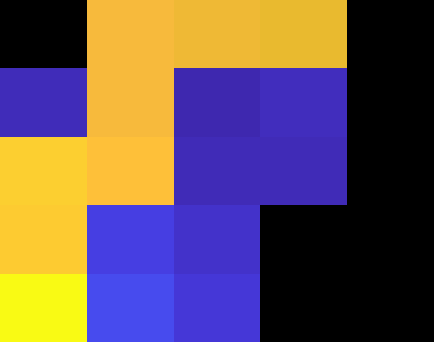}
	\end{subfigure} 
	\hfill
	\begin{subfigure}[b]{.18\linewidth} 
		\centering 
		\includegraphics[width=\linewidth]{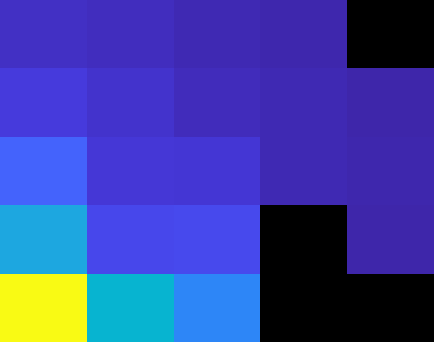}
	\end{subfigure}
	\hfill
	\begin{subfigure}[b]{.18\linewidth} 
		\centering 
		\includegraphics[width=\linewidth]{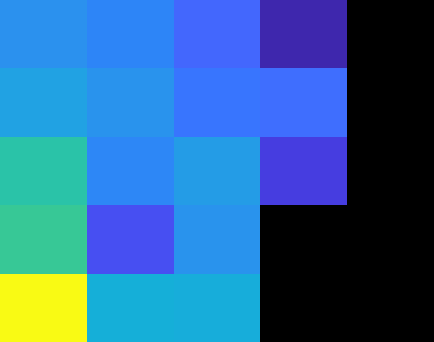}
	\end{subfigure} 
	\hfill
	\begin{subfigure}[b]{.18\linewidth} 
		\centering 
		\includegraphics[width=\linewidth]{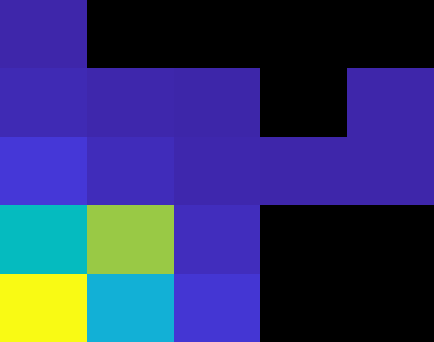}
	\end{subfigure} 
	\\[0.5em]
	\raisebox{22pt}{\rotatebox[origin=t]{90}{\textbf{Grid (Wall)}}}	\begin{subfigure}[b]{.18\linewidth} 
		\centering 
		\includegraphics[width=\linewidth]{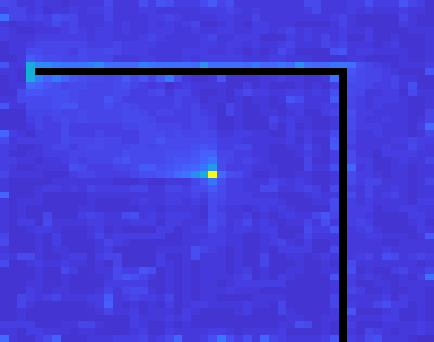}
	\end{subfigure}
	\hfill
	\begin{subfigure}[b]{.18\linewidth} 
		\centering 
		\includegraphics[width=\linewidth]{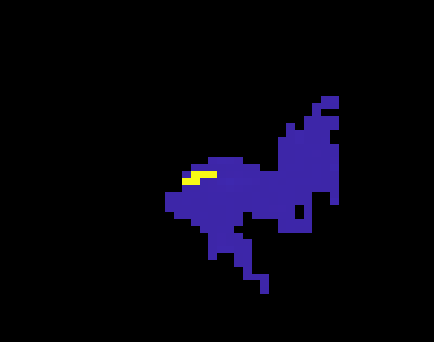}
	\end{subfigure} 
	\hfill
	\begin{subfigure}[b]{.18\linewidth} 
		\centering 
		\includegraphics[width=\linewidth]{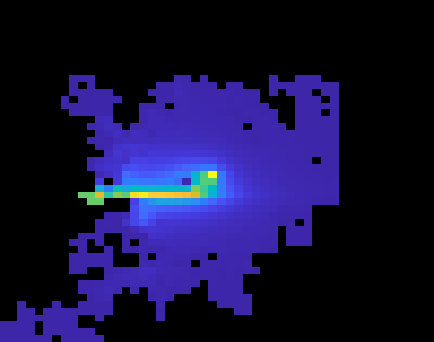}
	\end{subfigure}
	\hfill
	\begin{subfigure}[b]{.18\linewidth} 
		\centering 
		\includegraphics[width=\linewidth]{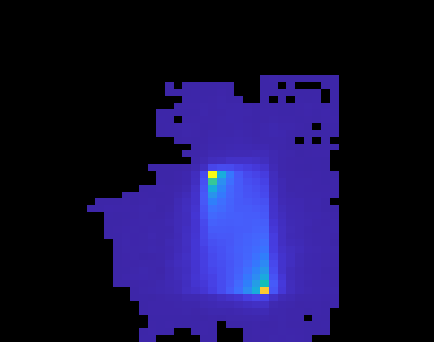}
	\end{subfigure} 
	\hfill
	\begin{subfigure}[b]{.18\linewidth} 
		\centering 
		\includegraphics[width=\linewidth]{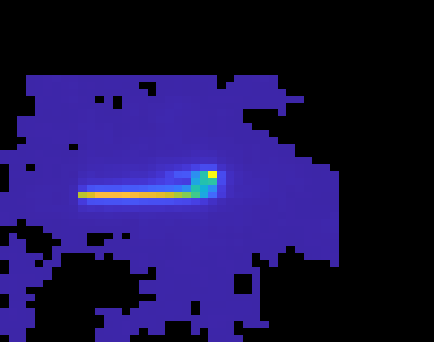}
	\end{subfigure} 
	\caption{\label{fig:count} \textbf{Visitation count at the end of the learning.} The seed is the same across all images. Initial states naturally have a higher count than other states. Recall that the upper portion of the deep sea and some gridworlds cells cannot be visited. 
	Only the proposed methods explore the environment uniformly\protect\footnotemark.  
	Other algorithms myopically focus on distractors.
	}
\end{figure} 
\footnotetext{Figures show the count of UCB-based W-function. The initial state (bottom left) has naturally higher counts because the agent always starts there.
Count-based W-function performed very similarly.}

\subsubsection{Visitation Count at the End of the Learning}
\label{sssec:visit_count_map}
In Section \ref{ssec:vv_init} and \ref{ssec:notes}, we discussed that the proposed behavior policies guarantee that an action is not executed twice before all actions are executed once, under the uniform count assumption. We also acknowledged that this assumption cannot hold in practice because of state resets, or because the agent may need to revisit the same state to explore new ones. For example, in the deep sea, the agent needs to traverse diagonal states multiple times to visit every state.
Nonetheless, in this section, we empirically show that our approach allows the agent to explore the environment as uniformly as possible. 

Figure \ref{fig:count} reports the state visitation count at the end of learning in the ``short-horizon zero-initialization'' scenario for some domains. 
Only the proposed method (first column) uniformly explores all states. Recall, in fact, that episodes reset after some steps or when terminal states are reached. Thus initial states (which are fixed) are naturally visited more often. The count in our method uniformly decreases proportionally to the distance from the initial states. This denotes that at each episode, starting from the same state, the agent followed different paths, exploring new regions of the environment uniformly. 
\\
Other algorithms suffer from distractors and local optima. 
In particular, UCB1 (third column) explores very slowly. The reason is that the agent myopically selects actions based on the \textit{immediate} count, which makes exploration highly inefficient. By selecting the action with the lowest immediate count, the agent does not take into account where the action will lead it, i.e., if future states have already been visited or not.
By contrast, our approach achieves deep exploration by taking into account the long-term visitation count of future states.
\\
The performance of bootstrapping (second column) is peculiar. It immediately converges to the first reward found in the gridworlds, barely exploring afterward, but performs very well on the deep sea. 
This hints that bootstrapping may be sensitive to any kind of reward, including negative penalties. 
In the deep sea, in fact, diagonal states provide the only intermediate feedback besides the final reward/penalty. Coincidentally, traversing the diagonal also leads to the only positive reward.
The agent may thus be guided by the reward received on the diagonal, even if it is a negative penalty.
This behavior can be explained by recalling that \citet{osband2019deep} regularize TD learning with the $\ell_2$-distance from a prior Q-table. The agent may therefore be attracted to any state providing some kind of feedback in order to minimize the regularization. 

\subsubsection{Impact of Visitation Value Discount Factor $\gamma_w$}
\label{sssec:gamma_sensitivity}
\begin{figure}[t]
	\centering
	\includegraphics[width=\linewidth,height=3.2cm]{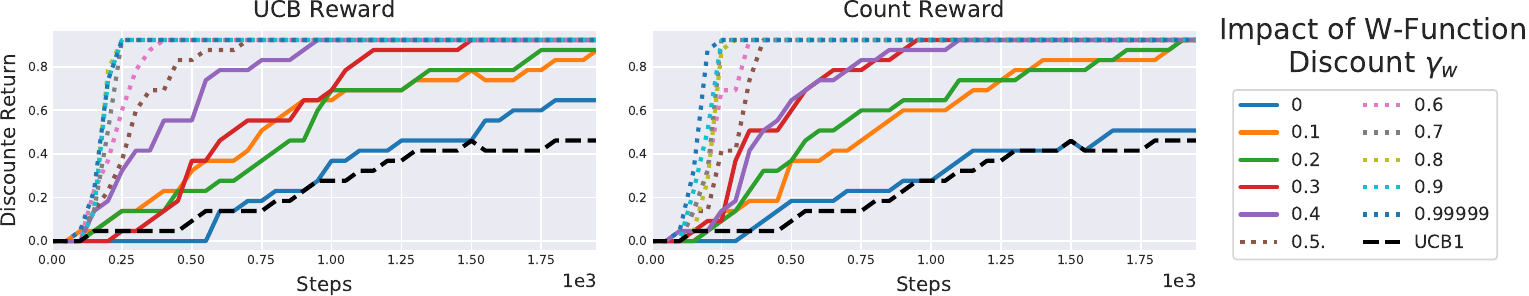}
	\caption{\label{fig:gamma_plot} \textbf{Performance of the our algorithms on varying of $\gamma_w$} on the ``toy'' gridworld (Figure \ref{fig:grid_small_env}). The higher $\gamma_w$, the more the agent explores, allowing to discover the reward faster.}
\end{figure}

\begin{figure}[t]
	\centering 
	\raisebox{33pt}{\rotatebox[origin=t]{90}{\textbf{Grid (Toy)}}}
	\begin{subfigure}[b]{.18\linewidth} 
		\centering 
		\includegraphics[width=\linewidth]{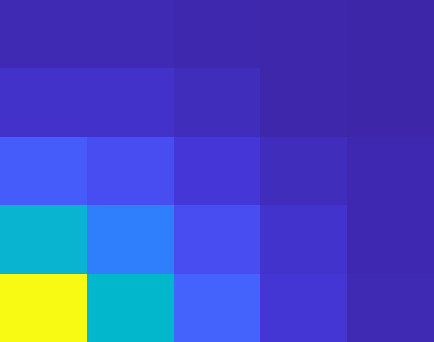}
		\raisebox{3pt}{\footnotesize{$\gamma_w = 0$}}
	\end{subfigure}
	\hfill
	\begin{subfigure}[b]{.18\linewidth} 
		\centering 
		\includegraphics[width=\linewidth]{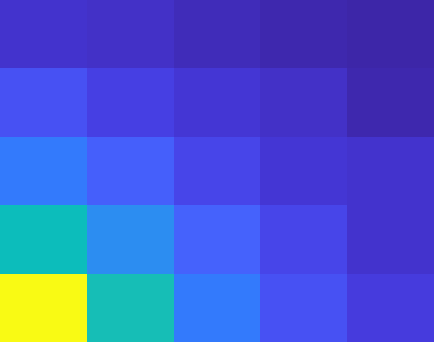}
		\raisebox{3pt}{\footnotesize{$\gamma_w = 0.3$}}
	\end{subfigure} 
	\hfill
	\begin{subfigure}[b]{.18\linewidth} 
		\centering 
		\includegraphics[width=\linewidth]{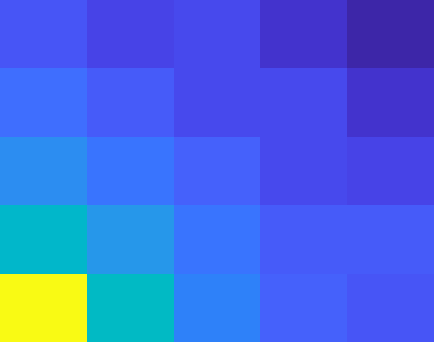}
		\raisebox{3pt}{\footnotesize{$\gamma_w = 0.7$}}
	\end{subfigure}
	\hfill
	\begin{subfigure}[b]{.18\linewidth} 
		\centering 
		\includegraphics[width=\linewidth]{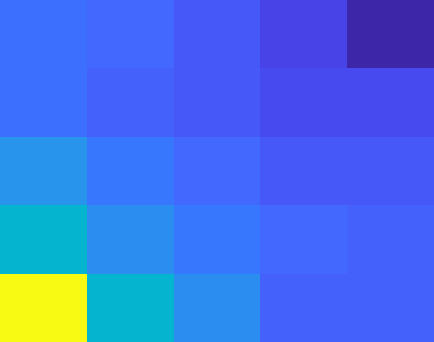}
		\raisebox{3pt}{\footnotesize{$\gamma_w = 0.9$}}
	\end{subfigure} 
	\hfill
	\begin{subfigure}[b]{.18\linewidth} 
		\centering 
		\includegraphics[width=\linewidth]{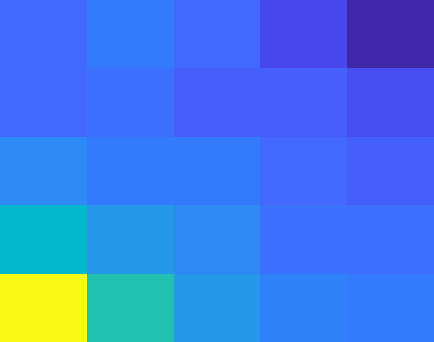}
		\raisebox{3pt}{\footnotesize{$\gamma_w = 0.99999$}}
	\end{subfigure} 
	\caption{\label{fig:gamma_count} \textbf{Visitation count for $W^\beta_{\textsub{n}}$ at the end of the learning} on the same random seed (out of 20). The higher the discount, the more uniform the count is. $W^\beta_{\textsub{ucb}}$ performed very similarly.}
\end{figure}

Here, we investigate the effect of the visitation discount factor $\gamma_w$ on the ``toy'' gridworld (Figure~\ref{fig:grid_small_env}). Figure~\ref{fig:gamma_plot} shows that the higher the discount, the better. As expected, with $\gamma_w = 0$ the algorithms behave very similarly to UCB1. In this case, in fact, the proposed approach and UCB1 are equivalent. However, the learning curves are not exactly the same because the W-function is updated with TD learning at every step. 
\\
The better exploration of our approach is also confirmed by Figure \ref{fig:gamma_count}, showing the count at the end of the learning. The higher $\gamma_w$, the more uniform the exploration is because with a higher discount the agent will look further in the future for visitation rewards. With smaller $\gamma_w$, instead, the agent rarely discovers states far from the initial position, as it does not look for future rewards. As a rule of thumb, it is appropriate to have $\gamma_w$ equal or larger than $\gamma$.

\subsubsection{Empirical Sample Complexity on the Deep Sea}
\label{sssec:empirical_sample_deep}

\begin{figure}[t]
	\centering
	\includegraphics[width=0.93\linewidth]{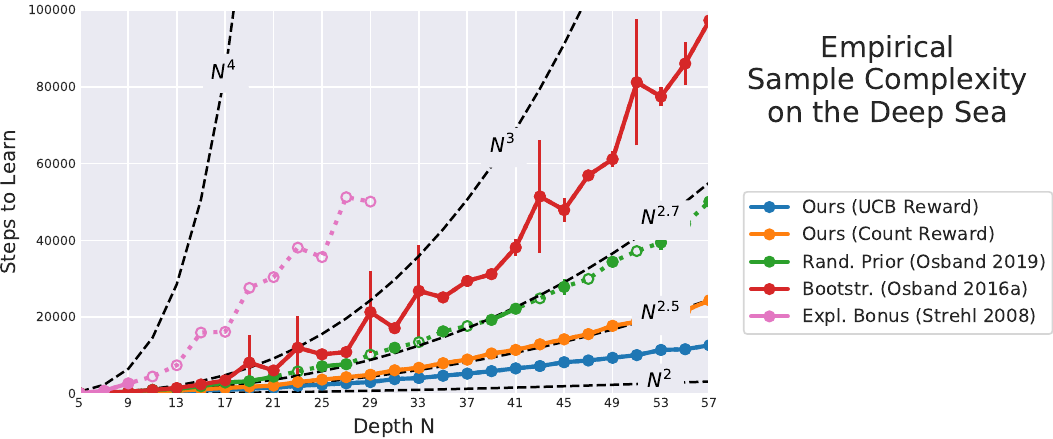}
	\caption{\label{fig:empirical_sample_deep} \textbf{Steps to learn on varying the deep sea size} $N = 5, 7, \ldots, 59$. Each filled dot denotes the average over ten runs with 500,000 steps limit. Error bars denote 95\% confidence interval. In empty dots connected with dashed lines, the algorithm did not learn within the step limit at least once. In this case, the average is over converged runs only. Missing algorithms (random, $\epsilon$-greedy, UCB1, Thompson sampling) performed poorly and are not reported. Only our algorithms learn with a confidence interval close to zero, and attain the lowest sample complexity.}
\end{figure}

Here, we propose the same evaluation presented by \citet{osband2018randomized} to investigate the empirical sample complexity of the proposed algorithms, and to assess how they scale to large problems. 
Figure \ref{fig:empirical_sample_deep} plots training steps to learn the optimal policy as a function of the environment size $N$.
Only our approach (blue and orange) scales gracefully to large problem sizes. Results suggest an empirical scaling of $\bigO(N^{2.5})$ for the visitation-value-based with count reward, and even smaller for UCB reward. 
Bootstrapping attains an empirical complexity of $\bigO(N^{3})$, confirming the findings of \citet{osband2019deep}.
However, in many cases ($N = 23, 27, 33, 37, 43, 47, 51, 55$) there was one seed for which bootstrapping either did not learn within the step limit (green dashed line, empty dots) due to premature convergence, or learned after substantially more steps than the average (red dots, large error bar). 
Missing algorithms (random, $\epsilon$-greedy, UCB1, approximate Thompson sampling) performed extremely poorly, often not learning within 500,000 steps even for small $N$, and are thus not reported.

\begin{figure}[t]
	\centering 
	\raisebox{55pt}{\rotatebox[origin=t]{90}{\scriptsize \sffamily Timestep}}
	\begin{subfigure}[b]{.3\linewidth} 
		\centering 
		\includegraphics[width=\linewidth]{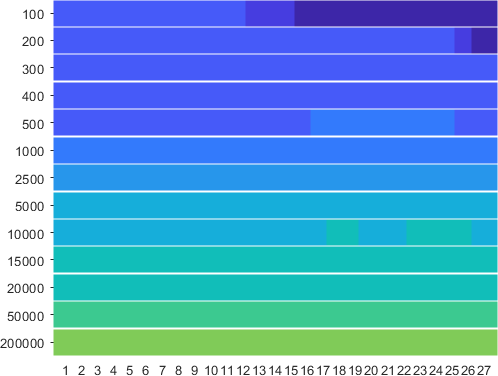}
		\caption{\label{fig:chain_count_vv3}Ours (UCB Reward)}
	\end{subfigure}
	\hfill
	\begin{subfigure}[b]{.3\linewidth} 
		\centering 
		\includegraphics[width=\linewidth]{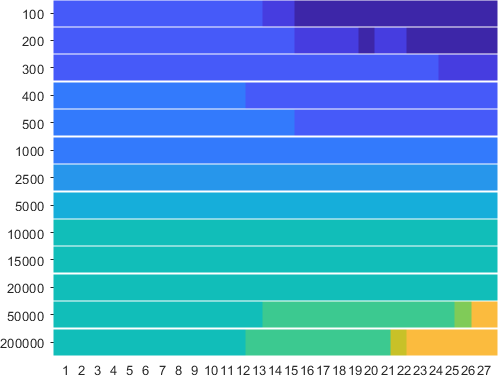}
		\caption{\label{fig:chain_count_vv2}Ours (Count Reward)}
	\end{subfigure}
	\hfill
	\begin{subfigure}[b]{.3\linewidth} 
		\centering 
		\includegraphics[width=\linewidth]{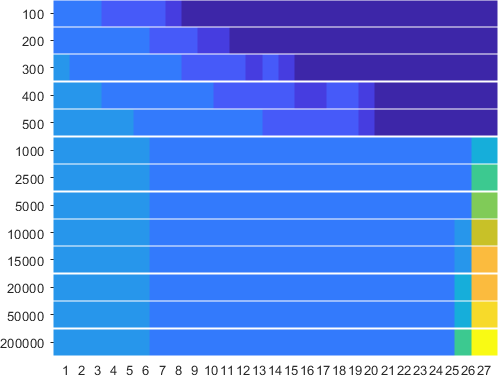}
		\caption{\label{fig:chain_count_ucb}UCB1}
	\end{subfigure} 
	\hfill
	\begin{subfigure}[b]{.04\linewidth} 
		\includegraphics[width=\linewidth]{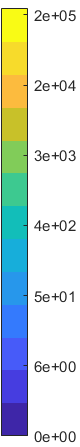}
		\centering \vspace*{6pt}
	\end{subfigure} 
	\\[0.7em]
	\raisebox{75pt}{\rotatebox[origin=t]{90}{\scriptsize \sffamily Timestep}}
	\begin{subfigure}[b]{.3\linewidth} 
		\centering 
		\includegraphics[width=\linewidth]{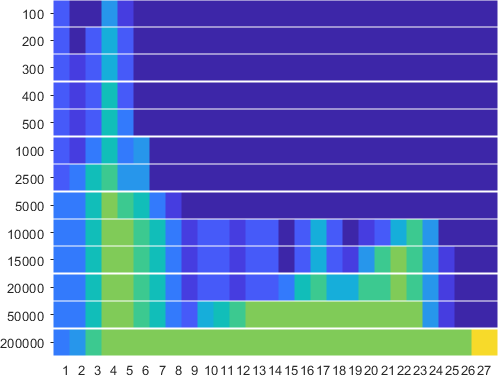}
		{\scriptsize \sffamily State}
		\caption{\label{fig:chain_count_mbie}Expl. Bonus}
	\end{subfigure}
	\hfill
	\begin{subfigure}[b]{.3\linewidth} 
		\centering 
		\includegraphics[width=\linewidth]{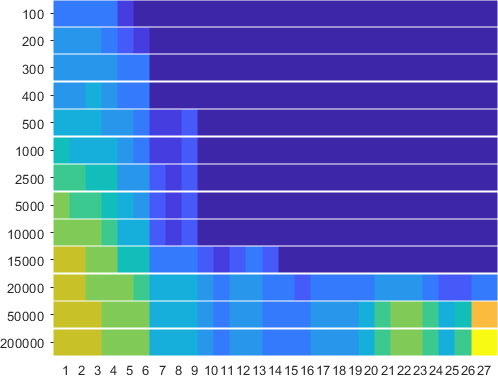}
		{\scriptsize \sffamily State}
		\caption{\label{fig:chain_count_thomp}Approx. Thompson Sampl.}
	\end{subfigure} 
	\hfill
	\begin{subfigure}[b]{.3\linewidth} 
		\centering 
		\includegraphics[width=\linewidth]{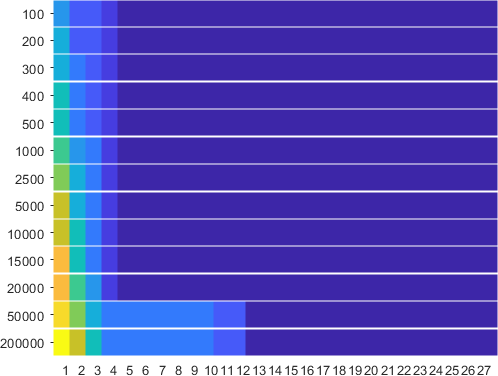}
		{\scriptsize \sffamily State}
		\caption{\label{fig:chain_count_egreedy}$\epsilon$-greedy}
	\end{subfigure} 
	\hfill
	\begin{subfigure}[b]{.04\linewidth} 
		\includegraphics[width=\linewidth]{img/Chain27_count_colorbar.png}
		\phantom{\scriptsize \sffamily State}
		\centering \vspace*{6pt}
	\end{subfigure} 
	\caption{\label{fig:chain_count} \textbf{Visitation counts for the chainworld} with 27 states (x-axis) over time (y-axis). 
	As time passes (top to bottom), the visitation count grows. Only our algorithms and UCB1 explore uniformly, producing a uniform colormap. However, UCB1 finds the last state (with the reward) much later. Once UCB1 finds the reward (Figure \ref{fig:chain_count_ucb}, step 1,000) the visitation count increases only in the last state. 
	This suggests that when UCB1 finds the reward contained in the last state it effectively stops exploring other states.
	By contrast, our algorithms keep exploring the environment for longer. This allows the agent to learn the true Q-function for all states, and explains why our algorithms achieve lower sample complexity and MSVE in Figures \ref{fig:empirical_sample_chain} and \ref{fig:chain_res}.}
\end{figure} 

\subsubsection{Infinite Horizon Stochastic Chainworld}
\label{sssec:chain}
This evaluation investigates how the algorithms perform in a stochastic MDP with infinite horizon. In particular, we are interested in (1) the mean squared value error (MSVE) at each timestep, (2) the sample complexity when varying the number of states, and (3) how behavior policies explore and if they converge to the greedy policy.

\paragraph{Evaluation Criteria.}
The MSVE is computed between the true value function of the optimal policy $V^*(s)$ and the learned value function of the current policy $V^{\pi_t}(s)$, i.e.,
\begin{equation}
	\smash{\text{MSVE} = \frac{1}{N}\sum_{i=1}^N{\left(V^*(s_i) - V^{\pi_t}(s)\right)^2}}.
\end{equation}
The value function is computed according to the Bellman expectation equation in matrix form, i.e., $V^\pi = (I - \gamma \prob^\pi)^{-1}\reward^\pi$, where $I$ is the identity matrix, and $\prob^\pi$ and $\reward^\pi$ are the transition and reward functions induced by the policy, respectively \citep{sutton2018reinforcement}. This error indicates how much the learned greedy policy $\pi_t$ deviates from the optimal policy.
\\
Similarly, the sample complexity is defined as the number of timesteps $t$ such that the non-stationary policy $\pi_t$ at time $t$ is not $\varepsilon$-optimal for current state $s_t$, i.e, $V^*(s_t) - V^{\pi_t}(s_t) > \varepsilon$ \citep{strehl2008analysis,dong2020q}.
\\
For the behavior of exploration policies and their empirical convergence, we show how the visitation count changes over time.

\begin{wrapfigure}{l}{0.5\textwidth}
	\begin{center}
		\vspace*{-0.7cm}
		\includegraphics[width=0.97\linewidth]{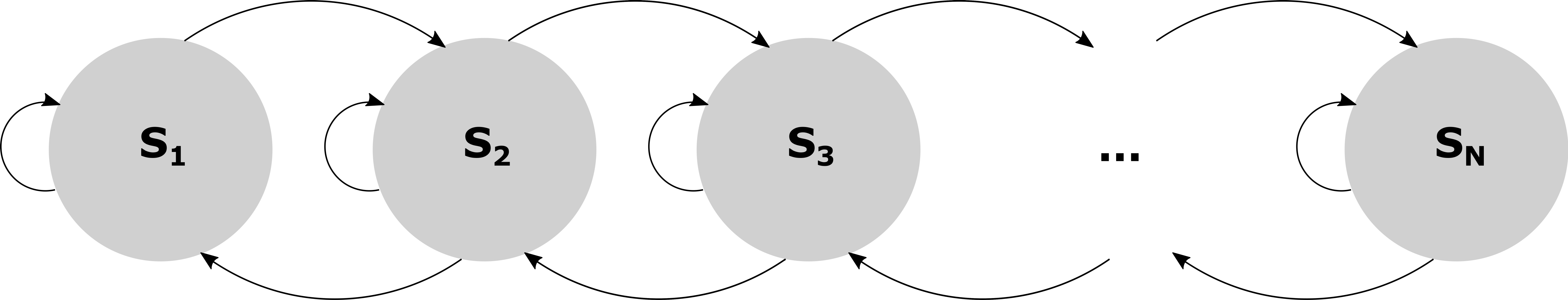}
	\end{center}
	\vspace*{-0.5cm}
	\caption{The ergodic chainworld.}
	\label{fig:chain_env}
	\vspace*{-10pt}
\end{wrapfigure}
\paragraph{MDP Characteristics.}
The MDP, shown in Figure \ref{fig:chain_env}, is a chainworld with $s = 1, \ldots, N$ states, three actions, and stochastic transition defined as follows. 
The first action moves the agent forward with probability $p$, backward otherwise. The second action moves the agent backward with probability $p$, forward otherwise. The third action keeps the agent in the current state with probability $p$, and randomly moves it backward or forward otherwise.
The initial state is the leftmost state, and no state is terminal.
This MDP is ergodic, i.e., all states are transient, positive recurrent, and aperiodic for any deterministic policy.
In our experiments, we set $p = 0.99$.
The reward is $10^{-8}$ for doing ``stay'' in the initial state $s_1$, $1$ for doing ``stay'' in the last state $s_N$, and 0 everywhere else.

Unlike previous MDPs, this has no terminal state, the horizon is infinite, and the state is never reset. Instead, to simulate infinite horizon, the agent explores the environment for one episode of 200,000 steps. 
To comply with the classic sample complexity definition \citep{strehl2008analysis,dong2020q}, we use classic Q-learning without replay memory. We do not compare against bootstrapping algorithms, because they are designed to select a different behavior Q-function at every episode, and this setup has only one episode. Approximate Thompson sampling by \citet{deramo2019exploiting}, instead, selects the Q-function at every step, and it is included in the comparison (without memory as well). All algorithms use zero Q-function initialization. 
Due to the stochasticity of the MDP, we increased the number of random seeds to 50.

\paragraph{Results.}
Figure \ref{fig:chain_count} shows how the algorithms explore over time. Only ours (\ref{fig:chain_count_vv3} and \ref{fig:chain_count_vv2}) and UCB1 (\ref{fig:chain_count_ucb}) explore uniformly, but UCB1 finds the last state (with the reward) later. The auxiliary rewards (\ref{fig:chain_count_mbie}) and approximate Thompson sampling (\ref{fig:chain_count_thomp}) also perform poorly, since the exploration is not uniform and the reward is found only late.  $\epsilon$-greedy exploration (\ref{fig:chain_count_egreedy}), instead, is soon stuck in the local optimum represented by the small reward in the initial state.

\begin{figure}[t]
	\centering
	\includegraphics[width=\linewidth]{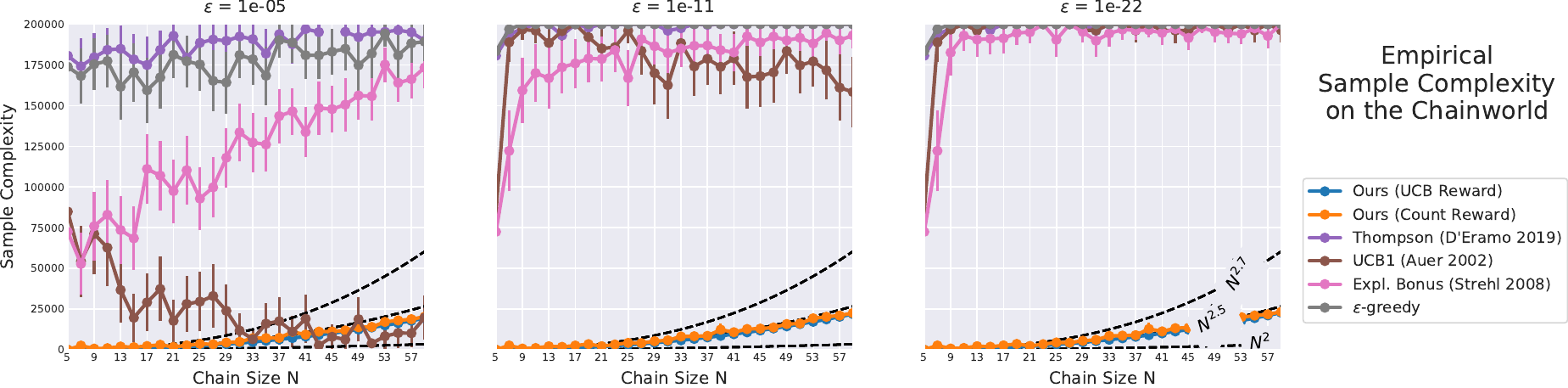}
	\caption{\label{fig:empirical_sample_chain} \textbf{Sample complexity on varying the chain size} $N = 5, 7, \ldots, 59$. Error bars denote 95\% confidence interval. Only the proposed algorithms (blue and orange lines almost overlap) show little sensitivity to $\varepsilon$, as their sample complexity only slightly increases as $\varepsilon$ decreases. By contrast, other algorithms are highly influenced by $\varepsilon$. Their estimate complexity has a large confidence interval, and for smaller $\varepsilon$ they rarely learn an $\varepsilon$-optimal policy.}
\end{figure}

\begin{figure}[t]
	\centering
	\includegraphics[width=\linewidth,trim=0 0em 15.2em 0em, clip]{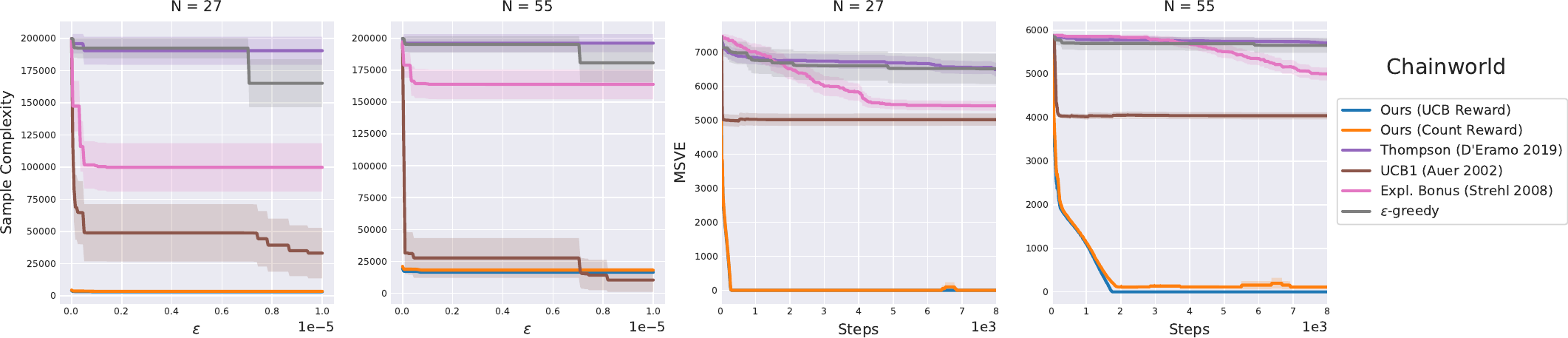}
	\caption{\label{fig:chain_res} \textbf{Sample complexity against $\varepsilon$, and MSVE} over time for 27- and 55-state chainworlds. Shaded areas denote 95\% confidence interval. As shown in Figure \ref{fig:empirical_sample_chain}, the sample complexity of our approach is barely influenced by $\varepsilon$. Even for $\varepsilon = 0$, our algorithms attain low sample complexity, whereas other algorithms complexity is several orders of magnitude higher. Similarly, only our algorithms learn an almost perfect Q-function, achieving an MSVE close to zero.}
\end{figure}

The visitation count also shows that all exploration policies but ours act greedily once the reward is found. For example, UCB1 finds the reward around step 1,000, and after that its visitation count increases only in the last state (and nearby states, because of the stochastic transition). 
This suggests that when these algorithms find the reward placed in the last state they effectively stop exploring other states.
By contrast, our algorithms explore the environment for longer, yielding a more uniform visitation count. This allows the agent to learn the true Q-function for all states, achieving lower sample complexity and MSVE as shown in Figures \ref{fig:empirical_sample_chain} and \ref{fig:chain_res}.

Figure \ref{fig:empirical_sample_chain} shows the sample complexity on varying the chain size over three values of $\varepsilon$. In all three cases, our algorithms scale gracefully with the number of states, suggesting a complexity of $\bigO(N^{2.5})$ and strengthening the findings of Section \ref{sssec:empirical_sample_deep}. By contrast, other algorithms performed poorly. Their sample complexity has a large confidence interval, and as $\varepsilon$ decreases the complexity increases to the point that they rarely learn an $\varepsilon$-optimal policy.

These results are confirmed by Figure \ref{fig:chain_res} (left plots), where only our algorithms attain low sample complexity even for $\varepsilon = 0$. Our algorithms are also the only ones to learn an almost perfect Q-function, with an MSVE close to zero (right plots). As anticipated, these results are explained by the uniform visitation count in Figure \ref{fig:chain_count}. By not converging to the greedy policy too quickly after discovering the reward, the agent keeps visiting old states and propagating to them the information about the reward, thus learning the true optimal Q-function for all states.

\clearpage

\begin{figure}[t]
	\centering
	\includegraphics[width=0.95\linewidth,trim=0 1.1em 15.5em 0em, clip]{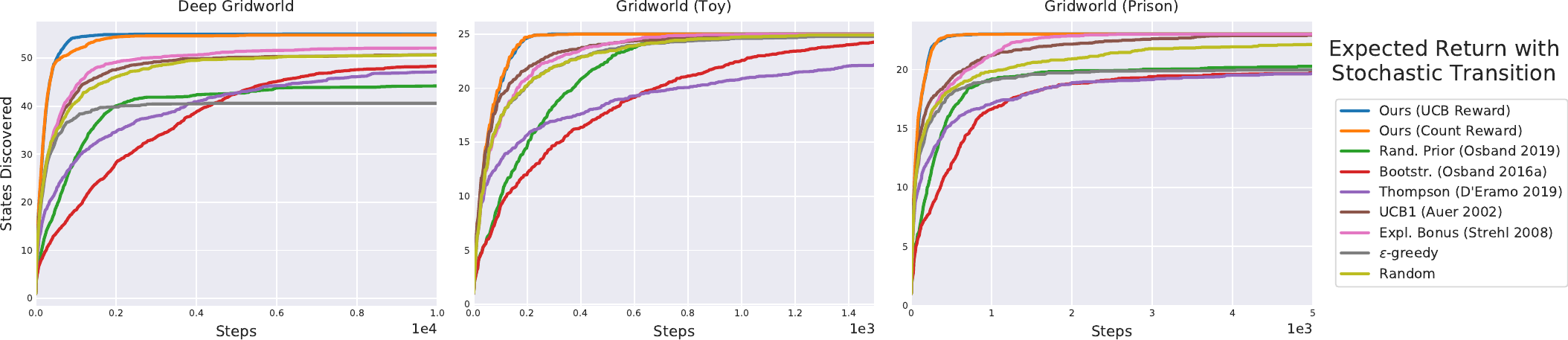}
	\\[0.5em]
	\includegraphics[width=0.95\linewidth,trim=0 0 15.5em 1em, clip]{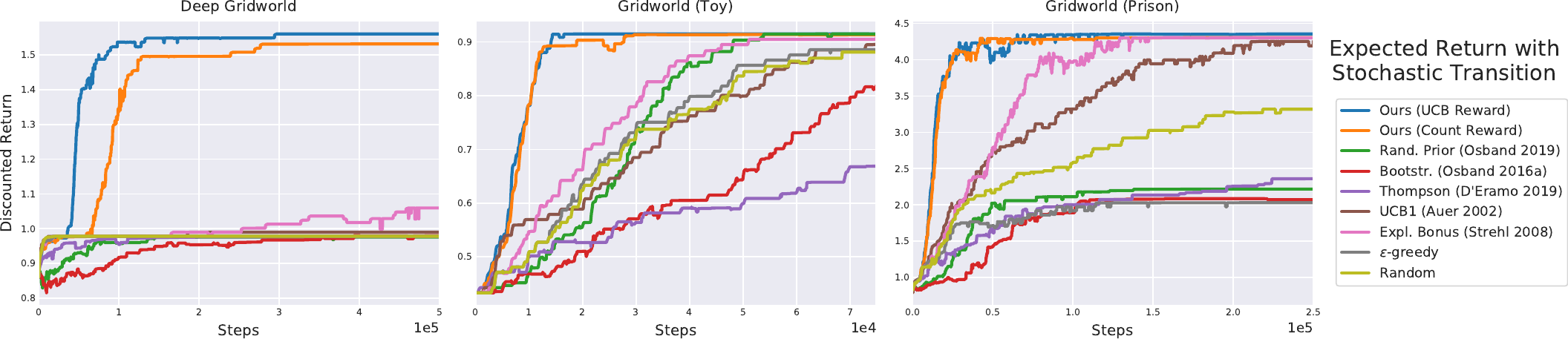}
	\\[0.5em]
	\includegraphics[width=\linewidth]{plot/legend_hor}
	\caption{\label{fig:noisy_res}\textbf{Results with stochastic transition function} on three of the previous MDPs. Once again, only our algorithms visit all states and learn the optimal policy within steps limit in all MDPs, and their performance is barely affected by the stochasticity.}
\end{figure}

\subsubsection{Stochastic Gridworlds}
\label{sssec:stochastic_grid}
As we discussed in Section \ref{ssec:notes}, the visitation rewards for training the W-functions penalize terminal state-action pairs. One may think this would lead to poor exploration in stochastic environments, where the same state-action pair can lead to either terminal or non-terminal states. 
Here, we evaluate all algorithms again on some of the previous environments but this time with stochastic transitions and show that our W-functions still solve the MDPs and outperform existing algorithms. 
Each transition $(s,a)$ has probability $p$ of succeeding, $1 - p/2$ of not moving the agent, and $1 - p/2$ of moving the agent to a random adjacent state. In our experiments, we set $p = 0.9$.  
Due to the stochasticity of transitions, we increased the number of random seeds from 20 to 50.

Figure \ref{fig:noisy_res} shows the number of states discovered and the expected discounted return computed in closed form as $V^\pi(s_0)$, where $V^\pi$ is defined according to the Bellman expectation equation as $V^\pi = (I - \gamma \prob^\pi)^{-1}\reward^\pi$. Similarly to the deterministic setting (Figures \ref{fig:deep_grid_res}, \ref{fig:grid_simple_res}, and \ref{fig:grid_small_res}, leftmost plots), our algorithms (blue and orange) outperform all baseline algorithms, being the only ones to solve all three MPDs within the steps limit\footnote{Because of the stochasticity we decreased the learning rate from 0.5 to 0.1, and increased the learning steps.}. With the UCB reward (blue), our approach always learns the optimal policy. With the count reward (orange), our approach almost always learns the optimal policy, and converges to a distractor only two times (out of 50) in the deep gridworld, and once in the ``prison'' gridworld. However, they both quickly discovered all states, as shown by top plots.
The better performance of the UCB-based reward is due to the optimistic initialization of the W-function. As seen in Section \ref{sssec:chain}, Figure \ref{fig:chain_count_vv3}, this version of the algorithm explores the environment for longer and converges to the greedy policy much later. Therefore, the agent will visit the high-reward state more often, and it is less likely to learn suboptimal policies.
\\Bootstrapping with a prior (green) is not affected by the stochasticity, and it performs as in the deterministic setting (it solves the ``toy'' gridworld but not the other two).
Vanilla bootstrapping (red), instead, is heavily affected by the stochasticity, and its performance is substantially worse compared to the deterministic setting (it cannot learn even the ``toy'' gridworld within steps limit). 
UCB1 (brown) and bonus-based exploration (pink), instead, benefit from the stochasticity. In the deterministic setting they could not solve these MDPs, while here they solve the two gridworlds (even though much later than ours). This improvement is explained by the larger number of visited states (top plots), which denotes an overall better exploration. This is not surprising if we consider that a stochastic transition function naturally helps exploration.

\clearpage

\section{Conclusion and Future Work}
\label{sec:concl}
Effective exploration with sparse rewards is an important challenge in RL, especially when sparsity is combined with the presence of ``distractors'', i.e., rewards that create suboptimal modes of the objective function. 
Classic algorithms relying on dithering exploration typically perform poorly, often converging to poor local optima or not learning at all.
Methods based on immediate counts have strong guarantees but are empirically not sample efficient, while we showed that methods based on intrinsic auxiliary rewards require hand-tuning and are prone to suboptimal behavior. 
In this paper, we presented a novel approach that (1) plans exploration actions far into the future by using a long-term visitation count, and (2) decouples exploration and exploitation by learning a separate function assessing the exploration value of the actions. 
Contrary to existing methods that use models of reward and dynamics, our approach is off-policy and model-free. 
Empirical results showed that the proposed approach outperforms existing methods in environments with sparse and distracting rewards, and suggested that our approach scales gracefully with the size of the environment. 

The proposed approach opens several avenues of research. First, in this work, we focused on empirical results. In the future, we will investigate the theoretical properties of the proposed approach in more detail.
Second, in this work, we considered model-free RL. In the future, we will extend the proposed approach and combine it with model-based RL.
Third, the experimental evaluation focused on identifying the challenges of learning with sparse and distracting rewards. In the future, we will consider more diverse tasks with continuous states and actions, and extend the proposed exploration to actor-critic methods.

\bibliographystyle{plainnat}
\bibliography{my_bib}

\clearpage

\appendix 
\setlength\parindent{0pt}

\section{Experiment Details}
\label{app:tab_details}

Below, we present the pseudocode and the hyperparameters of the algorithms used in Section \ref{sec:eval}. In all of them, we kept two separate Q-tables for the behavior policy $\beta(a|s)$ and the target (greedy) policy $\pi(a|s)$. The former can be either initialized to zero or optimistically, while the latter always to zero. 
With the optimistic initialization, all entries of ${Q}(s,a)$ are set to $r_{\max} / (1 - \gamma)$.
The reason why we do not initialize $Q^\pi(s,a)$ optimistically is that if the agent does not visit all the state-action pairs, and thus never updated the corresponding Q-table entries, it would still have an optimistic belief over some states. The performance of the target policy can therefore be poor until all state-action pairs are visited.
For example, in our experiments in the ``prison'' gridworld, the $\epsilon$-greedy policy was finding some low-value treasures, but during the evaluation the greedy policy was not trying to collect them because it still had an optimistic belief over unvisited empty states. 
\\
In most of the pseudocode, we explicitly distinguish between $Q^\beta(s,a)$ and $Q^\pi(s,a)$. If simply $Q(s,a)$ appears, it means that both Q-tables are updated using the same equation.
\\
In all algorithms, we evaluate $\pi(a|s)$, i.e., the greedy policy over $Q^\pi(s,a)$, every 50 training steps. Each evaluation episode has a horizon 10\% longer than the training one. The learning rate is $\eta = 0.5$ and the discount factor $\gamma = 0.99$. For MDPs with stochastic transition function (Sections \ref{sssec:chain} and \ref{sssec:stochastic_grid}) we used $\eta = 0.1$. 
The $\epsilon$-greedy initially has $\epsilon_0 = 1$, then it decays at step according to $\epsilon_{i+1} = \xi\epsilon_{i}$, where $\xi$ is chosen such that $\epsilon_{end} = 0.1$ when the learning is over.
\\
Finally, we break ties with random selection, i.e., if two or more actions have the same max Q-value the winner is chosen randomly.

\subsection{The Generic Q-Learning Scheme}
\label{app:generic_q}
In all experiments, we used Q-learning with infinite replay memory. Classic Q-learning by \citet{watkins1992q} updates the Q-tables using only the current transition. The so-called \textit{experience replay}, instead, keeps past transitions and uses them multiple times for successive updates. This procedure became popular with deep Q-learning \citep{mnih2013playing} and brought substantial improvement to many RL algorithms.
\\
In our experiments, we followed the setup proposed by \citet{osband2019deep} and used an infinite memory, i.e., we stored all transitions, since it allowed much faster learning than classic Q-learning.
This can also be seen as Dyna-Q \citep{sutton1990integrated} without random memory sampling. 
In our experiments, the size of the domains does not pose any storage issue. 
When storage size is an issue, it is possible to fix the memory size and sample random mini-batches as in DQN \citep{mnih2013playing}. More details in Appendix \ref{app:dyna_vs_q}.
\\
Algorithm \ref{alg:ql} describes the generic scheme of Q-learning with infinite replay memory. TD learning is used on both $Q^\pi(s,a)$ and $Q^\beta(s,a)$, with the only difference being the initialization ($Q^\pi(s,a)$ is always initialized to zero).

\begin{algorithm}[h]
	\caption{\label{alg:ql}Tabular Q-Learning with Replay Memory}
	Initialize ${Q}_0^\beta(s,a)$, ${Q}_0^\pi(s,a), i = 0$
	\\
	While $i < i_{\textsub{budget}}$ do
	\Cycle{
		Reset environment to state $s_1$
		\\
		For $t = 1 ... H$ or until $s_t$ is terminal
		\Cycle{
			Select action according to $a_t \sim \beta(\cdot|s_t)$
			\\
			Transition to $s_{t+1} = \pmodelt$ and receive reward $r_t = \rmodelt$
			\\
			Store tuple $(s_t,a_t,r_t,s_{t+1})$
			\\
			For all tuples $(s,a,s',r)$ in the replay memory 
			\Cycle{
				$\delta(s,a,s') = \begin{cases}
					r_t - Q_{i}(s,a) & \text{if $s$ is terminal}
					\\
					r_t + \gamma\max_{a}{Q}_{i}(s',a) - {Q}_{i}(s,a) & \text{otherwise}
				\end{cases}$
				\\
				${Q}_{i+1}(s,a) \leftarrow {Q}_{i}(s,a) + \eta\delta(s,a,s')$
			}
			Update budget counter: $i \leftarrow i + 1$
		}
	}
\end{algorithm}

Depending on $\beta(a|s)$, we have different exploration strategies, i.e.,
\begin{align}
	\beta(a_t|s_t)  &\sim \text{unif}\{\actionspace\}, & \text{random}
	\\
	\beta(a_t|s_t) & \begin{cases} 
	    = \arg\max_a \left\lbrace Q^\beta(s_t,a) \right\rbrace & \text{with probability $1\!-\!\epsilon$}, 
		\\
		\sim \text{unif}\{\actionspace\} & \text{with probability $\epsilon$},
		\end{cases} & \text{$\epsilon$-greedy}
	\\
	\beta(a_t|s_t) &= \arg\max_a \left\lbrace Q^\beta(s_t,a) + \coeff \sqrt{\frac{2\log \sum_{a_j} n(s_t,a_j)}{n(s_t,a)}} \right\rbrace. & \text{UCB1}
\end{align}

\textbf{Numerical stability and worst-case scenario.} For UCB1 \citep{auer2002finite}, a visitation count $n(s,a)$ is increased after every transition (see Algorithm \ref{alg:ql_aug}, line 7). In classic bandit problems, UCB1 is initialized by executing all actions once, i.e., with $n(s,a) = 1$. In MDPs we cannot do that, i.e., we cannot arbitrarily set the agent in any state and execute all actions, thus $n(s,a) = 0$. Following the W-function bound in Section \ref{ssec:vv_init}, we always add +1 inside the logarithm, and bound the square root to $(Q_{\max} - Q_{\min}) / \coeff + \sqrt{2\log|\actionspace|}$ when $n(s,a) = 0$. This correspond to the case where all actions but $\bar{a}$ has been executed once, and enforces the policy to choose $\bar{a}$. In our experiments, we set $Q_{\max} = r_{\max} / (1 - \gamma)$ and $Q_{\min} = 0$.

\subsection{Q-Learning with Augmented Reward}
This version follows the same scheme of Algorithm \ref{alg:ql} with $\epsilon$-greedy exploration. The only difference is that the reward used to train the behavior policy is augmented with the exploration bonus proposed by \citet{strehl2008analysis}. 
In our experiments, $\alpha = 0.1$, as used by \citet{strehl2008analysis} and \citet{bellemare2016unifying}.

\begin{algorithm}[h!]
	\caption{\label{alg:ql_aug}Tabular Q-Learning with Replay Memory and Augmented Reward}
	\setstretch{1.15}
	Initialize ${Q}_0^\beta(s,a)$, ${Q}_0^\pi(s,a) = 0, \textcolor{red}{n(s,a) = 0}, i = 0$
	\\
	While $i < i_{\textsub{budget}}$ do
	\Cycle{
		Reset environment to state $s_1$
		\\
		For $t = 1 ... H$ or until $s_t$ is terminal
		\Cycle{
			Select action according to the $\epsilon$-greedy policy of $Q_i^\beta(s,a)$
			\\
			Transition to $s_{t+1} = \pmodelt$ and receive reward $r_t = \rmodelt$
			\\
			\textcolor{red}{Update visitation count: $n(s_t,a_t) \leftarrow n(s_t,a_t) + 1$}
			\\
			Store tuple $(s_t,a_t,r_t,s_{t+1})$
			\\
			For all tuples $(s,a,s',r)$ in the replay memory 
			\Cycle{
				$\delta^\pi(s,a,s') = \begin{cases}
				r_t - {Q}^\pi_{i}(s,a) & \text{if $s$ is terminal}
				\\
				r_t + \gamma\max_{a}{Q}^\pi_{i}(s',a) - {Q}^\pi_{i}(s,a) & \text{otherwise}
				\end{cases}$
				\\
				${Q}^\pi_{i+1}(s,a) \leftarrow {Q}^\pi_{i}(s,a) + \eta\delta^\pi(s,a,s')$
				\\
				\textcolor{red}{Augment reward: $r^+_t = r_t + {\alpha}\:{{n(s_t,a_t)}^{-1/2}}$}
				\\
				$\delta^\beta(s,a,s') = \begin{cases}
				\textcolor{red}{r^+_t} - {Q}^\beta_{i}(s,a) & \text{if $s$ is terminal}
				\\
				\textcolor{red}{r^+_t} + \gamma\max_{a}{Q}_{i}^\beta(s',a) - {Q}^\beta_{i}(s,a) & \text{otherwise}
				\end{cases}$
				\\
				${Q}^\beta_{i+1}(s,a) \leftarrow {Q}^\beta_{i}(s,a) + \eta\delta^\beta(s,a,s')$
			}
			Update budget counter: $i \leftarrow i + 1$
		}
	}
\end{algorithm}

\subsection{Q-Learning with Visitation Value}
Algorithms \ref{alg:ql_vv3} and \ref{alg:ql_vv2} describe the novel method proposed in this paper. In our experiments $\coeff = r_{\max} / (1 - \gamma)$ and $\gamma_w = 0.99$. For the chainworld in Section \ref{sssec:chain} we set $\gamma_w = 0.999$. With infinite horizon, in fact, this yielded better results by allowing the agent to explore for longer. For the stochastic gridworld in Section \ref{sssec:stochastic_grid} we set $\gamma_w = 0.9$. As discussed in the results, in fact, a stochastic transition function naturally improves exploration, thus a smaller visitation discount was sufficient and yielded better results.
\\
In Algorithm \ref{alg:ql_vv3}, $W^\beta_{\textsub{ucb}}(s,a)$ is initialized as Eq. \eqref{eq:vv3_init}. In Algorithm \ref{alg:ql_vv2}, line 6, we bound the square root to Eq. \eqref{eq:vv2_condition2} when $n(s,a) = 0$. In our experiments, with $Q_{\max} = r_{\max} / (1 - \gamma)$ and $Q_{\min} = 0$.

\begin{algorithm}[h!]
	\caption{\label{alg:ql_vv3}Tabular Q-Learning with Replay Memory and Visit. Value (UCB)}
	Initialize ${Q}_0^\beta(s,a)$, ${Q}_0^\pi(s,a) = 0, \textcolor{red}{W^\beta_{\textsub{ucb},0}(s,a), n(s,a) = 0}, i = 0$
	\\
	While $i < i_{\textsub{budget}}$ do
	\Cycle{
		Reset environment to state $s_1$
		\\
		For $t = 1 ... H$ or until $s_t$ is terminal
		\Cycle{
			Select action according to $\textcolor{red}{\hspace*{6pt}\beta(a_t|s_t) = \arg\max_a \left\lbrace Q_i^\beta(s_t,a) + \coeff (1 - \gamma_w) W^\beta_{\textsub{ucb},i}(s_t,a)\right\rbrace}$
			\\
			Transition to $s_{t+1} = \pmodelt$ and receive reward $r_t = \rmodelt$
			\\
			\textcolor{red}{Update visitation count: $n(s_t,a_t) \leftarrow n(s_t,a_t) + 1$}
			\\
			Store tuple $(s_t,a_t,r_t,s_{t+1})$
			\\
			For all tuples $(s,a,s',r)$ in the replay memory 
			\Cycle{
				$\delta(s,a,s') = \begin{cases}
				r_t - {Q}_{i}(s,a) & \text{if $s$ is terminal}
				\\
				r_t + \gamma\max_{a}{Q}_{i}(s',a) - {Q}_{i}(s,a) & \text{otherwise}
				\end{cases}$
				\\
				${Q}_{i+1}(s,a) \leftarrow {Q}_{i}(s,a) + \eta\delta(s,a,s')$
				\\
				\textcolor{red}{Compute visitation reward according to Eq. \eqref{eq:vv3_rwd}}
				\\
				\textcolor{red}{$\delta^W(s,a,s') = \begin{cases}
				{r^W_t} - W^\beta_{\textsub{ucb},i}(s,a) & \text{if $s$ is terminal}
				\\
				{r^W_t} + \gamma\max_{a}W^\beta_{\textsub{ucb},i}(s',a) - W^\beta_{\textsub{ucb},i}(s,a) & \text{otherwise}
				\end{cases}$}
				\\
				\textcolor{red}{${W}^\beta_{\textsub{ucb},i+1}(s,a) \leftarrow {W}^\beta_{\textsub{ucb},i}(s,a) + \eta\delta^W(s,a,s')$}
			}
			Update budget counter: $i \leftarrow i + 1$
		}
	}
\end{algorithm}

\begin{algorithm}[h!]
	\caption{\label{alg:ql_vv2}Tabular Q-Learning with Replay Memory and Visit. Value (Count)}
	Initialize ${Q}_0^\beta(s,a)$, ${Q}_0^\pi(s,a) = 0, \textcolor{red}{n(s,a) = 0, W^\beta_{\textsub{n},0}(s,a) = 0}, i = 0$
	\\
	While $i < i_{\textsub{budget}}$ do
	\Cycle{
		Reset environment to state $s_1$
		\\
		For $t = 1 ... H$ or until $s_t$ is terminal
		\Cycle{
			\textcolor{red}{Compute pseudocount: $\hat n(s_{t},a) = (1 - \gamma_w) W^\beta_{\textsub{n},i}(s_{t},a)$}
			\\
			Select action according to $\textcolor{red}{\beta(a_t|s_t) = \arg\max_a \big\lbrace Q^\beta_i(s_t,a) + \coeff {\sqrt{\frac{2\log \sum_{a_j} \hat n(s_t,a_j)}{\hat n(s_t,a)}}} \big\rbrace
			}$
			\\
			Transition to $s_{t+1} = \pmodelt$ and receive reward $r_t = \rmodelt$
			\\
			\textcolor{red}{Update visitation count: $n(s_t,a_t) \leftarrow n(s_t,a_t) + 1$}
			\\
			Store tuple $(s_t,a_t,r_t,s_{t+1})$
			\\
			For all tuples $(s,a,s',r)$ in the replay memory 
			\Cycle{
				$\delta(s,a,s') = \begin{cases}
				r_t - {Q}_{i}(s,a) & \text{if $s$ is terminal}
				\\
				r_t + \gamma\max_{a}{Q}_{i}(s',a) - {Q}_{i}(s,a) & \text{otherwise}
				\end{cases}$
				\\
				${Q}_{i+1}(s,a) \leftarrow {Q}_{i}(s,a) + \eta\delta(s,a,s')$
				\\
				\textcolor{red}{Compute visitation reward according to Eq. \eqref{eq:vv2_rwd}}
				\\
				\textcolor{red}{$\delta^W(s,a,s') = \begin{cases}
					{r^W_t} - W^\beta_{\textsub{n},i}(s,a) & \text{if $s$ is terminal}
					\\
					{r^W_t} + \gamma\min_{a}W^\beta_{\textsub{n},i}(s',a) - W^\beta_{\textsub{n},i}(s,a) & \text{otherwise}
					\end{cases}$}
				\\
				\textcolor{red}{${W}^\beta_{\textsub{n},i+1}(s,a) \leftarrow {W}^\beta_{\textsub{n},i}(s,a) + \eta\delta^W(s,a,s')$}
			}
			Update budget counter: $i \leftarrow i + 1$
		}
	}
\end{algorithm}

\clearpage

\subsection{Bootstrapped Q-Learning}
Bootstrap algorithms have an ensemble of Q-tables $\{Q^b(s,a)\}_{b = 1 \ldots B}$. The behavior policy is greedy over one Q-table, randomly chosen from the ensemble either at the beginning of the episode or at every step.
\\
The behavior Q-tables are initialized randomly. After initializing each $Q^b(s,a)$ either optimistically or to zero, we add random noise drawn from a Gaussian $\gaussian(0,1)$.
Then, each one is trained on a random mini-batch from the replay memory. In our experiments, we used batches of size 1,024 and an ensemble of $B = 10$ behavior Q-tables.
The target Q-table is still updated as in previous algorithms, i.e., using the full memory.

\begin{algorithm}[h!]
	\caption{\label{alg:ql_boot}Tabular Bootstrapped Q-Learning with Replay Memory}
	\setstretch{1.15}
	Initialize ${Q}_{0}^{b}(s,a)$ for $b = 1 \ldots B$, ${Q}_0^\pi(s,a) = 0, i = 0$
	\\
	While $i < i_{\textsub{budget}}$ do
	\Cycle{
		Reset environment to state $s_1$
		\\
		\textcolor{red}{Select random Q-table: $Q_i^\beta(s,a) \sim \text{unif}\{{Q}_{i}^{1}(s,a), \ldots, {Q}_{i}^{B}(s,a)\}$}
		\\
		For $t = 1 ... H$ or until $s_t$ is terminal
		\Cycle{
			\textcolor{red}{Select action according to the $\epsilon$-greedy policy of $Q_i^\beta(s,a)$}
			\\
			Transition to $s_{t+1} = \pmodelt$ and receive reward $r_t = \rmodelt$
			\\
			Store tuple $(s_t,a_t,r_t,s_{t+1})$
			\\
			For all tuples $(s,a,s',r)$ in the replay memory 
			\Cycle{
				$\delta^\pi(s,a,s') = \begin{cases}
				r_t - {Q}^\pi_{i}(s,a) & \text{if $s$ is terminal}
				\\
				r_t + \gamma\max_{a}{Q}^\pi_{i}(s',a) - {Q}^\pi_{i}(s,a) & \text{otherwise}
				\end{cases}$
				\\
				${Q}^\pi_{i+1}(s,a) \leftarrow {Q}^\pi_{i}(s,a) + \eta\delta^\pi(s,a,s')$
			}
			\textcolor{red}{For all behavior Q-tables $Q_i^b(s,a)$}
			\Cycle{
				\textcolor{red}{Sample mini-batch from the replay memory}
				\\
				\textcolor{red}{For all tuples $(s,a,s',r)$ in the mini-batch}
				\Cycle{
					\textcolor{red}{$\delta^b(s,a,s') = \begin{cases}
					\textcolor{red}{r_t} - {Q}^b_{i}(s,a) & \text{if $s$ is terminal}
					\\
					{r_t} + \gamma\max_{a}{Q}_{i}^b(s',a) - {Q}^b_{i}(s,a) & \text{otherwise}
					\end{cases}$}
					\\
					\textcolor{red}{${Q}^b_{i+1}(s,a) \leftarrow {Q}^b_{i}(s,a) + \eta\delta^b(s,a,s')$}
				}
			}
			Update budget counter: $i \leftarrow i + 1$
		}
	}
\end{algorithm}

The above pseudocode defines the generic bootstrap approach proposed by \citet{osband2016deep}. In Section \ref{sec:eval} we also compared to two slightly different versions. The first uses approximate Thompson sampling and randomly selects the behavior Q-table at every step instead of at the beginning of the episode \citep{deramo2019exploiting}. The second keeps the sampling at the beginning of the episode, but further regularizes the TD error \citep{osband2018randomized,osband2019deep}. The regularization is the squared $\ell_2$-norm of the distance of the Q-tables from ``prior Q-tables'' $Q^p(s,a)$, resulting in the following regularized TD update
\begin{equation}
\delta^b_{reg}(s,a,s') = \delta^b(s,a,s') + {\nu} \left(Q^p(s,a) - Q^b(s,a)\right),
\end{equation}
wehere $\nu$ is the regularization coefficient which we set to $\nu = 0.1$ (in the original paper $\nu = 1$, but dividing it by ten worked best in our experiments).

In theory, $Q^p(s,a)$ should be drawn from a distribution. In practice, the same authors fix it at the beginning of the learning, and for the deep sea domain they set it to zero. This configuration also worked best in our experiments.

\clearpage

\paragraph{Horizons.} All environments in Section \ref{sec:eval} are infinite horizon MDPs, except for the deep sea which has a finite horizon equal to its depth. However, for practical reasons, we end the episode after $H$ steps and reset the agent to the initial state. 
Table \ref{tab:horizons} summarizes the training horizon $H$ for each environment, as well as the steps needed for the optimal policy to find the highest reward. 
Also, recall that an episode can end prematurely if a terminal state (e.g., a reward state) is reached.
\\
Finally, notice that the agent receives the reward on state-action transition, i.e., it needs to execute an additional action in the state with a treasure to receive the reward. This is why, for instance, the agent needs 9 steps instead of 8 to be rewarded in the $5\times5$ gridworlds.

\begin{table}[h] 
	\renewcommand{\arraystretch}{1.4} 
	\caption{\label{tab:horizons}Time horizons $H$ for the tabular MDPs presented in Section \ref{sec:eval}.} 
	\centering 
	\begin{tabular}{@{\extracolsep{\fill}}|l|c|c|c|c|c|c|} 
		\hline & Deep Sea & \hspace*{1.5pt} Taxi \hspace*{1.5pt} & Deep Grid. & Grid. (Toy) & Grid. (Prison) & Grid. (Wall) 
		\\ 
		\hline Optimal & Depth & 29 & 11 & 9 & 9 & 135
		\\ 
		\hline Short H. & Depth & 33 & 55 & 11 & 11 & 330
		\\
		\hline Long H. & Depth & 66 & 110 & 22 & 22 & 660
		\\
		\hline Stochastic & - & - & 55 & 15 & 25 & -
		\\
		\hline
	\end{tabular} 
\end{table}

\subsection{Infinite Memory vs No Memory}
\label{app:dyna_vs_q}
Except for the chainworld in Section \ref{sssec:chain}, the evaluation in Section \ref{sec:eval} was conducted with Q-learning with infinite replay memory, following the same setup of \citet{osband2019deep} as described in Section~\ref{app:generic_q}.
Here, we present additional results without replay memory, i.e., using classic Q-learning by \citet{watkins1992q} which updates the Q-tables using only the current transition. We show that the proposed method benefits the most from the replay memory, but can learn even without it while other algorithms cannot.
We compare $\epsilon$-greedy exploration, our proposed visitation-value-based exploration, and bootstrapped exploration. Nevertheless, the latter needs a replay memory to randomize the Q-tables update, as each Q-table is updated using different random samples. Thus, for bootstrapped Q-learning, we compare the use of finite memory and small batches to infinite memory and large batches. 
\\
Algorithm \ref{alg:ql_classic} describes the generic scheme of Q-learning without replay memory. The only difference from Algorithm \ref{alg:ql} is the absence of the loop over the memory. The same modification is done to Algorithms \ref{alg:ql_vv2} and \ref{alg:ql_boot} to derive the version of our algorithms without infinite memory. Bootstrapped Q-learning is the same as Algorithm \ref{alg:ql_vv3} but with finite memory. The finite memory keeps at most $20H$ steps, where $H$ is the episode horizon, and uses mini-batches of 32 samples instead of 1,024.

\begin{algorithm}[h!]
	\caption{\label{alg:ql_classic}Classic Tabular Q-Learning}
	\setstretch{1.15}
	Initialize ${Q}_0^\beta(s,a)$, ${Q}_0^\pi(s,a), i = 0$
	\\
	While $i < i_{\textsub{budget}}$ do
	\Cycle{
		Reset environment to state $s_1$
		\\
		For $t = 1 ... H$ or until $s_t$ is terminal
		\Cycle{
			Select action according to $a_t \sim \beta(\cdot|s_t)$
			\\
			Transition to $s_{t+1} = \pmodelt$ and receive reward $r_t = \rmodelt$
			\\
			Store tuple $(s_t,a_t,r_t,s_{t+1})$
			\\
			$\delta(s_t,a_t,s_{t+1}) = \begin{cases}
			r_t - Q_{i}(s_t,a_t) & \text{if $s_t$ is terminal}
			\\
			r_t + \gamma\max_{a}{Q}_{i}(s_{t+1},a) - {Q}_{i}(s_t,a_t) & \text{otherwise}
			\end{cases}$
			\\
			${Q}_{i+1}(s_t,a_t) \leftarrow {Q}_{i}(s_t,a_t) + \eta\delta(s_t,a_t,s_{t+1})$
			\\
			Update budget counter: $i \leftarrow i + 1$
		}
	}
\end{algorithm}

\paragraph{Results.}
\begin{figure}[t]
	\centering
	\includegraphics[width=\linewidth]{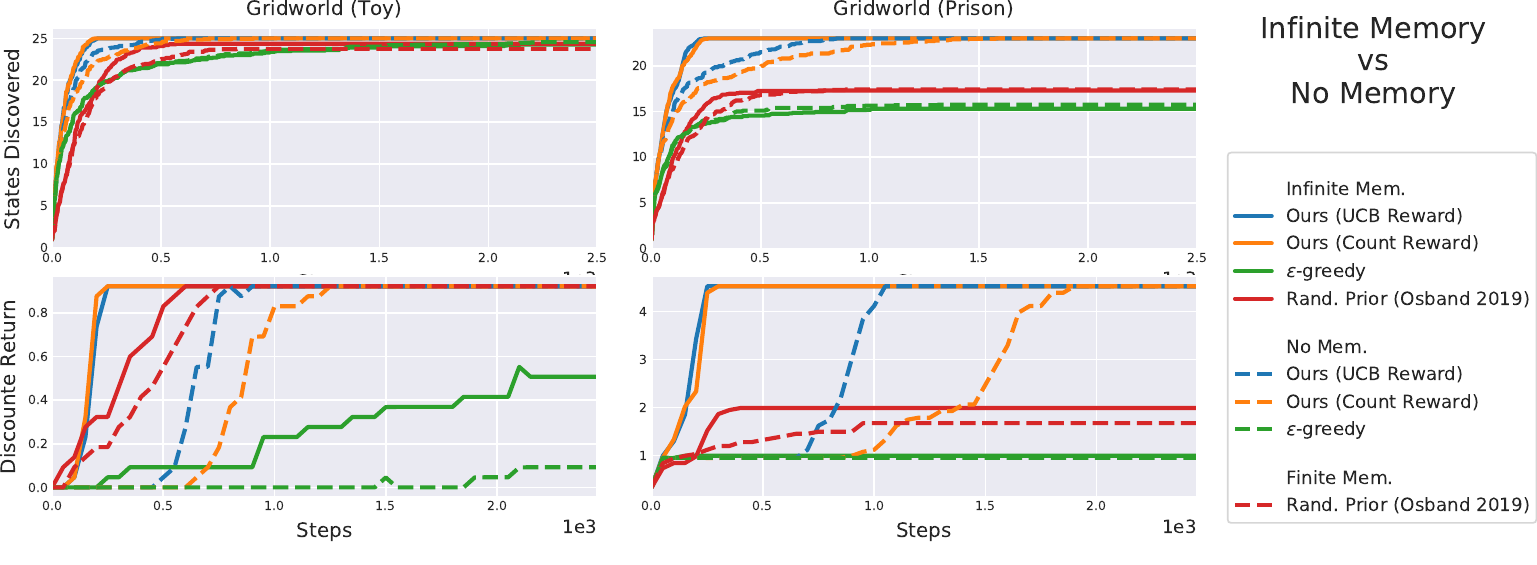}
	\caption{\label{fig:inf_vs_fin} Comparison of the proposed exploration against $\epsilon$-greedy and bootstrapped exploration with and without replay memory. Each line denotes the average over 20 seeds. Only exploration based on the visitation value always learns in both domains, i.e., both with and without local optima (distractor rewards), regardless of the memory.}
\end{figure}

Figure \ref{fig:inf_vs_fin} shows the results on two gridworlds. The ``toy'' one has a single reward, while the ``prison'' one has also distractors. We refer to Section \ref{ssec:eval_part1} for their full description.
In the first domain, all algorithms perform well except for $\epsilon$-greedy exploration (green), which missed the reward in some trials. The use of infinite memory helped all algorithms, allowing bootstrap and visitation-value-based to learn substantially faster. For example on average, without replay memory our algorithms (blue and orange) converged in 1,000 steps, while with replay memory they took only 250 steps, i.e., $\times 4$ times faster. Bootstrapped Q-learning (red), performance does not change substantially with finite and infinite memory.

However, in the second domain only visitation-value-based exploration always learns, regardless of the memory. As already discussed in Section \ref{ssec:eval_part1}, bootstrap performs poorly due to distractor rewards, and the infinite memory does not help it.
For visitation-value-based exploration, the speed-up gained from the infinite memory is even larger than before, due to the higher complexity of the domain. In this case in fact, with infinite memory, it learns $\times 4$ and $\times 8$ times faster than without memory (blue and orange, respectively).

This evaluation shows that the proposed method benefits the most from the replay memory, but can learn even without it while other algorithms cannot.

\clearpage

\subsection{Recap Tables}
Here, we report tables summarizing the results of Section \ref{ssec:eval_part1} in terms of discovery (percentage of states discovered during learning) and success (number of times the algorithm learned the optimal policy within steps limit). These are extended versions of Table \ref{tab:recap}, which reported results only for the ``zero initialization short horizon'' scenario.

\begin{table}[h]
	\centering
	\caption{Deep sea results recap.}
	\setstretch{1.15}
	\begin{tabular}{ c | c | c c }
		& Algorithm & Discovery (\%) & Success (\%) \\
		\hline
		& \textbf{Ours (UCB Reward)} & $\mathbf{100 \pm 0}$ & $\mathbf{100 \pm 0}$ \\
		& \textbf{Ours (Count Reward)} & $\mathbf{100 \pm 0}$ & $\mathbf{100 \pm 0}$ \\
		& Rand. Prior (Osband 2019) & $99.90 \pm 0.01$ & $100 \pm 0$ \\
		& Bootstr. (Osband 2016a) & $99.77 \pm 0.05$ & $100 \pm 0$ \\
		Zero Init. & Bootstr. (D'Eramo 2019) & $63.25 \pm 3.31$ & $0 \pm 0$ \\
		& UCB1 (Auer 2002) & $55.72 \pm 0.34$ & $0 \pm 0$ \\
		& Expl. Bonus (Strehl 2008) & $85.65 \pm 1.0$ & $0 \pm 0$ \\
		& $\epsilon$-greedy & $57.74 \pm 1.11$ & $0 \pm 0$ \\
		& Random & $58.59 \pm 1.35$ & $0 \pm 0$ \\
		\hline
		& \textbf{Ours (UCB Reward)} & $\mathbf{100 \pm 0}$ & $\mathbf{100 \pm 0}$\\
		& \textbf{Ours (Count Reward)} & $\mathbf{100 \pm 0}$ & $\mathbf{100 \pm 0}$ \\
		& Rand. Prior (Osband 2019) & ${100 \pm 0}$ & ${100 \pm 0}$ \\
		& Bootstr. (Osband 2016a) & ${100 \pm 0}$ & ${100 \pm 0}$ \\
		Opt. Init. & Bootstr. (D'Eramo 2019) & ${100 \pm 0}$ & ${100 \pm 0}$ \\
		& UCB1 (Auer 2002) & $76.85 \pm 0.3$ & $0 \pm 0$ \\
		& Expl. Bonus (Strehl 2008) & $98.79 \pm 0.21$ & $0 \pm 0$ \\
		& $\epsilon$-greedy & $98.79 \pm 0.20$ & $0 \pm 0$ \\
		& Random & $58.59 \pm 1.35$ & $0 \pm 0$ \\
	\end{tabular}
\end{table}

\begin{table}[h]
	\centering
	\caption{Taxi results recap.}
	\setstretch{1.15}
	\begin{tabular}{ c | c | c c }
		& Algorithm & Discovery (\%) & Success (\%) \\
		\hline
		& \textbf{Ours (UCB Reward)} & $\mathbf{100 \pm 0}$ & $\mathbf{100 \pm 0}$ \\
		& \textbf{Ours (Count Reward)} & $\mathbf{100 \pm 0}$ & $\mathbf{100 \pm 0}$ \\
		& Rand. Prior (Osband 2019) & $69.60 \pm 2.96$ & $13.07 \pm 2.96$ \\
		& Bootstr. (Osband 2016a) & $52.44 \pm 7.55$ & $18.77 \pm 9.24$ \\
		Zero Init. \& Short Hor. & Bootstr. (D'Eramo 2019) & $22.44 \pm 1.81$ & $1.9 \pm 1.47$ \\
		& UCB1 (Auer 2002) & $31.17 \pm 0.70$ & $1.53 \pm 1.37$ \\
		& Expl. Bonus (Strehl 2008) & $74.62 \pm 2.24$ & $17.6 \pm 2.56$ \\
		& $\epsilon$-greedy & $29.64 \pm 0.98$ & $1.92 \pm 1.49$ \\
		& Random & $29.56 \pm 0.98$ & $1.92 \pm 1.49$ \\
		\hline
		& \textbf{Ours (UCB Reward)} & $\mathbf{100 \pm 0}$ & $\mathbf{100 \pm 0}$ \\
		& \textbf{Ours (Count Reward)} & $\mathbf{100 \pm 0}$ & $\mathbf{100 \pm 0}$ \\
		& Rand. Prior (Osband 2019) & $100 \pm 0$ & $100 \pm 0$ \\
		& Bootstr. (Osband 2016a) & $100 \pm 0$ & $100 \pm 0$ \\
		Opt. Init. \& Short Hor. & Bootstr. (D'Eramo 2019) & $100 \pm 0$ & $100 \pm 0$ \\
		& UCB1 (Auer 2002) & $99.94 \pm 0.08$ & $100 \pm 0$ \\
		& Expl. Bonus (Strehl 2008) & $100 \pm 0$ & $100 \pm 0$ \\
		& $\epsilon$-greedy & $100 \pm 0$ & $100 \pm 0$ \\
		& Random & $29.56 \pm 0.98$ & $1.92 \pm 1.49$ \\
		\hline
		& \textbf{Ours (UCB Reward)} & $\mathbf{100 \pm 0}$ & $\mathbf{100 \pm 0}$ \\
		& \textbf{Ours (Count Reward)} & $\mathbf{100 \pm 0}$ & $\mathbf{100 \pm 0}$ \\
		& Rand. Prior (Osband 2019) & $76.38 \pm 2.31$ & $15.69 \pm 2.93$ \\
		& Bootstr. (Osband 2016a) & $53.80 \pm 6.63$ & $29.85 \pm 12.72$ \\
		Zero Init. \& Long Hor. & Bootstr. (D'Eramo 2019) & $36.34 \pm 3.16$ & $6.54 \pm 2.62$ \\
		& UCB1 (Auer 2002) & $49.66 \pm 2.45$ & $8.35 \pm 1.32$ \\
		& Expl. Bonus (Strehl 2008) & $92.83 \pm 2.26$ & $76.18 \pm 16.09$ \\
		& $\epsilon$-greedy & $42.26 \pm 2.52$ & $7.66 \pm 0.03$ \\
		& Random & $55.89 \pm 0.98$ & $2.06 \pm 1.49$ \\
		\hline
		& \textbf{Ours (UCB Reward)} & $\mathbf{100 \pm 0}$ & $\mathbf{100 \pm 0}$ \\
		& \textbf{Ours (Count Reward)} & $\mathbf{100 \pm 0}$ & $\mathbf{100 \pm 0}$ \\
		& Rand. Prior (Osband 2019) & $100 \pm 0$ & $100 \pm 0$ \\
		& Bootstr. (Osband 2016a) & $100 \pm 0$ & $100 \pm 0$ \\
		Opt. Init. \& Long Hor. & Bootstr. (D'Eramo 2019) & $95.45 \pm 0$ & $100 \pm 0$ \\
		& UCB1 (Auer 2002) & $98.74 \pm 0.09$ & $100 \pm 0$ \\
		& Expl. Bonus (Strehl 2008) & $100 \pm 0$ & $100 \pm 0$ \\
		& $\epsilon$-greedy & $100 \pm 0$ & $100 \pm 0$ \\
		& Random & $55.89 \pm 0.98$ & $2.06 \pm 1.49$ \\
	\end{tabular}
\end{table}

\begin{table}[h]
\centering
\setstretch{1.15}
\caption{Deep gridworld results recap.}
\begin{tabular}{ c | c | c c }
 & Algorithm & Discovery (\%) & Success (\%) \\
 \hline
 & \textbf{Ours (UCB Reward)} & $\mathbf{100 \pm 0}$ & $\mathbf{100 \pm 0}$ \\
 & \textbf{Ours (Count Reward)} & $\mathbf{100 \pm 0}$ & $\mathbf{100 \pm 0}$ \\
 & Rand. Prior (Osband 2019) & $75 \pm 9.39$ & $59.03 \pm 4.2$ \\
 & Bootstr. (Osband 2016a) & $60.82 \pm 11.78$ & $63.35 \pm 6.89$ \\
 Zero Init. \& Short Hor. & Bootstr. (D'Eramo 2019) & $63.73 \pm 10.35$ & $56.85 \pm 0.06$ \\
 & UCB1 (Auer 2002) & $92.18 \pm 0.36$ & $56.88 \pm 0$ \\
 & Expl. Bonus (Strehl 2008) & $95.45 \pm 1.57$ & $69.81 \pm 8.84$ \\
 & $\epsilon$-greedy & $74.36 \pm 4.42$ & $56.88 \pm 0$ \\
 & Random & $92.45 \pm 0.64$ & $56.88 \pm 0$ \\
 \hline
 & \textbf{Ours (UCB Reward)} & $\mathbf{100 \pm 0}$ & $\mathbf{100 \pm 0}$ \\
 & \textbf{Ours (Count Reward)} & $\mathbf{100 \pm 0}$ & $\mathbf{100 \pm 0}$ \\
 & Rand. Prior (Osband 2019) & $100 \pm 0$ & $100 \pm 0$ \\
 & Bootstr. (Osband 2016a) & $100 \pm 0$ & $100 \pm 0$ \\
 Opt. Init. \& Short Hor. & Bootstr. (D'Eramo 2019) & $100 \pm 0$ & $100 \pm 0$ \\
 & UCB1 (Auer 2002) & $100 \pm 0$ & $100 \pm 0$ \\
 & Expl. Bonus (Strehl 2008) & $100 \pm 0$ & $100 \pm 0$ \\
 & $\epsilon$-greedy & $100 \pm 0$ & $100 \pm 0$ \\
 & Random & $92.45 \pm 0.64$ & $56.88 \pm 0$ \\
 \hline
 & \textbf{Ours (UCB Reward)} & $\mathbf{100 \pm 0}$ & $\mathbf{100 \pm 0}$  \\
 & \textbf{Ours (Count Reward)} & $\mathbf{100 \pm 0}$ & $\mathbf{100 \pm 0}$ \\
 & Rand. Prior (Osband 2019) & $78.18 \pm 8.97$ & $56.88 \pm 0$ \\
 & Bootstr. (Osband 2016a) & $77.64 \pm 10.02$ & $69.81 \pm 8.84$ \\
 Zero Init. \& Long Hor. & Bootstr. (D'Eramo 2019) & $71.64 \pm 10.35$ & $56.85 \pm 0.06$ \\
 & UCB1 (Auer 2002) & $95.54 \pm 0.46$ & $56.4 \pm 3.12$ \\
 & Expl. Bonus (Strehl 2008) & $95 \pm 1.66$ & $65.5 \pm 7.71$ \\
 & $\epsilon$-greedy & $76.45 \pm 4.25$ & $56.88 \pm 0$ \\
 & Random & $92.09 \pm 0.69$ & $56.88 \pm 0$ \\
 \hline
 & \textbf{Ours (UCB Reward)} & $\mathbf{100 \pm 0}$ & $\mathbf{100 \pm 0}$ \\
 & \textbf{Ours (Count Reward)} & $\mathbf{100 \pm 0}$ & $\mathbf{100 \pm 0}$ \\
 & Rand. Prior (Osband 2019) & $100 \pm 0$ & $100 \pm 0$ \\
 & Bootstr. (Osband 2016a) & $100 \pm 0$ & $100 \pm 0$ \\
 Opt. Init. \& Long Hor. & Bootstr. (D'Eramo 2019) & $100 \pm 0$ & $100 \pm 0$ \\
 & UCB1 (Auer 2002) & $100 \pm 0$ & $100 \pm 0$ \\
 & Expl. Bonus (Strehl 2008) & $100 \pm 0$ & $100 \pm 0$ \\
 & $\epsilon$-greedy & $100 \pm 0$ & $100 \pm 0$ \\
 & Random & $92.09 \pm 0.69$ & $56.88 \pm 0$ \\
\end{tabular}
\end{table}

\begin{table}[h]
\centering
\caption{Gridworld (toy) results recap.}
\setstretch{1.15}
\begin{tabular}{ c | c | c c }
 & Algorithm & Discovery (\%) & Success (\%) \\
 \hline
 & \textbf{Ours (UCB Reward)} & $\mathbf{100 \pm 0}$ & $\mathbf{100 \pm 0}$ \\
 & \textbf{Ours (Count Reward)} & $\mathbf{100 \pm 0}$ & $\mathbf{100 \pm 0}$ \\
 & Rand. Prior (Osband 2019) & $99.8 \pm 0.39$ & $100 \pm 0$ \\
 & Bootstr. (Osband 2016a) & $99.2 \pm 0.72$ & $100 \pm 0$ \\
 Zero Init. \& Short Hor. & Bootstr. (D'Eramo 2019) & $84.8 \pm 4.33$ & $40 \pm 21.91$ \\
 & UCB1 (Auer 2002) & $97.4 \pm 0.99$ & $50 \pm 22.49$ \\
 & Expl. Bonus (Strehl 2008) & $99.8 \pm 0.39$ & $100 \pm 0$ \\
 & $\epsilon$-greedy & $97.4 \pm 1.17$ & $45 \pm 22.25$ \\
 & Random & $97.2 \pm 1.15$ & $45 \pm 22.25$ \\
 \hline
 & \textbf{Ours (UCB Reward)} & $\mathbf{100 \pm 0}$ & $\mathbf{100 \pm 0}$ \\
 & \textbf{Ours (Count Reward)} & $\mathbf{100 \pm 0}$ & $\mathbf{100 \pm 0}$ \\
 & Rand. Prior (Osband 2019) & $100 \pm 0$ & $100 \pm 0$ \\
 & Bootstr. (Osband 2016a) & $100 \pm 0$ & $100 \pm 0$ \\
 Opt. Init. \& Short Hor. & Bootstr. (D'Eramo 2019) & $100 \pm 0$ & $100 \pm 0$ \\
 & UCB1 (Auer 2002) & $100 \pm 0$ & $100 \pm 0$ \\
 & Expl. Bonus (Strehl 2008) & $100 \pm 0$ & $100 \pm 0$ \\
 & $\epsilon$-greedy & $100 \pm 0$ & $100 \pm 0$ \\
 & Random & $97.2 \pm 1.15$ & $45 \pm 22.25$ \\
 \hline
 & \textbf{Ours (UCB Reward)} & $\mathbf{100 \pm 0}$ & $\mathbf{100 \pm 0}$ \\
 & \textbf{Ours (Count Reward)} & $\mathbf{100 \pm 0}$ & $\mathbf{100 \pm 0}$ \\
 & Rand. Prior (Osband 2019) & ${100 \pm 0}$ & ${100 \pm 0}$ \\
 & Bootstr. (Osband 2016a) & $99.8 \pm 0.39$ & $100 \pm 0$ \\
 Zero Init. \& Long Hor. & Bootstr. (D'Eramo 2019) & $94.6 \pm 1.9$ & $70 \pm 20.49$ \\
 & UCB1 (Auer 2002) & $100 \pm 0$ & $98.5 \pm 0.22$ \\
 & Expl. Bonus (Strehl 2008) & ${100 \pm 0}$ & ${100 \pm 0}$ \\
 & $\epsilon$-greedy & ${100 \pm 0}$ & ${100 \pm 0}$ \\
 & Random & ${100 \pm 0}$ & ${100 \pm 0}$ \\
 \hline
 & \textbf{Ours (UCB Reward)} & $\mathbf{100 \pm 0}$ & $\mathbf{100 \pm 0}$ \\
 & \textbf{Ours (Count Reward)} & $\mathbf{100 \pm 0}$ & $\mathbf{100 \pm 0}$ \\
 & Rand. Prior (Osband 2019) & $100 \pm 0$ & $100 \pm 0$ \\
 & Bootstr. (Osband 2016a) & $100 \pm 0$ & $100 \pm 0$ \\
 Opt. Init. \& Long Hor. & Bootstr. (D'Eramo 2019) & $100 \pm 0$ & $100 \pm 0$ \\
 & UCB1 (Auer 2002) & $100 \pm 0$ & $100 \pm 0$ \\
 & Expl. Bonus (Strehl 2008) & $100 \pm 0$ & $100 \pm 0$ \\
 & $\epsilon$-greedy & $100 \pm 0$ & $100 \pm 0$ \\
 & Random & ${100 \pm 0}$ & ${100 \pm 0}$ \\
\end{tabular}
\end{table}

\begin{table}[h]
\centering
\caption{Gridworld (prison) results recap.}
\setstretch{1.15}
\begin{tabular}{ c | c | c c }
 & Algorithm & Discovery (\%) & Success (\%) \\
 \hline
 & \textbf{Ours (UCB Reward)} & $\mathbf{100 \pm 0}$ & $\mathbf{100 \pm 0}$ \\
 & \textbf{Ours (Count Reward)} & $\mathbf{100 \pm 0}$ & $\mathbf{100 \pm 0}$ \\
 & Rand. Prior (Osband 2019) & $72.39 \pm 6.9$ & $44.44 \pm 14.69$ \\
 & Bootstr. (Osband 2016a) & $64.35 \pm 11.1$ & $33.03 \pm 8.54$ \\
 Zero Init. \& Short Hor. & Bootstr. (D'Eramo 2019) & $54.78 \pm 6.69$ & $25.61 \pm 8.39$ \\
 & UCB1 (Auer 2002) & $80.78 \pm 2.56$ & $32.31 \pm 4.11$ \\
 & Expl. Bonus (Strehl 2008) & $85.87 \pm 3$ & $46.96 \pm 12.35$ \\
 & $\epsilon$-greedy & $58.38 \pm 5.27$ & $22.84 \pm 2.48$ \\
 & Random & $76.96 \pm 3.08$ & $35.36 \pm 10.37$ \\
 \hline
 & \textbf{Ours (UCB Reward)} & $\mathbf{100 \pm 0}$ & $\mathbf{100 \pm 0}$ \\
 & \textbf{Ours (Count Reward)} & $\mathbf{100 \pm 0}$ & $\mathbf{100 \pm 0}$ \\
 & Rand. Prior (Osband 2019) & ${100 \pm 0}$ & ${100 \pm 0}$ \\
 & Bootstr. (Osband 2016a) & ${100 \pm 0}$ & ${100 \pm 0}$ \\
 Opt. Init. \& Short Hor. & Bootstr. (D'Eramo 2019) & ${100 \pm 0}$ & ${100 \pm 0}$ \\
 & UCB1 (Auer 2002) & $100 \pm 0$ & $100 \pm 0$ \\
 & Expl. Bonus (Strehl 2008) & ${100 \pm 0}$ & ${100 \pm 0}$ \\
 & $\epsilon$-greedy & ${100 \pm 0}$ & ${100 \pm 0}$ \\
 & Random & $76.96 \pm 3.08$ & $35.36 \pm 10.37$ \\
 \hline
 & \textbf{Ours (UCB Reward)} & $\mathbf{100 \pm 0}$ & $\mathbf{100 \pm 0}$ \\
 & \textbf{Ours (Count Reward)} & $\mathbf{100 \pm 0}$ & $\mathbf{100 \pm 0}$ \\
 & Rand. Prior (Osband 2019) & $77.83 \pm 6.93$ & $56.04 \pm 14.79$ \\
 & Bootstr. (Osband 2016a) & $67.61 \pm 9.75$ & $44.12 \pm 13.02$ \\
 Zero Init. \& Long Hor. & Bootstr. (D'Eramo 2019) & $59.57 \pm 9.92$ & $29.31 \pm 4.11$ \\
 & UCB1 (Auer 2002) & $90.17 \pm 2.55$ & $51.55 \pm 10.67$ \\
 & Expl. Bonus (Strehl 2008) & $88.91 \pm 3.1$ & $54.48 \pm 11.63$ \\
 & $\epsilon$-greedy & $67.39 \pm 5.16$ & $24.69 \pm 3.31$ \\
 & Random & $86.09 \pm 3.77$ & $49.73 \pm 11.5$ \\
 \hline
 & \textbf{Ours (UCB Reward)} & $\mathbf{100 \pm 0}$ & $\mathbf{100 \pm 0}$ \\
 & \textbf{Ours (Count Reward)} & $\mathbf{100 \pm 0}$ & $\mathbf{100 \pm 0}$ \\
 & Rand. Prior (Osband 2019) & $100 \pm 0$ & $100 \pm 0$ \\
 & Bootstr. (Osband 2016a) & $100 \pm 0$ & $100 \pm 0$ \\
 Opt. Init. \& Long Hor. & Bootstr. (D'Eramo 2019) & $100 \pm 0$ & $100 \pm 0$ \\
 & UCB1 (Auer 2002) & $100 \pm 0$ & $100 \pm 0$ \\
 & Expl. Bonus (Strehl 2008) & $100 \pm 0$ & $100 \pm 0$ \\
 & $\epsilon$-greedy & $100 \pm 0$ & $100 \pm 0$ \\
 & Random & $86.09 \pm 3.77$ & $49.73 \pm 11.5$ \\
\end{tabular}
\end{table}

\begin{table}[h]
\centering
\caption{Gridworld (wall) results recap.}
\setstretch{1.15}
\begin{tabular}{ c | c | c c }
 & Algorithm & Discovery (\%) & Success (\%) \\
 \hline
 & \textbf{Ours (UCB Reward)} & $\mathbf{100 \pm 0}$ & $\mathbf{100 \pm 0}$ \\
 & \textbf{Ours (Count Reward)} & $\mathbf{100 \pm 0}$ & $\mathbf{100 \pm 0}$ \\
 & Rand. Prior (Osband 2019) & $9.44 \pm 3.14$ & $0.14 \pm 0.06$ \\
 & Bootstr. (Osband 2016a) & $6.35 \pm 2.74$ & $0.23 \pm 0.08$ \\
 Zero Init. \& Short Hor. & Bootstr. (D'Eramo 2019) & $12.55 \pm 4.45$ & $0.18 \pm 0.08$ \\
 & UCB1 (Auer 2002) & $24.7 \pm 4.45$ & $0.46 \pm 0.03$ \\
 & Expl. Bonus (Strehl 2008) & $45.16 \pm 2.29$ & $0.52 \pm 0.04$ \\
 & $\epsilon$-greedy & $38.15 \pm 1.95$ & $0.39 \pm 0.08$ \\
 & Random & $65.28 \pm 0.45$ & $0.59 \pm 0$ \\
 \hline
 & \textbf{Ours (UCB Reward)} & $\mathbf{100 \pm 0}$ & $\mathbf{100 \pm 0}$ \\
 & \textbf{Ours (Count Reward)} & $\mathbf{100 \pm 0}$ & $\mathbf{100 \pm 0}$ \\
 & Rand. Prior (Osband 2019) & $100 \pm 0$ & $100 \pm 0$ \\
 & Bootstr. (Osband 2016a) & $100 \pm 0$ & $100 \pm 0$ \\
 Opt. Init. \& Short Hor. & Bootstr. (D'Eramo 2019) & $100 \pm 0$ & $100 \pm 0$ \\
 & UCB1 (Auer 2002) & $100 \pm 0$ & $100 \pm 0$ \\
 & Expl. Bonus (Strehl 2008) & $100 \pm 0$ & $100 \pm 0$ \\
 & $\epsilon$-greedy & $100 \pm 0$ & $100 \pm 0$ \\
 & Random & $65.28 \pm 0.45$ & $0.59 \pm 0$ \\
 \hline
 & \textbf{Ours (UCB Reward)} & $\mathbf{100 \pm 0}$ & $\mathbf{100 \pm 0}$ \\
 & \textbf{Ours (Count Reward)} & $\mathbf{100 \pm 0}$ & $\mathbf{100 \pm 0}$ \\
 & Rand. Prior (Osband 2019) & $13.05 \pm 3.83$ & $0.2 \pm 0.08$ \\
 & Bootstr. (Osband 2016a) & $8.69 \pm 3.53$ & $0.2 \pm 0.08$ \\
 Zero Init. \& Long Hor. & Bootstr. (D'Eramo 2019) & $11.64 \pm 5.24$ & $0.23 \pm 0.08$ \\
 & UCB1 (Auer 2002) & $21.44 \pm 3.64$ & $0.44 \pm 0.03$ \\
 & Expl. Bonus (Strehl 2008) & $47.75 \pm 1.68$ & $0.55 \pm 0.03$ \\
 & $\epsilon$-greedy & $42.53 \pm 1.89$ & $0.54 \pm 0.03$ \\
 & Random & $70.51 \pm 0.75$ & $0.59 \pm 0$ \\
 \hline
 & \textbf{Ours (UCB Reward)} & $\mathbf{100 \pm 0}$ & $\mathbf{100 \pm 0}$ \\
 & \textbf{Ours (Count Reward)} & $\mathbf{100 \pm 0}$ & $\mathbf{100 \pm 0}$ \\
 & Rand. Prior (Osband 2019) & $100 \pm 0$ & $100 \pm 0$ \\
 & Bootstr. (Osband 2016a) & $100 \pm 0$ & $100 \pm 0$ \\
 Opt. Init. \& Long Hor. & Bootstr. (D'Eramo 2019) & $100 \pm 0$ & $100 \pm 0$ \\
 & UCB1 (Auer 2002) & $100 \pm 0$ & $100 \pm 0$ \\
 & Expl. Bonus (Strehl 2008) & $100 \pm 0$ & $100 \pm 0$ \\
 & $\epsilon$-greedy & $100 \pm 0$ & $100 \pm 0$ \\
 & Random & $70.51 \pm 0.75$ & $0.59 \pm 0$ \\
 \end{tabular}
\end{table}

\end{document}